\documentclass[10pt,letterpaper,twocolumn]{article}
\usepackage{tokaudit-preprint}
\usepackage{amsmath,amsfonts,bm}

\def\secref#1{section~\ref{#1}}
\def\eqref#1{(\ref{#1})}
\def\1{\bm{1}}

\DeclareMathAlphabet{\mathsfit}{\encodingdefault}{\sfdefault}{m}{sl}
\SetMathAlphabet{\mathsfit}{bold}{\encodingdefault}{\sfdefault}{bx}{n}

\usepackage{amsmath,amssymb,amsthm,bm}
\usepackage{booktabs}
\usepackage{graphicx}
\usepackage{multirow,multicol}
\usepackage{tabularx}
\usepackage{xcolor}
\usepackage{placeins}
\usepackage[pdfusetitle,hidelinks]{hyperref}
\usepackage{url}

\newcommand{\TAAgreeCoarse}{9}
\newcommand{\TAAgreeFine}{18}
\newcommand{\TAAgreeMid}{14}

\newcommand{\TABelowRefB}{51}

\newcommand{\TACTEmaBits}{5.04}
\newcommand{\TACTEmaRecon}{0.329}
\newcommand{\TACTEmaSupport}{9.4}
\newcommand{\TACTEmaSupportBase}{46}
\newcommand{\TACTEmaSupportCtl}{1024}
\newcommand{\TACTEntropyOff}{-0.93}
\newcommand{\TACTParamCount}{166340}
\newcommand{\TACTParamRecon}{0.985}
\newcommand{\TACTParamUtility}{6.7}
\newcommand{\TACVFourMedian}{0.961}
\newcommand{\TACVOneHighHi}{0.928}
\newcommand{\TACVOneHighLo}{0.766}
\newcommand{\TACVOneLowAHi}{0.998}
\newcommand{\TACVOneLowALo}{0.961}
\newcommand{\TACVOneLowBHi}{0.988}
\newcommand{\TACVOneLowBLo}{0.943}
\newcommand{\TACVOnePass}{9}

\newcommand{\TACVOneTotal}{9}
\newcommand{\TACVThreeModels}{135}
\newcommand{\TACVThreeViol}{0}
\newcommand{\TACVTwoSig}{9}
\newcommand{\TACVTwoTests}{9}
\newcommand{\TACells}{12}
\newcommand{\TACoarseReplicas}{32}
\newcommand{\TADGPBurnIn}{128}
\newcommand{\TADGPFactorLoad}{0.2}
\newcommand{\TADGPFactorNoise}{0.6}
\newcommand{\TADGPInnov}{0.25}
\newcommand{\TADGPRegimePeriod}{64}
\newcommand{\TADecGreedy}{-0.0217}
\newcommand{\TADecRecovery}{-0.0154}
\newcommand{\TADirectComparisons}{18}
\newcommand{\TADirectFlags}{1}
\newcommand{\TADirectMedianAbs}{0.0024}
\newcommand{\TADirectMedianRatio}{0.94}
\newcommand{\TADirectSeparated}{0}
\newcommand{\TADirectTests}{36}

\newcommand{\TADoseStateFour}{0.502}
\newcommand{\TADoseStateOne}{0.088}
\newcommand{\TADoseStateTwo}{0.215}

\newcommand{\TAFepsHigh}{0.5749}
\newcommand{\TAFepsLowA}{0.9068}
\newcommand{\TAFepsLowB}{0.9013}
\newcommand{\TAFreshExcessHi}{-0.002}
\newcommand{\TAFreshExcessLo}{-0.265}
\newcommand{\TAGainDetNoiseless}{0.000}
\newcommand{\TAGainDetNoisy}{1.000}
\newcommand{\TAGainRefNoiseless}{0.988}
\newcommand{\TAGainRefNoisy}{0.999}

\newcommand{\TAGaussExcessHi}{-0.007}
\newcommand{\TAGaussExcessLo}{-0.103}

\newcommand{\TAGaussN}{12}

\newcommand{\TAIVBase}{3.62\times 10^{-4}}
\newcommand{\TAIVCorrected}{-1.69\times 10^{-4}}
\newcommand{\TAIVEnhanced}{5.32\times 10^{-4}}
\newcommand{\TAIVInterval}{[-8.88\times 10^{-4}, -1.18\times 10^{-4}]}

\newcommand{\TAIntConfirmed}{4}
\newcommand{\TAIntNegativePoints}{15}
\newcommand{\TAIntTotal}{18}

\newcommand{\TAKOneRefHi}{0.9999}
\newcommand{\TAKOneRefLo}{0.9908}
\newcommand{\TAKThreeMedian}{1.000}

\newcommand{\TAKTwoComparisons}{48}
\newcommand{\TAKTwoConfigs}{24}
\newcommand{\TAKTwoErrHi}{2.58\times10^{-4}}
\newcommand{\TAKTwoErrLo}{1.83\times10^{-9}}
\newcommand{\TAKTwoRatioHi}{1.54}
\newcommand{\TAKTwoRatioLo}{1.01}

\newcommand{\TAKTwoViolMaxSe}{3}
\newcommand{\TAKTwoViolOwn}{5}

\newcommand{\TAMCBiasEndHi}{2.75\times 10^{-5}}
\newcommand{\TAMCBiasEndLo}{4.81\times 10^{-6}}
\newcommand{\TAMCBiasReplayHi}{1.76\times 10^{-3}}
\newcommand{\TAMCBiasReplayLo}{3.21\times 10^{-4}}
\newcommand{\TAMatchedEntropyTen}{0.0912}
\newcommand{\TAMatchedEntropyThree}{0.0144}
\newcommand{\TAMatchedEntropyTol}{0.05}
\newcommand{\TAMatchedPerSeedTen}{1}
\newcommand{\TAMatchedPerSeedThree}{1}
\newcommand{\TAMatchedReconTen}{0.202}
\newcommand{\TAMatchedReconThree}{0.053}
\newcommand{\TAMatchedReconTol}{0.10}
\newcommand{\TAMatchedRows}{135}
\newcommand{\TAMatchedSeedsTen}{10}
\newcommand{\TAMatchedShareHi}{0.998}
\newcommand{\TAMatchedShareLo}{0.887}

\newcommand{\TAMidReplicas}{128}
\newcommand{\TANEntities}{64}

\newcommand{\TANFitB}{315}

\newcommand{\TANSeeds}{5}
\newcommand{\TANSignalDim}{4}
\newcommand{\TANTimes}{256}
\newcommand{\TANTransitionsA}{18}
\newcommand{\TANTransitionsB}{54}
\newcommand{\TAOccSpanBsqHi}{1.76}
\newcommand{\TAOccSpanBsqLo}{1.73}
\newcommand{\TAOccSpanFsqHi}{2.03}
\newcommand{\TAOccSpanFsqLo}{1.90}
\newcommand{\TAOccSpanVqvaeHi}{0.48}
\newcommand{\TAOccSpanVqvaeLo}{0.44}
\newcommand{\TAPROBEHistory}{4}
\newcommand{\TAPSCellHi}{1.000}
\newcommand{\TAPSCellLo}{0.939}
\newcommand{\TAPSSign}{18/18}
\newcommand{\TAPSTransitions}{18}
\newcommand{\TAPanelsTest}{32}
\newcommand{\TAPanelsTrain}{8}
\newcommand{\TAPanelsValidation}{8}

\newcommand{\TAPopSeeds}{5}
\newcommand{\TAPopTransitions}{32}
\newcommand{\TAPositiveExcessB}{2}
\newcommand{\TAPositiveOccupancyB}{5}
\newcommand{\TAPositiveReconB}{5}
\newcommand{\TAPositiveSignalB}{5}
\newcommand{\TARateCorrectedStep}{6.3692}

\newcommand{\TARealBurnIn}{672}
\newcommand{\TARealGainHi}{0.663}
\newcommand{\TARealGainLo}{0.344}
\newcommand{\TARealLagOneHi}{0.990}
\newcommand{\TARealLagOneLo}{0.793}
\newcommand{\TARealRepeats}{3}
\newcommand{\TARealReplicas}{512}
\newcommand{\TARealSeasonalDays}{7}
\newcommand{\TARealSeasonalLag}{96}
\newcommand{\TARealShareHi}{0.837}
\newcommand{\TARealShareLo}{0.518}
\newcommand{\TARealStates}{2048}

\newcommand{\TARegRhoHigh}{0.90}
\newcommand{\TARegRhoLow}{0.55}
\newcommand{\TARegSNRHigh}{1.0}
\newcommand{\TARegSNRLow}{0.1}
\newcommand{\TARegScaleHigh}{1.10}
\newcommand{\TARegScaleLow}{0.65}
\newcommand{\TARegretLargest}{0.00116}
\newcommand{\TARegretOneSE}{0.00057}
\newcommand{\TARegretState}{0.00126}
\newcommand{\TARegretValidation}{4.71\times 10^{-6}}
\newcommand{\TAReplicaRefCoarse}{512}
\newcommand{\TAReplicaRefFine}{2048}
\newcommand{\TAReplicaRungFine}{512}
\newcommand{\TAResolvedPos}{17}

\newcommand{\TARhoIntervalCorrected}{[-0.569, -0.468]}
\newcommand{\TARhoIntervalFrozen}{[-0.584, -0.474]}
\newcommand{\TARhoPoint}{-0.538}
\newcommand{\TASaugeenBitsPerPoint}{0.25}
\newcommand{\TASaugeenBitsPerToken}{1}
\newcommand{\TASaugeenBitsPerWindow}{90}
\newcommand{\TASaugeenCleanModels}{12}
\newcommand{\TASaugeenCommitment}{0.25}
\newcommand{\TASaugeenEffSamples}{12}
\newcommand{\TASaugeenGenBatch}{1024}
\newcommand{\TASaugeenGenUpdates}{5000}

\newcommand{\TASaugeenInterval}{[-0.771, +0.571]}
\newcommand{\TASaugeenMeanEig}{1.0}
\newcommand{\TASaugeenMinEig}{-1.50}
\newcommand{\TASaugeenNDesigned}{9}
\newcommand{\TASaugeenNUsed}{8}

\newcommand{\TASaugeenPoints}{23741}

\newcommand{\TASaugeenRho}{-0.286}
\newcommand{\TASaugeenSampleRatio}{2.5}
\newcommand{\TASaugeenScratchBatch}{512}
\newcommand{\TASaugeenScratchUpdates}{4000}
\newcommand{\TASaugeenSign}{0.68}
\newcommand{\TASaugeenTranslate}{1024}

\newcommand{\TAScoringModelMax}{4.50\times 10^{-5}}
\newcommand{\TAScoringModelMedian}{5.76\times 10^{-8}}

\newcommand{\TASecondEfficiency}{-0.468}
\newcommand{\TASecondRate}{0.562}
\newcommand{\TASecondRecon}{0.799}
\newcommand{\TASecondSignal}{0.815}

\newcommand{\TASpearmanExcessB}{-0.470}
\newcommand{\TASpearmanOccupancyB}{0.440}
\newcommand{\TASpearmanReconB}{0.497}
\newcommand{\TASpearmanSignalB}{0.601}

\newcommand{\TAStrata}{9}

\newcommand{\TATokBSQEntropy}{0.01}
\newcommand{\TATokBatch}{8{,}192}
\newcommand{\TATokBatchesPerEpoch}{16}
\newcommand{\TATokClipNorm}{1.0}
\newcommand{\TATokEmbed}{64}
\newcommand{\TATokEpochs}{100}
\newcommand{\TATokFSQLevelsFour}{4\times 4}
\newcommand{\TATokFSQLevelsSeven}{8\times 4\times 4}
\newcommand{\TATokFSQLevelsTen}{8\times 8\times 4\times 4}
\newcommand{\TATokHidden}{192}
\newcommand{\TATokLR}{10^{-4}}
\newcommand{\TATokParamsMax}{158{,}532}
\newcommand{\TATokParamsMin}{27{,}972}
\newcommand{\TATokProbeAlphas}{10^{-4},\ldots,10^{4}}
\newcommand{\TATokProbeAlphasN}{9}
\newcommand{\TATokRows}{131{,}072}
\newcommand{\TATokSeeds}{42, 43, 44, 45, 46}
\newcommand{\TATokSnapEvery}{200}
\newcommand{\TATokSnapshots}{8}
\newcommand{\TATokStages}{2}
\newcommand{\TATokSteps}{1{,}600}
\newcommand{\TATokVQCommit}{0.25}
\newcommand{\TATolAbs}{10^{-5}}
\newcommand{\TATolRel}{0.10}
\newcommand{\TATotalA}{18}
\newcommand{\TATotalB}{54}

\newcommand{\TATrainFitPanels}{6}
\newcommand{\TAUnresolved}{1}

\newcommand{\TACrossBurnIn}{2048}
\newcommand{\TACrossCodecs}{96}
\newcommand{\TACrossFamilies}{2}
\newcommand{\TACrossGenOneBelowRef}{12}
\newcommand{\TACrossGenOneInterval}{[-0.829, -0.514]}
\newcommand{\TACrossGenOnePooled}{-0.364}
\newcommand{\TACrossGenOneQ}{-0.657}
\newcommand{\TACrossGenOneRefRankCorr}{0.958}
\newcommand{\TACrossGenOneShareMedian}{0.899}
\newcommand{\TACrossGenOneTransitions}{12}

\newcommand{\TACrossGenOneVerdict}{supported}
\newcommand{\TACrossGenTwoBelowRef}{12}
\newcommand{\TACrossGenTwoInterval}{[-0.829, -0.029]}
\newcommand{\TACrossGenTwoPooled}{-0.238}
\newcommand{\TACrossGenTwoQ}{-0.829}
\newcommand{\TACrossGenTwoRefRankCorr}{0.979}
\newcommand{\TACrossGenTwoShareMedian}{0.900}
\newcommand{\TACrossGenTwoTransitions}{12}

\newcommand{\TACrossGenTwoVerdict}{supported}
\newcommand{\TACrossGenerators}{2}
\newcommand{\TACrossGridBoundary}{71}
\newcommand{\TACrossGridPoints}{9}
\newcommand{\TACrossInitializations}{3}
\newcommand{\TACrossPairs}{1024}
\newcommand{\TACrossPerStratum}{6}
\newcommand{\TACrossPilotBiasHigh}{3.67\times 10^{-4}}
\newcommand{\TACrossPilotBiasLow}{4.58\times 10^{-5}}
\newcommand{\TACrossPilotBiasMid}{9.16\times 10^{-5}}
\newcommand{\TACrossPilotFits}{8}
\newcommand{\TACrossReplicaHigh}{4096}
\newcommand{\TACrossReplicaMid}{2048}
\newcommand{\TACrossReplicas}{2048}
\newcommand{\TACrossSnrs}{2}
\newcommand{\TACrossStates}{2048}
\newcommand{\TACrossTransitions}{24}

\newcommand{\TACrossWidths}{4}
\newcommand{\TAReaderAlphaBOne}{0.025}
\newcommand{\TAReaderAlphaBTwo}{0.0125}
\newcommand{\TAReaderBOneCodecs}{220}
\newcommand{\TAReaderBOneGru}{-0.702}
\newcommand{\TAReaderBOneGruHi}{-0.571}
\newcommand{\TAReaderBOneGruInterval}{[-0.792, -0.506]}
\newcommand{\TAReaderBOneGruLo}{-0.786}
\newcommand{\TAReaderBOneH}{4}
\newcommand{\TAReaderBOneMlp}{-0.720}
\newcommand{\TAReaderBOneMlpHi}{-0.667}
\newcommand{\TAReaderBOneMlpInterval}{[-0.804, -0.530]}
\newcommand{\TAReaderBOneMlpLo}{-0.762}
\newcommand{\TAReaderBOneRegimes}{4}
\newcommand{\TAReaderBTwoClaims}{4}
\newcommand{\TAReaderBTwoCodecs}{96}
\newcommand{\TAReaderBTwoGruGenOne}{-0.629}
\newcommand{\TAReaderBTwoGruGenOneInterval}{[-0.886, -0.629]}
\newcommand{\TAReaderBTwoGruGenTwo}{-0.743}
\newcommand{\TAReaderBTwoGruGenTwoInterval}{[-0.914, -0.086]}
\newcommand{\TAReaderBTwoH}{32}
\newcommand{\TAReaderBTwoMlpGenOne}{-0.629}
\newcommand{\TAReaderBTwoMlpGenOneInterval}{[-0.857, -0.600]}
\newcommand{\TAReaderBTwoMlpGenTwo}{-0.657}
\newcommand{\TAReaderBTwoMlpGenTwoInterval}{[-0.829, -0.043]}
\newcommand{\TAReaderCompetenceMlp}{0.764}
\newcommand{\TAReaderCompetenceRidge}{0.489}
\newcommand{\TAReaderInits}{2}
\newcommand{\TAReaderNearZeroBOne}{36}
\newcommand{\TAReaderNearZeroBand}{0.005}
\newcommand{\TAReaderNonlinear}{2}
\newcommand{\TAReaderOrderAlpha}{5.56\times 10^{-3}}
\newcommand{\TAReaderOrderComputed}{6}
\newcommand{\TAReaderOrderDeclared}{9}
\newcommand{\TAReaderOrderLoMax}{+0.833}
\newcommand{\TAReaderOrderLoMin}{+0.571}
\newcommand{\TAReaderOrderThreshold}{0.5}
\newcommand{\TAReaderUpdates}{2000}

\newcommand{\TAIVFolds}{4}
\newcommand{\TAIVFoldsImproving}{3}
\newcommand{\TAIVSecondaryBase}{0.0003623761875}
\newcommand{\TAIVSecondaryChange}{-0.0002407254215}
\newcommand{\TAIVSecondaryEnhanced}{0.000603101609}

\newcommand{\TAPatBootstrapTransitions}{32}
\newcommand{\TAPatBootstrapDrawSize}{5}
\newcommand{\TAPatBootstrapAtoms}{126}

\newcommand{\TAPatBootstrapRhoPoint}{-0.5381231671554252}
\newcommand{\TAPatBootstrapIdentityWeight}{0.0384}
\newcommand{\TAPatBootstrapStepLo}{-0.5692815249266863}
\newcommand{\TAPatBootstrapStepHi}{-0.4677419354838709}
\newcommand{\TAPatBootstrapLinearHi}{-0.4695381231671553}
\newcommand{\TAPatBootstrapDistinct}{103}
\newcommand{\TAPatBootstrapSupportMin}{-0.5879765395894428}
\newcommand{\TAPatBootstrapSupportMax}{-0.4266862170087977}
\newcommand{\TAPatBootstrapNegMass}{1.0000}

\newcommand{\TAPatBootstrapMaxDev}{0.000}
\newcommand{\TAPatBootstrapTv}{0.0365}
\newcommand{\TAPatBootstrapEqualLo}{-0.5762463343108505}
\newcommand{\TAPatBootstrapEqualHi}{-0.4475806451612903}
\newcommand{\TAPatBootstrapWidthPct}{27}
\newcommand{\TAPatCapacityABaseValue}{-0.5381}
\newcommand{\TAPatCapacityABaseN}{32}
\newcommand{\TAPatCapacityABaseLo}{-0.5693}
\newcommand{\TAPatCapacityABaseHi}{-0.4677}
\newcommand{\TAPatCapacityAWithinValue}{-0.6667}
\newcommand{\TAPatCapacityAWithinN}{32}
\newcommand{\TAPatCapacityAWithinLo}{-0.7440}
\newcommand{\TAPatCapacityAWithinHi}{-0.5476}
\newcommand{\TAPatCapacityALowerValue}{-0.5121}
\newcommand{\TAPatCapacityALowerN}{32}
\newcommand{\TAPatCapacityALowerLo}{-0.5526}
\newcommand{\TAPatCapacityALowerHi}{-0.3647}
\newcommand{\TAPatCapacityAStepValue}{-0.5459}
\newcommand{\TAPatCapacityAStepN}{32}
\newcommand{\TAPatCapacityAStepLo}{-0.5900}
\newcommand{\TAPatCapacityAStepHi}{-0.3698}
\newcommand{\TAPatCapacityAOccupancyValue}{-0.5458}
\newcommand{\TAPatCapacityAOccupancyN}{32}
\newcommand{\TAPatCapacityAOccupancyLo}{-0.5856}
\newcommand{\TAPatCapacityAOccupancyHi}{-0.3696}
\newcommand{\TAPatCapacityAFamilyValue}{-0.5466}
\newcommand{\TAPatCapacityAFamilyN}{32}
\newcommand{\TAPatCapacityAFamilyLo}{-0.5623}
\newcommand{\TAPatCapacityAFamilyHi}{-0.3440}
\newcommand{\TAPatCapacityARungDisjointValue}{-0.5008}
\newcommand{\TAPatCapacityARungDisjointN}{20}
\newcommand{\TAPatCapacityARungDisjointLo}{-0.5895}
\newcommand{\TAPatCapacityARungDisjointHi}{-0.2992}

\newcommand{\TAPatCapacityABetween}{-0.600}
\newcommand{\TAPatCapacityAWithin}{-0.667}
\newcommand{\TAPatCapacityAPairs}{28}
\newcommand{\TAPatCapacityAConcordant}{19}
\newcommand{\TAPatCapacityAContrastRho}{-0.352}
\newcommand{\TAPatCapacityAContrastP}{0.087}
\newcommand{\TAPatCapacityADesignRank}{4}
\newcommand{\TAPatCapacityAResidualDf}{28}
\newcommand{\TAPatCapacityBBaseValue}{-0.2870}
\newcommand{\TAPatCapacityBBaseN}{24}
\newcommand{\TAPatCapacityBBaseLo}{-0.3887}
\newcommand{\TAPatCapacityBBaseHi}{-0.1348}
\newcommand{\TAPatCapacityBWithinValue}{-0.7429}
\newcommand{\TAPatCapacityBWithinN}{24}
\newcommand{\TAPatCapacityBWithinLo}{-0.7429}
\newcommand{\TAPatCapacityBWithinHi}{-0.4429}
\newcommand{\TAPatCapacityBLowerValue}{-0.0450}
\newcommand{\TAPatCapacityBLowerN}{24}
\newcommand{\TAPatCapacityBLowerLo}{-0.2147}
\newcommand{\TAPatCapacityBLowerHi}{+0.0361}
\newcommand{\TAPatCapacityBStepValue}{-0.0554}
\newcommand{\TAPatCapacityBStepN}{24}
\newcommand{\TAPatCapacityBStepLo}{-0.2290}
\newcommand{\TAPatCapacityBStepHi}{+0.0440}
\newcommand{\TAPatCapacityBOccupancyValue}{+0.0259}
\newcommand{\TAPatCapacityBOccupancyN}{24}
\newcommand{\TAPatCapacityBOccupancyLo}{-0.2004}
\newcommand{\TAPatCapacityBOccupancyHi}{+0.2199}
\newcommand{\TAPatCapacityBFamilyValue}{+0.4171}
\newcommand{\TAPatCapacityBFamilyN}{24}
\newcommand{\TAPatCapacityBFamilyLo}{-0.0641}
\newcommand{\TAPatCapacityBFamilyHi}{+0.4171}
\newcommand{\TAPatCapacityBRungDisjointValue}{-0.2471}
\newcommand{\TAPatCapacityBRungDisjointN}{16}
\newcommand{\TAPatCapacityBRungDisjointLo}{-0.4676}
\newcommand{\TAPatCapacityBRungDisjointHi}{-0.0412}

\newcommand{\TAPatCapacityBBetween}{+0.400}
\newcommand{\TAPatCapacityBWithin}{-0.743}
\newcommand{\TAPatCapacityBPairs}{12}
\newcommand{\TAPatCapacityBConcordant}{10}
\newcommand{\TAPatCapacityBContrastRho}{-0.336}
\newcommand{\TAPatCapacityBContrastP}{0.039}

\newcommand{\TAPatCapacityAAtoms}{126}
\newcommand{\TAPatCapacityBAtoms}{10}
\newcommand{\TAPatCapacityANegAtoms}{126}
\newcommand{\TAPatCapacityAMin}{-0.6383}
\newcommand{\TAPatCapacityAMax}{-0.2746}
\newcommand{\TAPatCapacityBPosWeightPct}{63}
\newcommand{\TAPatCapacityBDf}{20}
\newcommand{\TAPatCapacityAVif}{2.8}
\newcommand{\TAPatCapacityBVif}{4.0}
\newcommand{\TAPatRidgeCodecs}{96}

\newcommand{\TAPatRidgeExtendedBoundary}{0}
\newcommand{\TAPatRidgeMoved}{21}
\newcommand{\TAPatRidgeInnerChange}{+1.02\times10^{-4}}
\newcommand{\TAPatRidgeTestMeanChange}{-3.52\times10^{-5}}
\newcommand{\TAPatRidgeTestMaxChange}{1.29\times10^{-3}}
\newcommand{\TAPatRidgeRankAll}{0.9997}
\newcommand{\TAPatRidgeRankTransitions}{0.9948}
\newcommand{\TAPatRidgeFrozenGrid}{9}
\newcommand{\TAPatRidgeExtendedGrid}{12}
\newcommand{\TAPatRidgeQOneFrozen}{-0.6571}

\newcommand{\TAPatRidgeQTwoFrozen}{-0.8286}

\newcommand{\TAPatRidgePooledOneFrozen}{-0.3636}
\newcommand{\TAPatRidgePooledOneExtended}{-0.3287}
\newcommand{\TAPatRidgePooledTwo}{-0.2378}
\newcommand{\TAPatRidgeDirectionAgreement}{0.995}
\newcommand{\TAPatScalarTransitions}{42}
\newcommand{\TAPatScalarShareMedian}{0.9985}
\newcommand{\TAPatScalarShareLo}{0.885}
\newcommand{\TAPatScalarShareHi}{1.006}
\newcommand{\TAPatScalarAboveHalf}{42}
\newcommand{\TAPatScalarBelow}{20}
\newcommand{\TAPatScalarAbove}{22}
\newcommand{\TAPatScalarExcessMedian}{+0.0001}
\newcommand{\TAPatScalarRho}{+0.378}
\newcommand{\TAPatScalarLearnedRhoOne}{-0.364}
\newcommand{\TAPatScalarLearnedRhoTwo}{-0.238}
\newcommand{\TAPatScalarLearnedShareOne}{0.899}
\newcommand{\TAPatScalarLearnedShareTwo}{0.900}
\newcommand{\TAPatScalarLearnedBelow}{24}
\newcommand{\TAPatScalarLearnedTransitions}{24}

\newcommand{\TAPatScalarUniformShare}{0.9995}
\newcommand{\TAPatScalarUniformExcess}{+0.0009}
\newcommand{\TAPatScalarUniformBelow}{5}
\newcommand{\TAPatScalarUniformAbove}{16}
\newcommand{\TAPatScalarUniformRho}{+0.699}

\newcommand{\TAPatScalarQuantileShare}{0.9885}
\newcommand{\TAPatScalarQuantileExcess}{-0.0103}
\newcommand{\TAPatScalarQuantileBelow}{15}
\newcommand{\TAPatScalarQuantileAbove}{6}
\newcommand{\TAPatScalarQuantileRho}{+0.484}
\newcommand{\TAPatScalarExpThirtyNineTransitions}{24}
\newcommand{\TAPatScalarExpThirtyNineShare}{0.9995}
\newcommand{\TAPatScalarExpThirtyNineBelow}{9}
\newcommand{\TAPatScalarExpThirtyNineAbove}{15}
\newcommand{\TAPatScalarExpThirtyNineRho}{+0.575}
\newcommand{\TAPatScalarExpSixteenTransitions}{18}
\newcommand{\TAPatScalarExpSixteenShare}{0.9877}
\newcommand{\TAPatScalarExpSixteenBelow}{11}
\newcommand{\TAPatScalarExpSixteenAbove}{7}
\newcommand{\TAPatScalarExpSixteenRho}{+0.236}
\newcommand{\TAPatScalarLearnedAbove}{0}
\newcommand{\TAPatScalarLearnedAboveHalf}{23}
\newcommand{\TAPatScalarBitsLo}{1}
\newcommand{\TAPatScalarBitsHi}{4}
\newcommand{\TAPatScalarChannels}{4}
\newcommand{\TAPatScalarStages}{2}
\newcommand{\TAPatScalarRateFactor}{8}
\newcommand{\TAPatScalarRateLevels}{8, 16, 24 and 32}
\newcommand{\TAPatScalarReaderHistory}{32}
\newcommand{\TAPatScalarRateMatchPairs}{24}
\newcommand{\TAPatScalarRateMatchGapMin}{0.210}
\newcommand{\TAPatScalarRateMatchGapMedian}{3.229}
\newcommand{\TAPatScalarRateMatchGapMax}{4.896}
\newcommand{\TAPatTemporalCodecs}{96}
\newcommand{\TAPatTemporalTransitions}{24}
\newcommand{\TAPatTemporalParamsPointLo}{27{,}972}
\newcommand{\TAPatTemporalParamsPointHi}{158{,}532}
\newcommand{\TAPatTemporalParamsTemporalLo}{29{,}700}
\newcommand{\TAPatTemporalParamsTemporalHi}{160{,}260}
\newcommand{\TAPatTemporalParamsAdded}{1{,}728}
\newcommand{\TAPatTemporalTestMsePoint}{0.24575}
\newcommand{\TAPatTemporalTestMseTemporal}{0.24564}
\newcommand{\TAPatTemporalTestRatioMedian}{0.9998}
\newcommand{\TAPatTemporalTestRatioMean}{0.9999}
\newcommand{\TAPatTemporalTestRatioLo}{0.958}
\newcommand{\TAPatTemporalTestRatioHi}{1.033}
\newcommand{\TAPatTemporalValidationMseTemporal}{0.24640}
\newcommand{\TAPatTemporalSampleSize}{96}
\newcommand{\TAPatTemporalOnePoint}{-0.00071}
\newcommand{\TAPatTemporalOneLo}{-0.00158}
\newcommand{\TAPatTemporalOneHi}{+0.00026}

\newcommand{\TAPatTemporalTwoPoint}{-0.00020}
\newcommand{\TAPatTemporalTwoLo}{-0.00171}
\newcommand{\TAPatTemporalTwoHi}{+0.00090}

\newcommand{\TAPatTemporalPointShare}{0.8993}
\newcommand{\TAPatTemporalPointShareLo}{0.4985}
\newcommand{\TAPatTemporalPointShareHi}{0.9798}
\newcommand{\TAPatTemporalPointBelow}{24}
\newcommand{\TAPatTemporalPointExcess}{-0.0527}
\newcommand{\TAPatTemporalPointUtility}{0.0131}
\newcommand{\TAPatTemporalPointReconstruction}{0.0775}
\newcommand{\TAPatTemporalPointOccupancy}{1.7662}
\newcommand{\TAPatTemporalPointPooledRho}{-0.2687}
\newcommand{\TAPatTemporalPointQOne}{-0.6571}
\newcommand{\TAPatTemporalPointQTwo}{-0.8286}
\newcommand{\TAPatTemporalTemporalShare}{0.8968}
\newcommand{\TAPatTemporalTemporalShareLo}{0.4964}
\newcommand{\TAPatTemporalTemporalShareHi}{0.9797}
\newcommand{\TAPatTemporalTemporalBelow}{24}
\newcommand{\TAPatTemporalTemporalExcess}{-0.0517}
\newcommand{\TAPatTemporalTemporalUtility}{0.0134}
\newcommand{\TAPatTemporalTemporalReconstruction}{0.0766}
\newcommand{\TAPatTemporalTemporalOccupancy}{1.7688}
\newcommand{\TAPatTemporalTemporalPooledRho}{-0.2635}
\newcommand{\TAPatTemporalTemporalQOne}{-0.6000}
\newcommand{\TAPatTemporalTemporalQTwo}{-0.8286}
\newcommand{\TAPatTemporalShareMove}{0.0025}
\newcommand{\TAPatTemporalOccupancyMove}{+0.0026}
\newcommand{\TAPatTemporalPooledMoveLow}{-0.263}
\newcommand{\TAPatTemporalPooledMoveHigh}{-0.269}
\newcommand{\TAPatReplayTruthPhiZero}{0.16563}
\newcommand{\TAPatReplayTruthPhiHalf}{0.25201}
\newcommand{\TAPatReplayTruthPhiNine}{0.15440}
\newcommand{\TAPatReplayDsPhiZero}{0.00518}
\newcommand{\TAPatReplayDsPhiHalf}{0.00579}
\newcommand{\TAPatReplayDsPhiNine}{0.00860}
\newcommand{\TAPatReplayRatioPhiZero}{0.0313}
\newcommand{\TAPatReplayRatioPhiHalf}{0.0230}
\newcommand{\TAPatReplayRatioPhiNine}{0.0557}
\newcommand{\TAPatReplayTargetEstimatorLo}{0.9873}
\newcommand{\TAPatReplayTargetEstimatorHi}{1.0026}
\newcommand{\TAPatReplayTargetTruthLo}{0.0218}
\newcommand{\TAPatReplayTargetTruthHi}{0.0606}
\newcommand{\TAPatReplayConditionalPhiZero}{0.0301}
\newcommand{\TAPatReplayOraclePhiZero}{0.0441}
\newcommand{\TAPatReplaySpreadZero}{0.0}
\newcommand{\TAPatReplaySpreadSmall}{5.6\times10^{-17}}
\newcommand{\TAPatReplaySpreadLast}{0.0}
\newcommand{\TAPatReplayExplainLo}{7}
\newcommand{\TAPatReplayExplainHi}{11}
\newcommand{\TAPatReplayStateLo}{2}
\newcommand{\TAPatReplayStateHi}{6}
\newcommand{\TAPatReplayResidualAc}{0.80}
\newcommand{\TAPatReplayBlockAc}{0.79}
\newcommand{\TAPatReplayIidAc}{0.0005}
\newcommand{\TAOccSpanHighPredHighNoiseBsqD}{1.76}
\newcommand{\TAOccSpanHighPredHighNoiseBsqU}{+0.0379}
\newcommand{\TAOccSpanHighPredHighNoiseFsqD}{1.90}
\newcommand{\TAOccSpanHighPredHighNoiseFsqU}{+0.0089}
\newcommand{\TAOccSpanHighPredHighNoiseVqvaeD}{0.47}
\newcommand{\TAOccSpanHighPredHighNoiseVqvaeU}{-0.0012}
\newcommand{\TAOccSpanHighPredLowNoiseBsqD}{1.73}
\newcommand{\TAOccSpanHighPredLowNoiseBsqU}{+0.0154}
\newcommand{\TAOccSpanHighPredLowNoiseFsqD}{1.99}
\newcommand{\TAOccSpanHighPredLowNoiseFsqU}{+0.0032}
\newcommand{\TAOccSpanHighPredLowNoiseVqvaeD}{0.44}
\newcommand{\TAOccSpanHighPredLowNoiseVqvaeU}{-0.0014}
\newcommand{\TAOccSpanLowPredLowNoiseBsqD}{1.73}
\newcommand{\TAOccSpanLowPredLowNoiseBsqU}{+0.0061}
\newcommand{\TAOccSpanLowPredLowNoiseFsqD}{2.03}
\newcommand{\TAOccSpanLowPredLowNoiseFsqU}{+0.0004}
\newcommand{\TAOccSpanLowPredLowNoiseVqvaeD}{0.48}
\newcommand{\TAOccSpanLowPredLowNoiseVqvaeU}{+0.0005}

\newcommand{\TAExpFortySevenQ}{-0.501}
\newcommand{\TAExpFortySevenInterval}{[-0.574,-0.138]}
\newcommand{\TAExpFortySevenRhoLow}{-0.545}
\newcommand{\TAExpFortySevenRhoHigh}{-0.457}
\newcommand{\TAExpFortySevenBelowLow}{24}
\newcommand{\TAExpFortySevenBelowHigh}{23}
\newcommand{\TAExpFortySevenTransitions}{24}
\newcommand{\TAExpFortySevenBootstrapReplicates}{10000}
\newcommand{\TAExpFortySevenCodecs}{48}
\newcommand{\TAExpFortySevenSeeds}{2}
\newcommand{\TAExpFortySevenStations}{12}
\newcommand{\TAExpFortySevenChannels}{4}
\newcommand{\TAExpFortySevenHours}{28513}
\newcommand{\TAExpFortySevenPanelHours}{168}
\newcommand{\TAExpFortySevenCleanWorst}{0.0}

\newcommand{\TAExpFortySevenLooSeasonLo}{-0.564}
\newcommand{\TAExpFortySevenLooSeasonHi}{-0.401}
\newcommand{\TAExpFortySevenMechBsq}{-0.794}
\newcommand{\TAExpFortySevenMechVqvae}{-0.378}
\newcommand{\TAExpFortySevenStationLo}{-0.538}
\newcommand{\TAExpFortySevenStationHi}{-0.456}

\newcommand{\TAExpFortySevenRetained}{28513}
\newcommand{\TAExpFortySevenHourlyPct}{99.40}
\newcommand{\TAExpFortySevenGapSteps}{172}
\newcommand{\TAExpFortySevenRetainedSteps}{28512}
\newcommand{\TAExpFortySevenLargestGapHours}{889}
\newcommand{\TAExpFortySevenWindowCrossTestPct}{3.83}
\newcommand{\TAExpFortySevenWindowCrossAllPct}{5.50}
\newcommand{\TAExpFortySevenLabelCrossTestPct}{0.77}
\newcommand{\TAExpFortySevenInterpFilled}{27962}
\newcommand{\TAExpFortySevenInterpMaxDistance}{6}
\newcommand{\TAExpFiftyQ}{-0.480}
\newcommand{\TAExpFiftyInterval}{[-0.593,-0.102]}

\newcommand{\TAExpFiftyDropped}{29}

\newcommand{\TAExpFortyEightCodecs}{20}
\newcommand{\TAExpFortyEightSeeds}{5}
\newcommand{\TAExpFortyEightLambda}{0.10}
\newcommand{\TAExpFortyEightBandGain}{0.10}
\newcommand{\TAExpFortyEightBandOccupancy}{0.05}
\newcommand{\TAExpFortyEightBandLevel}{0.15}
\newcommand{\TAExpFortyEightGainRelPct}{0.7}
\newcommand{\TAExpFortyEightOccGap}{0.0006}
\newcommand{\TAExpFortyEightLevelLowPct}{2.6}
\newcommand{\TAExpFortyEightLevelHighPct}{4.6}
\newcommand{\TAExpFortyEightGainPlain}{0.1096}
\newcommand{\TAExpFortyEightGainConsistent}{0.1089}
\newcommand{\TAExpFortyEightOccPlain}{2.333}
\newcommand{\TAExpFortyEightOccConsistent}{2.332}
\newcommand{\TAExpFortyEightSharePlain}{0.772}
\newcommand{\TAExpFortyEightShareConsistent}{0.734}
\newcommand{\TAExpFortyEightStatePlain}{0.0250}
\newcommand{\TAExpFortyEightStateConsistent}{0.0290}
\newcommand{\TAExpFortyEightNoisePlain}{0.0846}
\newcommand{\TAExpFortyEightNoiseConsistent}{0.0800}
\newcommand{\TAExpFortyEightExcessPlain}{-0.159}
\newcommand{\TAExpFortyEightExcessConsistent}{-0.197}
\newcommand{\TAExpFortyEightRefShare}{0.931}
\newcommand{\TAExpFortyEightDeltaShare}{-0.038}
\newcommand{\TAExpFortyEightDeltaShareLo}{-0.048}
\newcommand{\TAExpFortyEightDeltaShareHi}{-0.027}
\newcommand{\TAExpFortyEightDeltaUtility}{-0.00003}
\newcommand{\TAExpFortyEightSupportMin}{-0.055}
\newcommand{\TAExpFortyEightSupportMax}{-0.018}
\newcommand{\TAExpFortyEightAtoms}{126}

\newcommand{\TAAltConflA}{63}
\newcommand{\TAAltConflB}{44}

\newcommand{\TAAltEquivMedA}{0.000296}
\newcommand{\TAAltEquivMedB}{0.000198}

\newcommand{\TAAltRhoConsA}{-0.807}
\newcommand{\TAAltRhoConsB}{-0.685}

\newcommand{\TAAltRhoGXA}{0.939}
\newcommand{\TAAltRhoGXB}{0.868}
\newcommand{\TAAltSignA}{54}
\newcommand{\TAAltSignB}{29}
\newcommand{\TAAltSignRawA}{56}
\newcommand{\TAAltSignRawB}{50}
\newcommand{\TAAltTransitionsA}{64}
\newcommand{\TAAltTransitionsB}{54}

\newcommand{\TAExpFiftyThreeDetBelow}{1}
\newcommand{\TAExpFiftyThreeDetExcess}{+0.016}
\newcommand{\TAExpFiftyThreeDetShare}{0.999}
\newcommand{\TAExpFiftyThreeGapMax}{0.107}
\newcommand{\TAExpFiftyThreeLearnedBelow}{51}
\newcommand{\TAExpFiftyThreeLearnedExcess}{-0.044}
\newcommand{\TAExpFiftyThreeLearnedShare}{0.940}
\newcommand{\TAExpFiftyThreeTotal}{54}
\newcommand{\TAExpFiftyTwoDeltaA}{-0.037}
\newcommand{\TAExpFiftyTwoDeltaB}{-0.056}
\newcommand{\TAExpFiftyTwoFits}{64}
\newcommand{\TAExpFiftyTwoGainRelA}{0.64}
\newcommand{\TAExpFiftyTwoGainRelB}{2.96}
\newcommand{\TAExpFiftyTwoIntervalA}{[-0.043, -0.032]}
\newcommand{\TAExpFiftyTwoIntervalB}{[-0.060, -0.053]}
\newcommand{\TAExpFiftyTwoLevelHighA}{4.90}
\newcommand{\TAExpFiftyTwoLevelHighB}{5.76}
\newcommand{\TAExpFiftyTwoLevelLowA}{2.65}
\newcommand{\TAExpFiftyTwoLevelLowB}{4.99}

\newcommand{\TAExpFiftyTwoNA}{16}

\newcommand{\TAExpFiftyTwoOccGapA}{0.0002}
\newcommand{\TAExpFiftyTwoOccGapB}{0.0374}
\newcommand{\TAPanelDraws}{8}
\newcommand{\TAPanelRhoFrozen}{-0.538}
\newcommand{\TAPanelRhoMax}{-0.518}
\newcommand{\TAPanelRhoMean}{-0.550}
\newcommand{\TAPanelRhoMin}{-0.583}
\newcommand{\TAPanelRhoSd}{0.022}
\newcommand{\TAReplayCorrectPointwiseArHigh}{-0.0001}

\newcommand{\TAReplayCorrectPointwiseHeavyHigh}{-0.0000}

\newcommand{\TAReplayCorrectTemporalArHigh}{0.0001}

\newcommand{\TAReplayCorrectTemporalHeavyHigh}{0.0000}

\newcommand{\TAReplayWrongPointwiseArHigh}{-0.0001}

\newcommand{\TAReplayWrongPointwiseHeavyHigh}{0.0007}

\newcommand{\TAReplayWrongTemporalArHigh}{-0.0015}

\newcommand{\TAReplayWrongTemporalHeavyHigh}{-0.0002}

\newcommand{\TATempAepsLocalHigh}{0.342}
\newcommand{\TATempAepsLocalLow}{0.877}
\newcommand{\TATempAepsWholeHigh}{0.099}
\newcommand{\TATempAepsWholeLow}{0.839}
\newcommand{\TATempBlocks}{16}
\newcommand{\TATempDeltaGsHigh}{+0.0138}
\newcommand{\TATempDeltaGsHighInterval}{[+0.0125, +0.0152]}
\newcommand{\TATempDeltaGsLow}{+0.0029}
\newcommand{\TATempDeltaGsLowInterval}{[+0.0025, +0.0034]}

\newcommand{\TATempFinalFits}{192}
\newcommand{\TATempFinalSnapshot}{127}

\newcommand{\TATempFits}{192}

\newcommand{\Gx}{G_{x}}
\newcommand{\Gs}{G_{s}}
\newcommand{\Geps}{G_{\varepsilon}}
\newcommand{\GU}{G_{U}}
\newcommand{\Aeps}{A_{\varepsilon}}
\newcommand{\Aref}{A_{\varepsilon}^{\mathrm{ref}}}
\newcommand{\Eeps}{E_{\varepsilon}}
\newcommand{\Dx}{D_{x}}
\newcommand{\Ds}{D_{s}}
\newcommand{\Deps}{D_{\varepsilon}}
\newcommand{\sref}{\mathrm{ref}}
\newcommand{\TA}{TokAudit}
\newcommand{\Rocc}{R_{\mathrm{occ}}}

\title{TokAudit: Auditing What Extra Capacity Buys in Time-Series Tokenizers}

\author{Haofeng Guo, De-hua Chend}

\begin{document}
\begin{preprintabstract}
What improves when additional tokenizer capacity reduces reconstruction error? We introduce \TA{}, a framework for measuring what extra capacity buys in trained time-series tokenizers. Given a known or declared state and a conditional observation-noise model, \TA{} separates marginal reconstruction gains into a conditional-mean state component and a noise-related component through noise replay. It reports this split alongside realized occupancy, an occupancy-matched Gaussian reference and independently measured downstream utility. Training contrasts at approximately matched reconstruction and occupancy reveal reproducible composition differences. Across capacity levels, output variance can increase enough to worsen single-draw state error even as the conditional-mean response approaches the state. While gains are predominantly noise-related in the tested pointwise codecs, a window-compression bottleneck shifts to state-dominated gains at higher SNR. The window reference agrees with the observed direction of change. \TA{} makes marginal gain composition measurable, allowing capacity increases to be compared by what their reconstruction improvements preserve.
\end{preprintabstract}
\maketitle

\begin{figure*}[t]
  \centering
  \includegraphics[width=\textwidth]{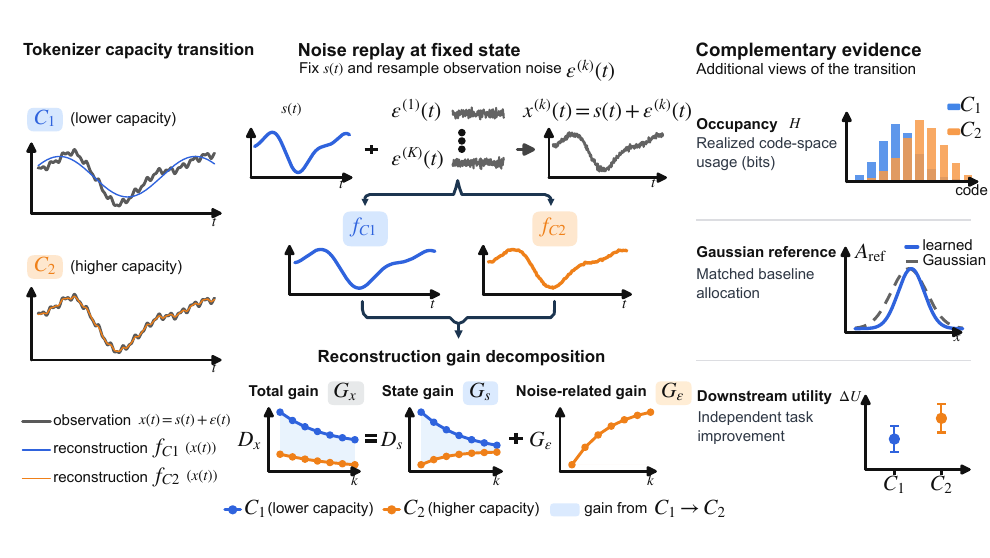}
  \caption{TokAudit measures reconstruction-gain composition across capacities. Left: a capacity increase within one tokenizer family. Centre: the gain decomposes exactly into conditional-mean state and noise-related components, estimated through noise replay at fixed states. Right: realized occupancy, an occupancy-matched Gaussian reference and independently measured downstream utility.}
  \label{fig:audit}
\end{figure*}

\section{Introduction}
\label{sec:intro}
In many time-series modeling pipelines, tokenizer capacity is fixed before
downstream model training and is therefore costly to revisit. Capacity---how
many discrete levels the representation may use---must balance two risks: too
little can discard useful structure, while too much can preserve
observation-specific detail that does not recur. Common diagnostics such as
reconstruction error, vocabulary utilization, token occupancy, and downstream
performance describe different aspects of this trade-off, but they do not
reveal what an improvement in reconstruction is made of.

A lower reconstruction error can arise because a tokenizer recovers more of
the latent state, or because it reproduces more of the observation noise.
Utilization and occupancy indicate how much of the available code space is
used, but not what the resulting representation preserves. Improved state
fidelity need not translate into greater utility for a particular downstream
task. How much capacity is added, what the resulting reconstruction gain
preserves, and how downstream utility changes are distinct quantities.

We introduce \TA{}, a measurement framework that takes the \emph{marginal
tokenizer capacity transition} as its unit of analysis and makes the
\emph{marginal reconstruction gain} from a capacity increase an explicit
object of measurement. For one tokenizer family
$T$ trained at capacities $C_1 < C_2$, it holds the state fixed and resamples
observation noise to decompose the marginal reconstruction gain into state- and
noise-related components, compares their allocation with a same-occupancy
second-moment-matched Gaussian reference, and measures downstream utility
independently (Figure~\ref{fig:audit}).

We ask: holding the state fixed and varying observation noise, does additional capacity bring the mean reconstruction response closer to the state? This differs from single-draw reconstruction quality and downstream utility. Reconstruction gain already ranks state gain closely in the studied populations; the audit adds absolute composition measurements and discrimination when reconstruction and occupancy are approximately matched.

Our contributions follow three steps. \textbf{Define composition:} decompose a capacity transition's reconstruction gain into conditional-mean state and noise-related components, alongside nominal capacity, realized occupancy, a reference and independent utility. \textbf{Distinguish differences:} measure reproducible composition changes in constructed contrasts with approximately matched reconstruction and occupancy. \textbf{Explain disagreement:} account for opposing movements of single-draw error and conditional-mean response, and compare composition across working conditions through a window-compression case.

\section{Related Work}
\label{sec:related}

\paragraph{Discrete tokenization for time series.}
Our population spans three representative quantizer mechanisms:
learned-codebook vector quantization~\citep{oord2017vqvae}, finite scalar
quantization~\citep{mentzer2024fsq} and binary spherical
quantization~\citep{zhao2025bsq}.  The third is instantiated as residual
binary quantization (RBQ), a two-stage residual construction with
a binary-spherical quantizer at each residual stage.  Time-series foundation models tokenize by
scalar binning, task-independent VQ-VAE tokens, localized frequency
representations, hierarchical discrete targets, or byte-pair
encoding~\citep{ansari2024chronos,talukder2024totem,masserano2025wavetoken,feng2025hdt,gotz2026bpe}.
We study capacity transitions in trained tokenizers across these mechanisms through a common
measurement framework.

\paragraph{What is measured about a tokenizer.}
The closest empirical studies are \citet{roger2025small}, who examine scaling,
quantization, vocabulary size and pretrained versus random initialization,
analyze ideal-predictor discretization error and relate token utilization to
forecasting performance, and \citet{gotz2026bpe}, who compare vocabulary sizes
within a forecasting pipeline.  Both relate tokenizer configurations to
downstream performance.  \TA{} instead takes the capacity transition between
two trained configurations of the same tokenizer as its unit of analysis.

\paragraph{Rate--distortion, predictive information, and benchmarks.}
Predictive rate--distortion asks how compression trades off against predictive
information~\citep{hahn2019predictive}, and the information bottleneck frames a
related trade-off as a constraint~\citep{tishby2000information}.  Remote-source,
or indirect, rate--distortion asks how well a compressed noisy observation can
represent a hidden source~\citep{wolf1970transmission,kipnis2015indirect}.
\TA{} addresses a complementary empirical question: once a codec has been trained
to reconstruct the observation, where did the reconstruction gain from additional
capacity actually go?  Forecasting benchmarks standardize datasets, splits,
metrics and evaluation protocols~\citep{godahewa2021monash,informer2021,autoformer2021};
\TA{} standardizes the quantities used to characterize reconstruction gains across capacities.

\section{\TA: Measurement Specification}
\label{sec:spec}

\subsection{The audit transition}

\TA{} takes the \emph{audit transition} as its unit of analysis: a tuple
$\tau = (T, C_1, C_2, \mathcal{E})$ comprising a tokenizer mechanism $T$, two
capacities $C_1 < C_2$ at which $T$ has been trained using the same recipe, and
an environment $\mathcal{E}$.  The capacity index is the nominal
per-stage code width in bits.  The ledger records the nominal step and the
realized occupancy increment separately, since the two need not coincide.
Training and snapshot-selection details are given in \secref{app:core}.

\subsection{Conditions for measurement}

\TA{} measures each transition under an observation model
$x_t = s_t + \varepsilon_t$. Holding $s_t$ fixed and replaying noise from its
conditional distribution makes the noise-averaged reconstruction measurable.
The core studies use a known latent state and conditionally zero-mean Gaussian
noise; the antithetic estimator additionally requires symmetric noise and
independent replay pairs. A declared state and noise model define a
model-relative measurement; interpreting it as fidelity to a physical latent
state requires the state and replay law to match that environment
(\secref{app:real}). We instantiate the core environment with a
regime-switching vector autoregression across four settings spanning
predictability and observation-noise level. Generator, regime and panel
specifications are in Appendices~\ref{app:regimes}--\ref{app:core}.

\subsection{The ledger}

The ledger records three aspects of a transition: the reconstruction gain and
its state--noise decomposition; the fraction of that gain assigned to noise
relative to the matched reference; and downstream utility.

For a transition $\tau$, the ledger is
\begin{equation}
\mathcal{L}(\tau) = \bigl(\Delta R_{\mathrm{nom}}, \Delta H, \Gx, \Gs, \Geps,
\Aeps, \Aref, \Eeps, \GU \bigr).
\label{eq:ledger}
\end{equation}
$\Delta R_{\mathrm{nom}}$ is the nominal per-token step in bits, twice the
per-stage width increment for a two-stage codec, and $\Delta H$ is the realized
joint-token occupancy increment in the same unit.  Write $f_C$ for the complete
trained codec at capacity $C$ --- encoder, quantizers and decoder --- and let
all squared-error distortions be reported as per-channel means in the
environment's standardized coordinates.  Then
\begin{equation}
\begin{aligned}
\Dx(C) &= \mathbb{E}\,\lVert s + \varepsilon - f_C(s+\varepsilon) \rVert^2,
\\
\Ds(C) &= \mathbb{E}\,\lVert s - \mathbb{E}_\varepsilon f_C(s+\varepsilon) \rVert^2,
\end{aligned}
\label{eq:dxds}
\end{equation}
with $\Deps(C) = \Dx(C) - \Ds(C)$.  The transition gains are
$\Gx = \Dx(C_1)-\Dx(C_2)$, and likewise for $\Gs$ and $\Geps$, so
$\Gx = \Gs + \Geps$ exactly.

$\Ds$ is the conditional-mean state distortion --- the error of the codec's
noise-averaged reconstruction rather than of any single reconstruction.  A
vanishing conditional-mean state distortion is therefore not a denoised reconstruction: the identity
codec has $\Ds=\Deps=\Dx=0$ while reproducing the observation noise exactly.
$\Gs$ measures a fall in conditional-mean state distortion, not a gain in usable
state information.  $\Ds$ and $\Deps$ are estimated with a bias-corrected
finite-replica estimator over resampled observation noise, with corrected
estimates retained rather than clipped (\secref{app:controls}).

The marginal noise share is $\Aeps = \Geps/\Gx$.  It is a ratio of two
reconstruction gains, not a share of $\Delta R_{\mathrm{nom}}$ or $\Delta H$.
It is defined where $\Gx > 0$ and is not clipped to $[0,1]$: $\Aeps>1$ exactly
when $\Gs<0$, a measurement outcome rather than an error.

To interpret $\Aeps$, the ledger places it against a direct
observation-reconstruction second-moment-matched Gaussian reference at the same
realized occupancies and computes the same transition-level noise gain share
under that reference.  Write $H(C)$ for the codec's realized joint-token
occupancy at nominal capacity $C$; the reference is evaluated at $H(C)$, not at
$C$:
\begin{equation}
\begin{aligned}
\Aref(\tau) &=
\frac{D^{\sref}_{\varepsilon}(H(C_1)) - D^{\sref}_{\varepsilon}(H(C_2))}
     {D^{\sref}_{x}(H(C_1)) - D^{\sref}_{x}(H(C_2))},
\\
\Eeps &= \Aeps - \Aref .
\end{aligned}
\label{eq:refshare}
\end{equation}
$\Aref$ (Eq.~\eqref{eq:refshare}) reports what fraction of its reconstruction
gain belongs to the noise-related component of the Gaussian reference over the same
two occupancies.  It is an interpretive baseline, not an optimal codec, and
``below the reference'' is an observation about this population rather than a
distance to an optimum (Appendix~\ref{app:reference}).
State-information capture uses a distinct indirect, remote-source reference
(\secref{app:closure}).  Appendix~\ref{app:ledger} summarizes
the ledger quantities and their interpretation.

\subsection{Single-draw error and output variance}
\label{sec:simple}
For a fixed state, let $M(s)=\mathbb E_\varepsilon\|s-f(s+\varepsilon)\|^2$ and $V(s)=d^{-1}\operatorname{tr}\operatorname{Cov}_\varepsilon[f(s+\varepsilon)\mid s]$, where the squared norm is a mean over $d$ channels. Averaging over states gives $M=\mathbb E_s M(s)$ and $V=\mathbb E_s V(s)$. Then
\begin{equation}M=D_s+V,\qquad G_M=G_s+G_V.\label{eq:mv}\end{equation}
All gains are low minus high capacity; $G_V$ is the reduction in output variance. $D_s$ asks whether the conditional-mean response approaches the state, whereas $M$ measures single-draw state error. Downstream models see tokens of single observations, not noise-averaged reconstructions. The combination $M-V$ is an equivalent implementation of the state component. Independent dual views estimate $V$; their finite-sample estimates need not satisfy the identity exactly.

\subsection{Independent downstream utility}

$\GU$ is the increment in downstream utility between the two capacities.  In the
primary analysis, utility is the per-step cross-sectional Spearman rank
correlation (RankIC) of a ridge readout that predicts the next-step state from a
four-step latent history; we report the first state channel, and
\secref{sec:generalization} varies the reader family.  Because task utility
depends on what a reader can extract, $\GU$ is a \emph{readout quantity},
reported separately from the reconstruction decomposition.  Implementation and
measurement details are given in Appendices~\ref{app:core} and~\ref{app:accounts}.

\section{Estimator Stability and Calibration}
\label{sec:validate}

Replica estimates converge with budget: \TAAgreeCoarse{} of the
\TANTransitionsA{} calibration transitions meet the operational tolerance at
\TACoarseReplicas{} replicas, \TAAgreeMid{} at \TAMidReplicas{}, and all
\TAAgreeFine{} do at \TAReplicaRungFine{} relative to the
\TAReplicaRefFine{}-replica endpoint.  The measured noise-related component recovers the
known observation-noise ordering and changes monotonically with added training
noise; Gaussian and distortion-matched controls preserve the reference-relative
ordering.  A noiseless analytic case yields a vanishing noise-related component, while a
stricter finite-replica accuracy criterion of the analytic study is not met at the finest rung:
\TAKTwoViolOwn{} of the \TAKTwoComparisons{} component comparisons deviate
from the closed-form value by more than one Monte Carlo standard error
(Appendix~\ref{app:validity}).

Two further checks assess panel sensitivity and noise-model dependence.  The prognostic interval
resamples initializations with the test panels fixed; re-evaluating the same
frozen population on \TAPanelDraws{} fresh panel draws gives
$\rho=\TAPanelRhoMean$ (sd \TAPanelRhoSd, range \TAPanelRhoMin\ to
\TAPanelRhoMax) against the fixed-panel \TAPanelRhoFrozen, so the panel draw alone
moves the association by roughly \TAPanelRhoSd{}
(\secref{app:panel}).  A separate stress test measures the effect of replaying a mismatched noise law: under a heavy-tailed
law, replaying Gaussian noise biases $\Ds$ in both codec families, while
replaying the declared law is consistent with the high-budget truth; under an
AR(1) law the pointwise codec is unchanged relative to its Gaussian marginal and
the window codec is not, so a window codec requires the joint law
(\secref{app:replay}).

% Generated by research_rebuild/revision_20260915/make_population_assets.py
% Main-text evidence map.  Every transition count is re-derived or checked
% against the frozen tree; do not hand-edit.
\begin{table*}[t]
  \centering\footnotesize
  \setlength{\tabcolsep}{3pt}
  \begin{tabularx}{\dimexpr\textwidth-2pt\relax}{@{}>{\raggedright\arraybackslash}X>{\raggedright\arraybackslash}Xrl@{}}
    \toprule
    study design & analysis & trans. & details \\
    \midrule
    Calibration & measurement validation and core allocation & 18 & App.~\ref{app:populations} \\
    Capacity sweep & allocation characterization & 54 & App.~\ref{app:allocation} \\
    Prognostic & allocation--utility endpoint & 32 & App.~\ref{app:prognostic} \\
    Cross-generator & dynamics robustness & 24 & App.~\ref{app:crossgenerator} \\
    Frozen re-read & reader robustness & 56 & App.~\ref{app:readers} \\
    Operational & observational-data construction & 4 & App.~\ref{app:real} \\
    External & real-data stress test & 9 & App.~\ref{app:saugeen} \\
    Multi-site & multi-site external endpoint & 72 & App.~\ref{app:multisite} \\
    Initial training-objective contrast & allocation discrimination & 2 & App.~\ref{app:discrimination} \\
    Independent training-objective confirmation & fresh-data confirmation and mechanism extension & 4 & App.~\ref{app:exp52} \\
Window-compression comparison & input range, capacity and SNR & 8 & App.~\ref{app:temporal} \\
\bottomrule
  \end{tabularx}
  \caption{Study designs and audit-transition counts. Full grids and replication: Table~\ref{tab:population-detail}.}
  \label{tab:populations}
\end{table*}

\section{Measuring Gain Composition: Discrimination and Explanation}\label{sec:character}
We first compare similar total gains, then explain disagreement between error measures, and finally compare composition across working conditions. Tables~\ref{tab:populations} and~\ref{tab:population-detail} identify the study designs and their populations.

\subsection{Composition at approximately matched reconstruction and occupancy}
We begin with the initial training-objective contrast on generator g1, SNR1.0 and the $4\to7$ capacity step. Plain and noise-view-consistency training share the base recipe; strength selection uses practical reconstruction/occupancy bands. The contrast introduced the constructed case; the independent training-objective confirmation (Appendix~\ref{app:exp52}) confirms RBQ (A) with fresh data and initializations and extends it to VQ-VAE (B), with 16 independent training blocks per unit.

Table~\ref{tab:matched-evidence} shows that both mechanisms have lower noise-related gain shares under consistency-regularized training while satisfying the practical reconstruction and occupancy matching bands; validation and test gaps are reported in full in Table~\ref{tab:new-matching}.
\begin{table*}[t]
\centering\small
\setlength{\tabcolsep}{4pt}
\begin{tabular}{@{}llllll@{}}
\toprule
Mechanism & $G_x$ gap (\%) & $\Delta H$ gap (bit) & Max. $D_x$ gap (\%) & $\delta A_\varepsilon$ & Interval \\
\midrule
RBQ & 0.61 & 0.000136 & 4.86 & -0.037 & [-0.043, -0.032] \\
VQ-VAE & 3.09 & 0.037317 & 5.76 & -0.056 & [-0.060, -0.053] \\
\bottomrule
\end{tabular}
\caption{Composition differences at approximately matched reconstruction and occupancy. Gaps are measured on test panels; the maximum $D_x$ gap is taken over the two capacities. $\delta A_\varepsilon$ denotes consistent minus plain. Matching criteria and study-specific interval methods are detailed in Appendix~\ref{app:exp52}.}
\label{tab:matched-evidence}
\end{table*}

Post-hoc 128-pair component measurements (Appendix Table~\ref{tab:new-readouts}) give positive $\delta G_M$ and $\delta G_s$ in both units; $\delta G_V$ is positive in A and unresolved in B, where $\delta$ denotes consistent minus plain. Simple variance measurements therefore also detect part of the intervention: discrimination is relative to reconstruction and occupancy, not all diagnostics. Replay budgets and interval definitions are in \secref{app:exp52}.

\subsection{Single-draw and conditional-mean errors can disagree}
\label{sec:simple-evidence}
Among \TAAltTransitionsA{} repetition-level transitions in the independent training-objective confirmation and \TAAltTransitionsB{} aggregated transitions in the capacity sweep, $G_M$ and $G_s$ disagree in sign on \TAAltSignRawA{} and \TAAltSignRawB{} transitions; both bootstrap intervals exclude zero on \TAAltSignA{} and \TAAltSignB{} of these (panels within a training repetition for the confirmation; paired initialization and panel resampling for the sweep; \secref{app:simple}). Every mismatch has $G_M<0$ and $G_s>0$: output variance can increase enough to worsen single-draw state error while the conditional mean approaches the state. Consistency improvement $G_V$ correlates with $G_s$ at \TAAltRhoConsA{} and \TAAltRhoConsB{}, whereas the $G_x$--$G_s$ correlations are \TAAltRhoGXA{} and \TAAltRhoGXB{}.
The $G_x$--$G_s$ rank agreement shows that total and state gains order similarly.
\begin{figure*}[t]\centering\includegraphics[width=\textwidth]{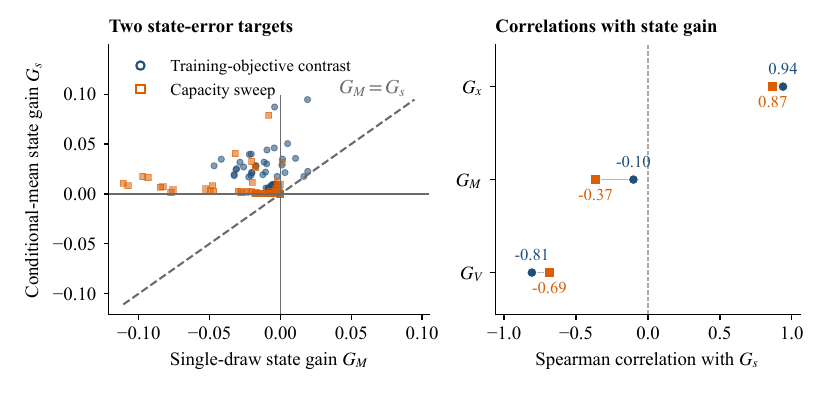}
\caption{State-error gains under two study designs. Left: $G_M$ versus $G_s$; circles represent individual training repetitions of the training-objective contrast, and squares represent initialization-averaged capacity-sweep transitions. Open and filled marks indicate $G_V\ge0$ and $G_V<0$, respectively. Right: within-design average-rank Spearman correlations with $G_s$. Positive gains denote error or variance reductions. Design and aggregation: Appendix~\ref{app:simple}.}\label{fig:simple}\end{figure*}

\subsection{A window-compression boundary and its interpretation}
\label{sec:temporal}
In a simplified construction on generator g1, a single discrete token compresses a 16-step, four-channel window at widths $4/7/10$. Two SNRs, two input views and \TATempBlocks{} independent training blocks give \TATempFits{} fits. The whole encoder reads all steps; the local encoder reads only the last four but reconstructs the entire window. For the first capacity step, median block noise shares are \TATempAepsWholeLow{}/\TATempAepsLocalLow{} at low SNR and \TATempAepsWholeHigh{}/\TATempAepsLocalHigh{} at high SNR. Dominance thus depends on the construction and conditions, rather than being a fixed property of scaling.

The post-hoc position decomposition locates the whole encoder's larger state gain mainly in the first twelve targets, which the local encoder cannot see; it diagnoses target visibility, not a causal effect of temporal modeling. On the last four steps, its advantage is negative at low SNR and unresolved at high SNR. The whole encoder also has larger reconstruction gains and realized rates. Doubling the budget on a subset fixed by block ID preserves the dominance boundary.

\begin{figure*}[t]\centering\includegraphics[width=\textwidth]{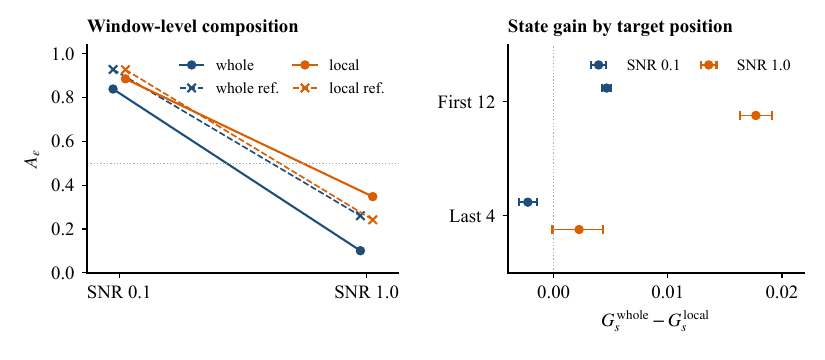}
\caption{Left: measured and window-reference noise shares (ratios of mean gains). Right: whole-minus-local state gains for targets outside (first 12) and inside (last 4) the local encoder's input window; bars are approximate 95\% paired block-bootstrap intervals over 16 blocks.}\label{fig:temporal}\end{figure*}

A post-hoc 64-dimensional Gaussian window reference matches occupancy in bits per window. Comparing ratios of mean gains on both sides, all four cells agree on which side of 0.5 the noise share falls, and both designs agree on the direction of change across SNRs (Figure~\ref{fig:temporal}). The reference captures the dominance boundary rather than fitting every cell: at high SNR, whole lies below its reference and local above it (Table~\ref{tab:window-reference}).
Block medians and the ratios of means used with the reference are labeled separately. Full position-wise components and the budget sensitivity are in \secref{app:temporal}.

\section{Populations, Reference and Independent Utility}\label{sec:generalization}

\subsection{Noise dominance in the pointwise populations}

In the calibration population, the median noise-related share of reconstruction gain exceeds ninety-five per cent ($\Aeps=\TACVFourMedian{}$).  Noise
dominance strengthens with observation noise and remains in nearly every
low-noise transition (Table~\ref{tab:regime-summary}).

\subsection{Learned codecs are noise-dominated and usually below the reference}

Noise dominance is an absolute statement about composition; comparing
$\Aeps$ with its matched reference gives a relative statement about allocation.
The learned codecs are noise-dominated, but less so than the matched reference:
all \TATotalA{} calibration transitions and \TABelowRefB{} of \TATotalB{}
capacity-sweep transitions lie below it (Figure~\ref{fig:character};
Table~\ref{tab:full-ledger}).  A deterministic uniform quantizer, matched to learned-codec occupancy
on validation data, lies below the same reference in only
\TAExpFiftyThreeDetBelow{} of \TAExpFiftyThreeTotal{} transitions while the
learned codecs do so in \TAExpFiftyThreeLearnedBelow{}; determinism alone does
not produce the pattern (\secref{app:exp53}).

\subsection{Occupancy and utility change differently across mechanisms}

The three quantizers differ in occupancy and utility gains.  Over their shared capacity
range, VQ-VAE adds the least realized occupancy
(\TAOccSpanVqvaeLo--\TAOccSpanVqvaeHi{} bits against
\TAOccSpanBsqLo--\TAOccSpanBsqHi{} for RBQ and \TAOccSpanFsqLo--\TAOccSpanFsqHi{}
for FSQ) and has the smallest utility increment in the two high-predictability
regimes, where that increment is negative; \secref{app:mechanism} prints the
regime-by-regime contrast.
A smaller occupancy increment does not imply a larger utility gain.
Targeted controls weaken several implementation-specific explanations for these
differences (\secref{app:mechanism}).

\subsection{The negative allocation--utility relation is population-dependent}
\label{sec:prognostic}

In a disjoint population with fresh fits, more reference-relative noise
allocation is associated with smaller downstream utility gains
(Spearman $\rho=\TARhoPoint$, initialization-bootstrap interval \TARhoIntervalCorrected{} (fixed transition population); Figure~\ref{fig:association}).  A post-hoc conditioning on
capacity position and realized occupancy leaves the association essentially
unchanged in this population (\secref{app:conditioning}).

\subsection{The relation is a within-stratum pattern}

Across nonlinear and seasonal dynamics the allocation--utility relation remains
negative within strata, and every transition lies below its matched reference;
the same within-stratum relation persists under nonlinear readers, and the readers agree on the ordering of utility gains across capacity transitions
(\secref{app:crossgenerator} and \secref{app:readers}).  The pooled
cross-generator association disappears after conditioning on capacity position
and realized occupancy. The cross-generator evidence therefore supports a within-stratum negative
association, rather than a pooled prognostic relation surviving capacity conditioning
(\secref{app:conditioning}).

\section{External Stress Tests on Real Dynamics}
\label{sec:externalstress}
\label{sec:external}

On the Beijing multi-site air-quality series, we inject known Gaussian
observation noise while retaining the observed multivariate dynamics, with
twelve stations as the cross-section and train-only global normalization.  The
stratified endpoint reproduces the negative association
($Q=\TAExpFortySevenQ$; 95\% interval $\TAExpFortySevenInterval$).  This
extends the result from synthetic generators to real multivariate dynamics
under identifiable injected noise; it does not identify natural noise (\secref{app:multisite}).

A separate stress test injects known corruption into the Monash Saugeen series;
its endpoint is inconclusive, and differences in training budget
and provenance prevent a like-for-like replication claim
(\secref{app:saugeen}).

\section{Discussion and Conclusion}\label{sec:discussion}
\TA{} moves from how much reconstruction improves to what that improvement consists of. Training contrasts with similar total gains can have different compositions; single-draw error and conditional-mean response can move in opposite directions; and the same measurement describes pointwise and window-compression codecs across conditions.

The window reference reproduces the observed dominance directions using occupancy and second moments. This agreement helps interpret the boundary without isolating its cause. Position-wise components locate state gains in particular targets, helping interpret which input information a design difference reflects.

\label{sec:selection}Capacity selection is a separate question. In the tested candidate set, $G_x$ and $G_s$ favor the same design, so the audit does not add selection value there (\secref{app:selection}). Identifying natural states and noise remains an application challenge; known-noise injection on real dynamics, operationally declared states and controlled generators are reported as distinct settings.
\label{sec:conclusion}
By placing gain composition alongside occupancy, a reference and independently measured utility, \TA{} enables researchers to compare what capacity increases preserve and explain disagreements among conventional readings.

\FloatBarrier
\label{main_text_end}

\onecolumn
\setlength{\multicolsep}{0pt}
\begin{multicols}{2}
\section*{AI Use Statement}

Generative AI tools were used extensively in this work.  We used them to assist
with research methodology and experimental design, implementation and validation
of experimental pipelines, analysis and interpretation of experimental results,
literature search, and manuscript drafting and revision.  In particular,
substantial parts of the experiment authoring, analysis, and reporting tooling
were developed with AI assistance.  Generative AI was additionally used to
locate relevant literature and to revise the presentation of the manuscript.
The authors take responsibility for the final content of this work, including
all AI-assisted text, code, analyses, and artifacts.

\bibliography{paper}
\bibliographystyle{plainnat}
\end{multicols}

\newgeometry{left=0.95in,right=0.95in,top=0.75in,bottom=0.78in}
\onecolumn
\appendix
\raggedbottom
\section*{Notation at a glance}
\label{app:notation}

\begin{center}\small
\begin{tabular}{@{}ll@{}}
  \toprule
  symbol & meaning \\
  \midrule
  $\tau=(T,C_1,C_2,\mathcal{E})$ & audit transition \\
  $x_t=s_t+\varepsilon_t$ & observation, latent state, observation noise \\
  $f_C$ & trained codec at capacity $C$ \\
  $H(C)$ & realized joint-token occupancy \\
  $D_x,D_s,D_\varepsilon$ & observation, state and noise-related distortions \\
  $G_x,G_s,G_\varepsilon$ & corresponding marginal reconstruction gains \\
  $A_\varepsilon$ & measured marginal noise share $G_\varepsilon/G_x$ \\
  $A_\varepsilon^{\mathrm{ref}}$ & same-occupancy matched-reference noise share \\
  $E_\varepsilon$ & $A_\varepsilon-A_\varepsilon^{\mathrm{ref}}$ \\
  $G_U$ & downstream utility gain \\
  $\rho,\tau_b$ & Spearman and Kendall rank correlations (the latter distinct from the transition $\tau$) \\
  $Q,Q_g$ & study-specific stratified association endpoints \\
  $m,d$ & number of antithetic pairs and channel dimension \\
  \bottomrule
\end{tabular}
\end{center}

\noindent Table~\ref{tab:ledger} gives the fuller interpretation of the ledger
entries; local symbols are defined where they are used.

\section*{Guide to the appendix}
The appendix follows the measurement workflow, then presents the supporting
studies.  Study status is collected at the end; conditions
needed to interpret a result remain with that result.

\begin{center}\small
\begin{tabularx}{\textwidth}{@{}>{\raggedright\arraybackslash}X>{\raggedright\arraybackslash}X@{}}
\toprule
main-text topic & supporting material \\
\midrule
Experimental settings and training & Sections~\ref{app:regimes}--\ref{app:prognostic}; Table~\ref{tab:population-detail} \\
Measurement and uncertainty & Sections~\ref{app:controls}--\ref{app:aggregation} \\
Estimator validation & Section~\ref{app:validity} \\
Noise allocation and mechanism differences & Tables~\ref{tab:full-ledger}--\ref{tab:regime-summary}; Section~\ref{app:mechanism} \\
Conditioning, dynamics and readers & Sections~\ref{app:conditioning}--\ref{app:readers} \\
Tokenizer and matched-training controls & Sections~\ref{app:codec-controls}--\ref{app:discrimination} \\
Simple statistics and the state component & Section~\ref{app:simple} \\
Temporal bottleneck, selection, panel and replay stress & Sections~\ref{app:temporal}--\ref{app:replay} \\
Independent confirmation and reference sensitivity & Sections~\ref{app:exp52} and~\ref{app:exp53} \\
Utility analyses and selection & Sections~\ref{app:interface}--\ref{app:rules} \\
External studies & Sections~\ref{app:multisite}--\ref{app:saugeen} \\
Reproduction & Section~\ref{app:repro} \\
\bottomrule
\end{tabularx}
\end{center}

\FloatBarrier
\section{Experimental setup and populations}
\subsection{Regime definitions}
\label{app:regimes}

The core generator is a regime-switching vector autoregression with additive
Gaussian observation noise, a latent state known by construction and
resampleable noise:
\begin{align}
s_{i,t} &= s_{i,t-1} A_{r(t)}^{\top} + \lambda\, \ell_i f_t + \xi_{i,t},
  & \xi_{i,t} &\sim \mathcal{N}(0, \sigma_\xi^2 I), \label{eq:state}\\
f_t &= \rho_f f_{t-1} + \TADGPFactorNoise{}\,\eta_t, & \eta_t &\sim \mathcal{N}(0,1),
  \label{eq:factor}\\
x_{i,t} &= s_{i,t} + \varepsilon_{i,t}, & \varepsilon_{i,t} &\sim \mathcal{N}(0, \sigma_\varepsilon^2 I),
  \label{eq:obs}
\end{align}
with $s_{i,t}, x_{i,t}, \varepsilon_{i,t}, \xi_{i,t} \in
\mathbb{R}^{\TANSignalDim{}}$ and $\ell_i \in \mathbb{R}^{\TANSignalDim{}}$.
The index $i$ runs over entities within a panel; all vectors are row vectors.
Each panel has its own scalar factor path $f_t$, shared across that panel's
entities.  The regime
index is $r(t) = \lfloor t/\TADGPRegimePeriod{} \rfloor \bmod 2$, and the two
transition matrices $A_0$ and $A_1$ are diagonal, each scaled by that setting's
state scale $\varsigma$.  Observation noise is scaled so that the raw noise
energy equals the signal energy divided by the target signal-to-noise ratio, so
predictability and observation-noise level vary independently through
$(\varsigma, \rho_f)$ and the target SNR.

The four regimes complete a predictability-by-observation-noise square.  The
generator's state scale $\varsigma$ and factor persistence $\rho_f$ set
predictability, and the target SNR sets the observation-noise level.
Table~\ref{tab:regimes} reports the realized noise fraction alongside the nominal
one, because the two differ.  Each regime trains on \TAPanelsTrain{} panels.  The internal tag
\texttt{lowSNR} denotes high observation noise: at $\mathrm{SNR}=0.1$,
roughly ninety per cent of the expected observation energy comes from noise.  The table uses
observation-noise levels throughout.

\begin{table}[ht]
  \centering\small
  \begin{tabular}{@{}lcccc@{}}
    \toprule
    Regime & $\varsigma$ & $\rho_f$ & target SNR & realized $f_\varepsilon$ \\
    \midrule
    low predictability, high obs.\ noise & \TARegScaleLow & \TARegRhoLow & \TARegSNRLow & \TAFepsLowA \\
    high predictability, high obs.\ noise & \TARegScaleHigh & \TARegRhoHigh & \TARegSNRLow & \TAFepsLowB \\
    high predictability, low obs.\ noise & \TARegScaleHigh & \TARegRhoHigh & \TARegSNRHigh & \TAFepsHigh \\
    low predictability, low obs.\ noise & \TARegScaleLow & \TARegRhoLow & \TARegSNRHigh & --- \\
    \bottomrule
  \end{tabular}
  \caption{Regime definitions. The fourth cell is used only by the prognostic population (\secref{sec:prognostic}); its realized noise fraction was not measured in the synthetic core and is shown as a dash.}
  \label{tab:regimes}
  \par\smallskip
  \begin{minipage}{\textwidth}\footnotesize\raggedright\textit{Shared generator parameters:} the
  state matrices are $\varsigma\,\mathrm{diag}(0.85,0.70,0.55,0.40)$ and
  $\varsigma\,\mathrm{diag}(0.55,0.40,0.25,0.10)$; loadings are
  $\ell_i \sim \mathrm{Unif}(0.5,1.5)^{4}$; the factor enters at
  $\lambda = \TADGPFactorLoad{}$ with innovation
  $\TADGPFactorNoise{}\,\eta_t$; the state innovation is
  $\sigma_\xi = \TADGPInnov{}$; and the burn-in is \TADGPBurnIn{} steps.
  \end{minipage}

\end{table}

\subsection{The synthetic core: codec, budget and optimizer}
\label{app:core}

The three learned quantizer families share the architecture, training budget
and readout described here.  The regime definitions are in
\secref{app:regimes}, and Table~\ref{tab:population-detail} gives the
capacity grids and replication schemes.

\paragraph{Codec.}
The environment emits one \TANSignalDim{}-dimensional observation per time step,
and the codec is applied per time step rather than to a window.  The three
mechanisms share one MLP skeleton and differ only in the two residual quantizers
between its halves:
\begin{equation}
z = E(x), \qquad z_1 = q_1(z), \qquad z_2 = q_2(z - z_1), \qquad
\hat x = D(z_1 + z_2),
\label{eq:codec}
\end{equation}
\begin{equation}
E = \operatorname{LN}\circ
    \operatorname{Lin}(\TATokHidden{}\!\to\!\TATokEmbed{})\circ
    \operatorname{GELU}\circ
    \operatorname{LN}\circ\operatorname{Lin}(4\!\to\!\TATokHidden{}),
\label{eq:encoder}
\end{equation}
\begin{equation}
D = \operatorname{Lin}(\TATokHidden{}\!\to\!4)\circ\operatorname{GELU}\circ
    \operatorname{LN}\circ
    \operatorname{Lin}(\TATokEmbed{}\!\to\!\TATokHidden{}).
\label{eq:decoder}
\end{equation}
The \TATokStages{} residual stages quantize the \TATokEmbed{}-dimensional latent
in sequence, and the decoder reconstructs from the sum of the two stage codes.  A
capacity is a per-stage width $w$ applied to both stages, so a transition's
nominal per-token step is $2\Delta w$ bits.  Every quantizer uses a
straight-through estimator.  Fitted codecs range from \TATokParamsMin{} to
\TATokParamsMax{} parameters.

\paragraph{The three quantizers.}
\emph{Residual binary quantization} (RBQ) applies the binary-spherical
quantizer of \citet{zhao2025bsq} as the stage operator of the two residual
quantizers in \eqref{eq:codec}: it normalizes the latent, projects it through
a learned bias-free hyperplane to $w$ signs, and decodes through a learned map
back to \TATokEmbed{} dimensions.  Its $2^w$ codes are implicit, and the nominal code space contains all sign patterns.  Its objective adds to the reconstruction error an entropy regularizer
on the batch-mean bit marginals, with weight \TATokBSQEntropy{}.  \emph{Codebook
VQ-VAE} gives each stage an explicit learned codebook of $2^w$ vectors in
$\mathbb{R}^{\TATokEmbed{}}$ and nearest-neighbour quantizes.  Its objective adds a
unit-weight codebook term and an encoder-side commitment term of weight
\TATokVQCommit{}.  \emph{Finite scalar quantization} projects the latent to a
small product grid through a learned linear map, rounds coordinate-wise on a
\emph{fixed} grid, and projects back.  The level factorization is fixed by width:
$\TATokFSQLevelsFour{}$ at $w=4$, $\TATokFSQLevelsSeven{}$ at $w=7$ and
$\TATokFSQLevelsTen{}$ at $w=10$, each giving exactly $2^w$ codes.  Its objective is
the reconstruction error alone.  Table~\ref{tab:core} collects the per-family
objective.  The binary implementation's stored commitment coefficient is not
applied; its fitted objective is reconstruction plus
entropy.  The VQ-VAE commitment coefficient \TATokVQCommit{} is active.

\begin{table}[ht]
  \centering\small
  \begin{tabular}{@{}lll@{}}
    \toprule
    mechanism & code structure & term added to the reconstruction error \\
    \midrule
    residual binary quantization & implicit $2^w$, signed hyperplanes
      & entropy, weight \TATokBSQEntropy{} \\
    codebook VQ-VAE & explicit learned $2^w\times\TATokEmbed{}$
      & codebook $+$ \TATokVQCommit{}\(\times\) commitment \\
    finite scalar quantization & fixed grid, level product $=2^w$ & none \\
    \bottomrule
  \end{tabular}
  \caption{Implemented quantizer objectives under a shared codec architecture and update budget.}
  \label{tab:core}
\end{table}

\paragraph{Budget, optimizer and snapshot.}
Each fit takes exactly \TATokSteps{} optimization steps at minibatch size
\TATokBatch{}, using Adam at learning rate $\TATokLR{}$ with the library's default
moments, no weight decay, the gradient norm clipped to \TATokClipNorm{}, and a
cosine schedule that anneals the rate to zero over the \TATokSteps{} steps.  The
training split is \TAPanelsTrain{} panels of \TANEntities{} entities over
\TANTimes{} usable steps, which is \TATokRows{} observation rows, so one epoch is
\TATokBatchesPerEpoch{} minibatches and the budget is exactly \TATokEpochs{}
epochs, with batches drawn by a per-epoch seeded permutation.  Every fit in the
population, for every mechanism, regime, width and initialization, shares this
budget and optimizer.  The controlled differences are mechanism and capacity.  The codec is evaluated every
\TATokSnapEvery{} steps, at \TATokSnapshots{}
checkpoints, on the held-out validation panels.  It is scored by the mean squared
error with which it reconstructs those panels' \emph{observations}; the latent
state is not used for selection.  We retain the checkpoint with the lowest target-free
validation reconstruction error; the validation split is used only for this choice.

\paragraph{Reader.}
The fitted codec's frozen latent is read by a ridge probe whose features are the
last \TAPROBEHistory{} steps of the summed \TATokEmbed{}-dimensional stage
latents, concatenated into
$h_t = [\,z_{t-3};z_{t-2};z_{t-1};z_t\,] \in \mathbb{R}^{256}$, with
$z_t = z_{1,t}+z_{2,t} \in \mathbb{R}^{64}$;
the label is the known state at the next step.  The primary score is the first
state channel's per-step cross-sectional Spearman rank correlation (the
rank-information coefficient); the four-channel average is a secondary
readout.  The penalty is chosen on the
inner training panels only (\TATrainFitPanels{} of them) from a
\TATokProbeAlphasN{}-point logarithmic grid $\TATokProbeAlphas{}$, by the mean
per-step cross-sectional rank information coefficient on those panels.  The
probe is then refit with that penalty on all training panels.

\paragraph{Population and splits.}
A cell is one (mechanism, regime, width) triple trained at \TANSeeds{}
initializations seeded \TATokSeeds{}, and a transition is a pair of widths.  Each
regime trains on \TAPanelsTrain{} panels and holds out \TAPanelsValidation{}
validation panels.  All fits are scored on \TAPanelsTest{} test panels generated
once, after the capacity choices are locked.

\subsection{The prognostic endpoint}
\label{app:prognostic}

The prognostic population is the four-regime square of Table~\ref{tab:regimes}:
RBQ and VQ-VAE at per-stage widths $\{2,4,7,10\}$ and FSQ at $\{4,7,10\}$,
\TAPopSeeds{} initializations at seeds $47$--$51$, all disjoint from every
earlier population in this project.  Training and validation panels are
generated under new seeds; the test panels are generated once under a further
new seed and are read by no earlier experiment.  No capacity is selected: every
rung enters the endpoint.  The codec skeleton, update budget, snapshot rule and
reader recipe are those of \secref{app:core}; the endpoint is the one stated in
\secref{sec:prognostic}.

The main text reports the primary endpoint and its corrected interval.  Four
descriptive secondaries were declared alongside it: Spearman correlations of
the utility gain with the state gain (\TASecondSignal{}), with the total
reconstruction gain (\TASecondRecon{}), with the occupancy gain
(\TASecondRate{}), and with the state-information efficiency at the
higher-capacity end of the transition (\TASecondEfficiency{}; the
reference-to-measured ratio of \secref{app:closure}).  All
\TAPopTransitions{} transitions lie below their reference, with excesses from
\TAFreshExcessLo{} to \TAFreshExcessHi{} and no exception in either direction.
The cross-generator study is reported separately in \secref{app:crossgenerator}.

\subsection{Population grids and replication}
\label{app:populations}
Table~\ref{tab:population-detail} records the codec counts, capacity grids and
replication schemes behind the main-text evidence map.  Rows carry the
study-design names of Table~\ref{tab:populations}: the capacity sweep is the
extended population (Exp~29 Part B); the initial training-objective contrast is
the matched-contrast study (Exp~48); the independent training-objective
confirmation is Exp~52; and the window-compression comparison is Exp~55.

% Generated by research_rebuild/revision_20260915/make_population_assets.py
% Every count and every grid/replication cell is re-derived from a frozen
% bundle.  Do not hand-edit.
\begin{table}[t]
  \centering\footnotesize
  \setlength{\tabcolsep}{3pt}
  \begin{tabularx}{\dimexpr\textwidth-\tabcolsep\relax}{@{}>{\hsize=1.25\hsize\raggedright\arraybackslash}Xrr>{\hsize=1.05\hsize\raggedright\arraybackslash}X>{\hsize=0.70\hsize\raggedright\arraybackslash}X@{}}
    \toprule
    population & codecs & trans. & capacity grid & replication \\
    \midrule
    Calibration & 135 & 18 & 3 mech., 4/7/10 bits & 3 regimes, 5 inits \\
    Capacity sweep & 315 & 54 & RBQ/VQ 2--10, FSQ 4/7/10 & 3 regimes, 5 inits \\
    Prognostic & 220 & 32 & RBQ/VQ 2/4/7/10, FSQ 4/7/10 & 4 regimes, 5 inits \\
    Cross-generator & 96 & 24 & 2 mech., 2/4/7/10 bits & 2 gen., 2 noise, 3 inits \\
    Frozen re-read & 316 & 56 & as above, ridge/MLP/GRU readers & fresh panels \\
    Operational & 30 & 4 & 2 mech., 4/7/10 bits & ETTm1, 5 inits \\
    External & 49 & 9 & VQ, $K{=}32$--$256$ & 3 corruptions $+$ clean, 3 inits \\
    Multi-site & 48 & 72 & RBQ/VQ, widths 2/4/7/10 & 3 cells (clean + 2 noise), 2 inits, seasons as strata \\
    Initial training-objective contrast & 20 & 2 & RBQ, widths 4/7 & 2 conditions, 5 paired inits \\
    Independent training-objective confirmation & 128 & 4 & 2 units, 2 arms, 4/7 & 16 blocks/unit \\
Window-compression comparison & 192 & 8 & 2 SNR, 2 views, 4/7/10 & 16 blocks \\
\bottomrule
  \end{tabularx}
  \caption{Study configurations and replication. Calibration and frozen-re-read rows reuse existing codecs.}
  \label{tab:population-detail}
\end{table}

\FloatBarrier
\section{Measurement and uncertainty}
\subsection{Estimator and its analytic controls}
\label{app:controls}

\paragraph{Estimator form.}
The estimator computes, per state, the pair means
$Y_j = \tfrac12(f(s+e_j) + f(s-e_j))$ for independent noise draws $e_j$, and
their average $\bar Y$.  Its
population identity is
\begin{equation}
\mathbb{E}_\varepsilon\!\left[\lVert s - \bar Y\rVert^2 \mid s\right]
  = \lVert s - \mu(s)\rVert^2
  + \frac{\operatorname{tr}\operatorname{Cov}(Y\mid s)}{m\,d},
\qquad \mu(s) = \mathbb{E}_\varepsilon\!\left[f(s+\varepsilon)\mid s\right],
\end{equation}
where $m$ is the number of independent pairs (the replica count is $2m$) and $d$
the channel count ($d=4$ in these studies).  Every squared norm here is a
per-channel average, $\lVert v\rVert^2 = d^{-1}\sum_{k=1}^{d}v_k^2$, so the bias
term is the channel-averaged covariance trace over $m$ pairs: with
$\operatorname{tr}$ read as the usual sum over channels,
$\operatorname{tr}\operatorname{Cov}(Y)$ carries the channel sum and must be
divided by $m\,d$.  The sample variance of the pair means with $m-1$ degrees of
freedom, averaged over channels, estimates that term, and subtracting
it gives the bias-removed state component.  The same term is added to the noise-related
component, so the two still sum to the observation-reconstruction error $\Dx$.
The estimator is therefore
\begin{equation}
\widehat b(s) = \frac{1}{m(m-1)}\sum_{j=1}^m \lVert Y_j-\bar Y\rVert^2
  = \frac{1}{m(m-1)d}\sum_{j=1}^m\sum_{k=1}^d
  (Y_{jk}-\bar Y_k)^2,
\end{equation}
The right-hand expression expands the per-channel norm explicitly.  We compute
the correction per state, then average over sampled states, retaining negative
corrected estimates.  Independent, identically distributed draws and a fixed
codec are assumptions; the estimator cannot verify them from its inputs.

\paragraph{The two distortion components.}
Write $\mu_C(s) = \mathbb{E}[f_C(x)\mid s]$ for the codec's noise-averaged
reconstruction.  Because $\mu_C$ is the conditional mean, the single-sample
state-reconstruction error splits exactly,
\begin{equation}
\mathbb{E}\lVert s - f_C(x)\rVert^2
  = \Ds(C) + \mathbb{E}\lVert f_C(x) - \mu_C(s)\rVert^2 ,
\label{eq:ds-split}
\end{equation}
where the left side takes its error against the \emph{state} rather than against
the observation, so it is neither $\Ds$ nor $\Dx$.  $\Ds$ scores the averaged
reconstruction, and the second term is the codec's own conditional spread around
it.  Under $\mathbb{E}[\varepsilon\mid s]=0$ the same split reads on the residual
$r_C = x - f_C(x)$ as
\begin{equation}
\Ds(C) = \mathbb{E}\bigl\lVert\mathbb{E}[r_C\mid s]\bigr\rVert^2, \qquad
\Deps(C) = \mathbb{E}\bigl\lVert r_C - \mathbb{E}[r_C\mid s]\bigr\rVert^2 ,
\label{eq:residual-split}
\end{equation}
a conditional mean and a conditional variance.

For the identity codec $f_C(x)=x$, all three distortion components vanish,
while its single-sample error against the state equals the observation-noise
energy.  Thus $\Gs$ measures a reduction in conditional-mean state distortion;
it is neither usable state information nor a noise mutual information, bit
rate, or count of encoded noise bits.  Equation~\eqref{eq:residual-split}
requires conditionally zero-mean noise.  The antithetic estimator additionally
requires symmetric noise, independent pairs and a fixed codec.

\paragraph{Valid gains.}
The share is defined on positive reconstruction gains (\secref{sec:spec});
denominators that are non-finite or below the
declared minimum gain in distortion units are marked invalid, and valid ratios
are not clipped to $[0,1]$.

\subsection{The ledger's entries}
\label{app:ledger}

The ledger has three layers.  The first is the reconstruction gain and its two
components, $\Gx=\Gs+\Geps$.  The second is the share of that gain assigned to
noise, $\Aeps=\Geps/\Gx$, and its reference-relative excess $\Eeps=\Aeps-\Aref$.
The third is an independently measured utility gain $\GU$.
Table~\ref{tab:ledger} defines all nine entries of Eq.~\eqref{eq:ledger},
including the nominal capacity step and realized occupancy increment.
The latter is the entropy increment of the paired two-stage token
$z = z_1V + z_2$ and determines the two rates used by the reference.

\begin{table}[ht]
  \centering\scriptsize
  \begin{tabular}{@{}lp{0.62\textwidth}@{}}
    \toprule
    entry & what it is, and how it may be read \\
    \midrule
    $\Delta R_{\mathrm{nom}},\ \Delta H$ & nominal per-token step and realized
      joint-token occupancy increment, in bits. \\
    $\Gx,\ \Gs,\ \Geps$ & reconstruction gain, split into a fall in
      conditional-mean state distortion ($\Gs$) and a fall in the noise-related
      residual component ($\Geps$); $\Gx=\Gs+\Geps$ by construction, so a zero
      residual is a consistency check, not evidence. \\
    $\Aeps$ & marginal noise share $\Geps/\Gx$, a ratio of two reconstruction
      gains and not a share of $\Delta R_{\mathrm{nom}}$ or $\Delta H$; defined
      where $\Gx>0$, above one exactly when $\Gs<0$ and below zero exactly when
      $\Geps<0$. \\
    $\Aref$ & the same-occupancy second-moment-matched reference put through the
      same gain arithmetic as the measurement, Eq.~\eqref{eq:refshare}; a
      Gaussian proxy rather than an optimal codec. \\
    $\Eeps$ & excess $\Aeps-\Aref$: both are noise gain shares over the same
      transition, the reference also includes test-channel randomness, so this is not a distance to an optimum. \\
    $\GU$ & future-utility gain, measured on its own readout and derived from
      none of the entries above. \\
    \bottomrule
  \end{tabular}
  \caption{Ledger quantities and their interpretation. Measured and reference noise shares describe reconstruction-gain composition; the reference also includes test-channel randomness.}
  \label{tab:ledger}
\end{table}

\subsection{Measurement targets and readouts}
\label{app:accounts}

Table~\ref{tab:accounts} distinguishes the measurements by target, readout, metric
and time index.  State and noise distortions measure same-time reconstruction
under the codec; utility measures one-step-ahead prediction under a specified
readout.  The two state-informed linear readouts in Table~\ref{tab:ceilings} agree to within
$10^{-4}$ in every regime, consistent with the state carrying the future
information and the observation adding none given it.  The observation-only
linear readout is a baseline, so the gap to the state-informed arms is not a recovery rate.

% Generated by research_rebuild/make_tokaudit_assets.py
% Every value in this file is read from a frozen bundle.  Do not hand-edit.
\begin{table}[ht]
  \centering\scriptsize
  \begin{tabularx}{\textwidth}{@{}>{\raggedright\arraybackslash}X>{\raggedright\arraybackslash}X>{\raggedright\arraybackslash}X>{\raggedright\arraybackslash}Xc r@{}}
    \toprule
    measure & target & read-out & loss / metric & index & in \\
    \midrule
    conditional-mean state distortion $\Ds$ & $s_t$ & codec $f_C$ & squared error, ch-avg. & $t$ & \eqref{eq:dxds} \\
    noise-related residual component $\Deps$ & residual around its conditional mean & codec $f_C$ & conditional variance, ch-avg. & $t$ & \eqref{eq:dxds} \\
    utility gain $\GU$ & $s_{t+1}$ & ridge, 4-step latent history & per-step cross-sectional rank IC (first channel) & $t+1$ & \secref{sec:spec} \\
    linear-readout baseline & $s_{t+1}$ & linear on obs.\ or state & per-step cross-sectional rank IC (first channel) & $t+1$ & \secref{app:ceilings} \\
    state-information capture & $s_t$ & ridge on the latent & state-error drop / base & $t$ & \secref{app:rules} \\
    closure probes & $s_t$, $\mathbb{E}[s_{t+1}\!\mid\! H_t]$ & ridge on the latent & $R^2$ & $t$, $t{+}1$ & \secref{app:closure} \\
    \bottomrule
  \end{tabularx}
  \caption{Measurement targets, readouts, metrics and time indices. Conditional-mean reconstruction, downstream utility and state-readout performance are distinct targets.}  \label{tab:accounts}
\end{table}

\subsection{The matched reference: construction and scope}
\label{app:reference}
\label{app:refscope}

The reference evaluates the same transition-level gain share as the measured
ledger, using a Gaussian test channel at the codec's realized occupancies.
Its noise-related component includes test-channel randomness.  It provides an
allocation baseline, not an optimal allocation of physical observation noise;
the noiseless control in \secref{app:validity} makes this distinction explicit.

\paragraph{Second moments and water-filling.}
Let the standardized observation and signal satisfy $x = s + e$ elementwise and
be flattened to $N$ rows of dimension $d$.  The reference forms the empirical
second-moment matrices
\begin{equation}
\hat\Sigma_x = \frac1N\sum_{n=1}^N x_n x_n^\top, \qquad
\hat\Sigma_s = \frac1N\sum_{n=1}^N s_n s_n^\top,
\end{equation}
symmetrizes each as $(\hat\Sigma+\hat\Sigma^\top)/2$, and defines the noise
second moment by exact subtraction, $\hat\Sigma_e = \hat\Sigma_x - \hat\Sigma_s$.
These are uncentered second moments of the training-standardized arrays: the
construction does not subtract a sample mean, so it matches the second-moment
matrix rather than the centered sample covariance, and the reference carries that
name.  Taking the signal second moment as measured and the noise as the residual
makes the identity below hold by construction and folds the finite-sample cross
term into the noise-related component.  The construction requires
$\lambda_{\min}(\hat\Sigma_e) \ge 0$.

Diagonalize $\hat\Sigma_x = U\operatorname{diag}(\lambda)U^\top$, floor the
eigenvalues at $10^{-12}$, and for a level $\theta$ define
\begin{equation}
R(\theta) = \sum_{i:\,\lambda_i>\theta} \tfrac12\log_2\frac{\lambda_i}{\theta},
\end{equation}
zero when no direction is retained.  The level saturating the requested rate is
found by bisecting \emph{geometrically} --- the trial point is
$\sqrt{\theta_{\rm lo}\theta_{\rm hi}}$, starting from
$\lambda_{\max}10^{-14}$ --- and stops when
$\theta_{\rm hi}/\theta_{\rm lo}-1 < 10^{-14}$, so the operative tolerance is a
relative one on $\theta$.

\paragraph{Shrinkage and the three distortions.}
At level $\theta$ the test channel's complement is diagonal \emph{in the
eigenbasis}: $\widetilde W = \operatorname{diag}(w)$ with
$w_i = \max(0, 1-\theta/\lambda_i)$, whose original-coordinate form is
$W = U\widetilde W U^\top$.  The traces below are therefore taken in the
eigenbasis, against the rotated second-moment matrices
$\tilde\Sigma_s = U^\top\hat\Sigma_s U$ and
$\tilde\Sigma_e = U^\top\hat\Sigma_e U$.  All factors in each trace use
this eigenbasis.  With the coding noise on retained directions
$\nu_i = (\lambda_i-\theta)\theta/\lambda_i$ and zero on the rest, the
reference's per-channel-mean distortions at the same $1/d$ normalization the
measured components use are
\begin{equation}
D_x^{\sref} = \frac1d\sum_i \min(\lambda_i,\theta), \qquad
D_s^{\sref} = \frac1d\operatorname{tr}\!\bigl[(I-\widetilde W)
              \tilde\Sigma_s(I-\widetilde W)\bigr],
\end{equation}
\begin{equation}
D_\varepsilon^{\sref} = \frac1d\Bigl[\operatorname{tr}\!\bigl[(I-\widetilde W)
      \tilde\Sigma_e(I-\widetilde W)\bigr] + \sum_i \nu_i\Bigr],
\end{equation}
so $D_x^{\sref} = D_s^{\sref} + D_\varepsilon^{\sref}$ identically.  The reference quantities carry the $\sref$
superscript throughout to keep them apart from the measured components of the
same name.

The rate handed to the reference is the model's own realized joint-token entropy
at that configuration, not a nominal codebook size, and the reference is
evaluated panel by panel over the frozen test panels.  For a transition the
reference gains and their share are
\begin{equation}
G^{\sref}_s = D^{\sref}_s(H(C_1))-D^{\sref}_s(H(C_2)), \qquad
G^{\sref}_{\varepsilon} = D^{\sref}_{\varepsilon}(H(C_1))
  -D^{\sref}_{\varepsilon}(H(C_2)),
\end{equation}
\begin{equation}
G^{\sref}_x = G^{\sref}_s + G^{\sref}_{\varepsilon}, \qquad
\Aref = G^{\sref}_{\varepsilon}/G^{\sref}_x \ \ (G^{\sref}_x>0),
\end{equation}
which is Eq.~\eqref{eq:refshare}; the measured share is formed the same way from
the bias-corrected estimates and is undefined when $G_x$ is not positive.

\paragraph{Scalar proxy for Saugeen.}
This construction is estimable only when the observation second-moment matrix
rests on enough effective samples.  On the Saugeen series of
\secref{app:saugeen} it is not: the window second-moment matrix has on the order of
\TASaugeenEffSamples{} effective samples and the measured residual has a
minimum eigenvalue of \TASaugeenMinEig{} against a mean eigenvalue of
\TASaugeenMeanEig{}.  The reference is therefore replaced there by a
per-time-point scalar variance-matched construction at the model's realized
occupancy per time point, and that replacement is declared a proxy.  The measured
components are unchanged by the substitution; only the reference share is.  This substitution is specific to the measured second moments of that series.

\subsection{Aggregation and uncertainty}
\label{app:aggregation}

\paragraph{Aggregation order and bias removal.}
Every per-configuration distortion is the mean over that configuration's
initializations of the seed's own distortion, and each seed's distortion is
already a channel average and an average over its sampled states.  A transition
gain is the difference of the two configuration means,
$\overline{D}(C_1)-\overline{D}(C_2)$; because both configurations share the same
initialization set this equals the mean over initializations of the per-seed
paired difference.  An allocation share is formed \emph{once} from those
aggregated gains --- a ratio of means,
$\Aeps=\overline{G}_{\varepsilon}/\overline{G}_{x}$ and
$\Aref=\overline{G}^{\sref}_{\varepsilon}/\overline{G}^{\sref}_{x}$, so
$\Eeps=\Aeps-\Aref$ --- and is never an average of per-seed shares.  The
reference is put through the same arithmetic, so measurement and reference use the same gain-share functional, evaluated under their respective models.

The state-distortion estimator removes the finite-pair Monte-Carlo bias in expectation under
the estimator's stated assumptions --- independent, identically distributed
symmetric pairs and a fixed codec, as stated with the estimator in
\secref{app:controls}; it is not an exact removal in every finite realization.

\paragraph{Reported intervals.}
The prognostic endpoint's interval $\TARhoIntervalCorrected$ resamples the
$\TANSeeds$ initializations with replacement, one draw shared by all transitions
inside a replicate, and recomputes the transition means, differences, ratios and
excesses before taking the Spearman correlation, so the resampled functional
matches the point estimate; $10{,}000$ replicates and $2.5/97.5$ percentiles are
used, and a replicate is dropped only if it has fewer than three usable
transitions or a constant vector.  The superseded \TARhoIntervalFrozen{} averaged
per-seed ratios instead and is quoted only where flagged.  The replica-convergence
criterion applies the operational tolerance
$\max(\TATolAbs{},\TATolRel{}\times|\Gs|)$ to the bootstrap interval of the
difference between a coarse rung's corrected state gain and the finest rung's,
as an equivalence band.  The \TAResolvedPos{} of \TANTransitionsA{}
calibration transitions counted as resolved use a family-wise adjusted
per-transition interval ($\alpha=0.05/18$ over the 18 transitions): a transition
counts only if its corrected state gain is positive and the lower end of that
interval is above zero.

\paragraph{Exact seed-support enumeration.}
The $n=\TAPatBootstrapDrawSize$ seed bootstrap behind the prognostic interval was
enumerated exactly.  Every draw of five labels from five seeds is a multiset, so
the replicate distribution is supported on $\binom{9}{5}=\TAPatBootstrapAtoms$ atoms
with multinomial probabilities $5!/(\prod_i n_i!)\,/\,5^5$.  Each atom was
evaluated once through the same functional as the corrected analysis (seed
means, gains, ratio of means, reference share, excess, Spearman over
\TAPatBootstrapTransitions{} transitions).  The identity multiset reproduces the prognostic point estimate exactly.
The \TAPatBootstrapAtoms{} atoms take \TAPatBootstrapDistinct{}
distinct correlation values.
\begin{center}\small
\begin{tabular}{@{}lr@{}}
\toprule
enumeration quantity & value \\
\midrule
identity-multiset correlation & $\TAPatBootstrapRhoPoint$ \\
maximum deviation from stored draws & $\TAPatBootstrapMaxDev$ \\
exact step interval & $[\TAPatBootstrapStepLo,\TAPatBootstrapStepHi]$ \\
smoothed linear interval & $[\TAPatBootstrapStepLo,\TAPatBootstrapLinearHi]$ \\
support & $[\TAPatBootstrapSupportMin,\TAPatBootstrapSupportMax]$ \\
identity-atom weight & $\TAPatBootstrapIdentityWeight$ \\
frequency total-variation distance & $\TAPatBootstrapTv$ \\
weighted probability of $\rho<0$ & $\TAPatBootstrapNegMass$ \\
\bottomrule
\end{tabular}
\end{center}
The exact step interval equals the corrected Monte-Carlo interval.
Reading the \TAPatBootstrapAtoms{} atoms as equally
likely would widen the interval to $[\TAPatBootstrapEqualLo,\TAPatBootstrapEqualHi]$,
\TAPatBootstrapWidthPct\% wider than the exact interval.  Multinomial weights
are required for the bootstrap distribution.  This
is a post-hoc supplementary analysis of the study's endpoint: it removes
Monte-Carlo resampling error from the reported percentiles, but it does not
enlarge the population and does not change the point estimate, decision table or
verdict.

\FloatBarrier
For the corrected matched-pair study, intervals are reported only when the
bootstrap mean and point estimate agree in sign; all bootstrap draws enter
the percentile calculation.

\subsection{A deterministic comparison under the same reference (Exp 53)}
\label{app:exp53}

The Gaussian reference is a stochastic test channel, so its noise-related component
contains channel randomness in addition to the observation noise, while a
trained codec is deterministic.  To check whether below-reference behavior is a
generic consequence of that comparison, we added a fixed zero-offset uniform
scalar quantizer, applied per channel in the standardized coordinates, to the
Exp 29 Part B population (\TANFitB{} codecs, \TAExpFiftyThreeTotal{} transitions).
Its step was chosen on validation data only, to match the learned codec's
five-seed mean occupancy (log-step bisection; accept at $5\times10^{-3}$ bit).  The measurements use the same sampled states,
noise replay and estimator as Exp 29, and the reference is evaluated at each
object's own realized test occupancy.

Learned codecs lie below the reference in \TAExpFiftyThreeLearnedBelow{} of
\TAExpFiftyThreeTotal{} transitions; the deterministic quantizer lies below it
in \TAExpFiftyThreeDetBelow{} and above it in the rest.  Median measured noise
shares are \TAExpFiftyThreeLearnedShare{} and \TAExpFiftyThreeDetShare{}, with
median excess \TAExpFiftyThreeLearnedExcess{} and
\TAExpFiftyThreeDetExcess{}.  Test occupancy drifts above the validation target
by up to \TAExpFiftyThreeGapMax{} bit; a post-hoc recomputation of the reference at the learned
codecs' test rates changes no sign.

The reading is a sensitivity check, not a correction of the reference.  The
deterministic object differs from the Gaussian reference in transform,
quantization and randomness, so the experiment does not decompose channel
randomness; it shows that determinism alone does not produce the observed
below-reference pattern.

\section{Estimator validation}
\label{app:validity}

\begin{figure}[t]
  \centering
  \includegraphics[width=\textwidth]{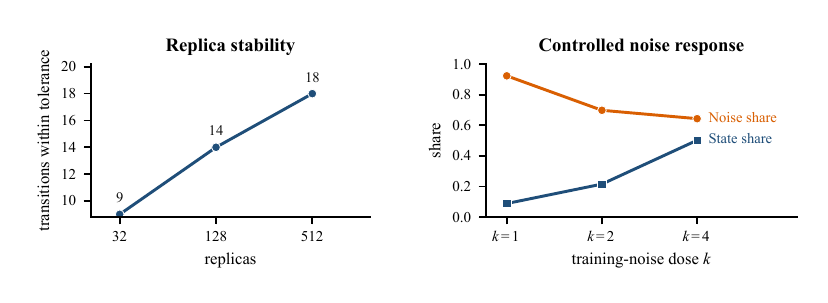}
  \caption{Left: stability across replica budgets. Right: reconstruction-error composition under added training noise. The separate analytic accuracy criterion failed (Appendix~\ref{app:validity}).}
  \label{fig:validity-detail}
\end{figure}

Figure~\ref{fig:validity-detail} collects the replica-convergence and
dose-response summaries discussed in this subsection.

\paragraph{Known observation-noise ordering.}
The measured noise-related component recovers the ordering of the known noise fractions.
The per-case means are \TACVOneLowALo{}--%
\TACVOneLowAHi{} and \TACVOneLowBLo{}--\TACVOneLowBHi{} in the two
high-observation-noise regimes, against \TACVOneHighLo{}--\TACVOneHighHi{} in the
low-observation-noise regime.  The criterion passes
\TACVOnePass{} of \TACVOneTotal{} cases, a one-sided rank test of the pooled
high-noise regimes against the low-noise one is significant in \TACVTwoSig{} of
\TACVTwoTests{} cases with and without a Bonferroni divisor of nine, and no state
component exceeds the signal energy (\TACVThreeViol{} violations in
\TACVThreeModels{} models).

\paragraph{Training-noise dose.}
The dose study re-trains one mechanism with an encoder-input corruption
$c\,\eta$ added to the observation, where $c = \sqrt{k-1}$ and $\eta$ is drawn
once and scaled so that its mean square equals the observation noise's; $k=1$
and $k=2$ reuse checkpoints from an earlier arm and only $k=4$ is newly
trained.  The dose study
predeclared no direction and makes no utility claim; the
degraded reconstruction at high dose is expected and is reported as such.  The per-record noise shares behind the three curve means
are recorded per capacity and initialization --- the state share of the
$7{\to}10$ increment runs \TADoseStateOne{}, \TADoseStateTwo{} and
\TADoseStateFour{} across the three doses.

\paragraph{Gaussian and distortion-matched controls.}
The Gaussian control source is constructed to be Gaussian with the same covariance
structure as the benchmark.  The ordering reported in the main text is
reproduced on this source --- every one of its \TAGaussN{} transitions falls
below the reference, with excesses from \TAGaussExcessLo{} to
\TAGaussExcessHi{}.  The ordering therefore also occurs without a regime mixture.  Moving each reference point to the
model's own measured distortion instead of its own rate preserves the ordering
over the \TAMatchedRows{} frozen models, whose reference shares then span
\TAMatchedShareLo{}--\TAMatchedShareHi{}.  Like the main construction, both
controls compare a deterministic codec with a stochastic reference.  The
closed-form quantizer below isolates that distinction.

\paragraph{Replica detail.}
The ladder is read in two layers.  The \TACoarseReplicas{}- and
\TAMidReplicas{}-replica rungs are assessed against the
\TAReplicaRefCoarse{}-replica reference, and the \TAReplicaRungFine{}-replica
rung is then assessed against the \TAReplicaRefFine{}-replica endpoint.  Against
\TAReplicaRefCoarse{}, \TAAgreeCoarse{} and \TAAgreeMid{} of the
\TANTransitionsA{} transitions fall within the operational tolerance; against
\TAReplicaRefFine{}, all \TAAgreeFine{} do.  With the endpoint count,
\TAResolvedPos{} of the \TANTransitionsA{} transitions have a positive corrected
state gain with an adjusted interval excluding zero, and \TAUnresolved{} does
not.
The bias the estimator removes at the coarse rung spans
$\TAMCBiasReplayLo$--$\TAMCBiasReplayHi$, against
$\TAMCBiasEndLo$--$\TAMCBiasEndHi$ at the endpoint.  It is small in absolute
terms, but of the same order as the gains themselves, which is why the endpoint
count is the one reported.  The across-stream spread that underpins the endpoint
prediction is a descriptive spread over the three declared streams, not a
variance decomposition and not an error bound.

The closed-form construction is a scalar Gaussian source with unit variance, a
deterministic uniform quantizer and a decoder equal to the quantizer, so the
conditional mean and both components are available through the standard normal
CDF.  Rate is the entropy of the quantized output, computed from the
output-bin probabilities (the finest step's realized rate is
\TARateCorrectedStep{} bit), and the second-moment-matched reference is evaluated at that
same realized rate.  The grid is three noise settings, four step sizes and two
codec offsets: a zero-offset uniform quantizer alone has a nearly unbiased
conditional mean that would leave the state component untested.

The noiseless criterion (K1) passes: at zero observation noise the corrected noise-related component is
below $10^{-30}$ in the noiseless cells, while the reference's \emph{level}
share $q^{\sref}$ assigns \TAKOneRefLo{}--\TAKOneRefHi{} of its reconstruction error to
the noise-related component.  $q^{\sref}$ is its noise-related component over its own reconstruction error at one
configuration, which is the analytic study's own quantity and not the ledger's
Eq.~\eqref{eq:refshare}.  Where observation noise dominates the two agree to a
median ratio of \TAKThreeMedian{}.

The second criterion, the finite-replica accuracy criterion (K2),
failed.  It required two conditions: the corrected error against the closed form had to
fall as the replica count rose, and at the finest rung it had to lie within one
Monte Carlo standard error.  Judged component by component, \TAKTwoViolOwn{} of
the \TAKTwoComparisons{} comparisons --- \TAKTwoConfigs{} configurations times two
components, not \TAKTwoComparisons{} independent configurations --- exceed their own
standard error, and \TAKTwoViolMaxSe{} exceed it when both components are judged
against the larger of the two.  Among the violations, the error runs from $\TAKTwoErrLo{}$ to $\TAKTwoErrHi{}$ and is \TAKTwoRatioLo{}--\TAKTwoRatioHi{}
of the component's own standard error.  The calibration claim of that study is withdrawn.

The study's \emph{gain} share over adjacent step sizes has the same qualitative
behavior as its level share.  In the noisy settings the deterministic codec's noise
gain share is \TAGainDetNoisy{} against a reference gain share of
\TAGainRefNoisy{}, and in the noiseless cell it is \TAGainDetNoiseless{}
against \TAGainRefNoiseless{}.  The two agree where observation noise dominates
and separate exactly where it vanishes, which is the same boundary K1 draws.

\FloatBarrier
\section{Allocation results and mechanism controls}
\subsection{Allocation across capacities and regimes}
\label{app:allocation}

\begin{figure}[t]
  \centering
  \includegraphics[width=\textwidth,trim=0bp 4.5bp 0bp 4bp,clip]{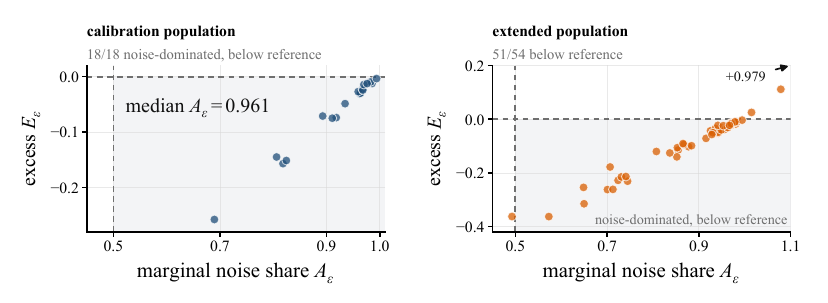}
  \caption{Noise-related gain shares and reference-relative excesses in the calibration (left) and capacity-sweep (right) populations. Dashed guides mark $A_\varepsilon=0.5$ and $E_\varepsilon=0$. The arrow labels the out-of-range transition.}
  \label{fig:character}
\end{figure}

Table~\ref{tab:full-ledger} reports the extended population's
\TANTransitionsB{} transitions, each aggregated over the \TANSeeds{}
initializations of its cell.  Columns are the nominal per-token step, the
realized occupancy increment, the reconstruction gain, its state and noise
components, the marginal noise share, and the same-occupancy reference gain
share.
A transition whose reconstruction gain is not positive has no defined share and
is marked as such rather than dropped.  The excess over the reference is the
difference of the last two columns.

% Generated by research_rebuild/make_tokaudit_assets.py
% Every value in this file is read from a frozen bundle.  Do not hand-edit.
\begin{table}[ht]
  \centering\scriptsize
  \setlength{\tabcolsep}{3.2pt}
  \begin{tabular}{@{}llrrrrrrrr@{}}
    \toprule
    regime & family & $C_1{\to}C_2$ & $\Delta R_{\mathrm{nom}}$ & $\Delta H$ & $\Gx$ & $\Gs$ & $\Geps$ & $\Aeps$ & $\Aref$ \\
    \midrule
    highpred\_highSNR & bsq & $2{\to}3$ & 2 & 1.27 & 0.1569 & 0.0796 & 0.0773 & 0.493 & 0.855 \\
    highpred\_highSNR & bsq & $3{\to}4$ & 2 & 1.18 & 0.0926 & 0.0326 & 0.0601 & 0.649 & 0.902 \\
    highpred\_highSNR & bsq & $4{\to}5$ & 2 & 1.02 & 0.0750 & 0.0320 & 0.0430 & 0.573 & 0.935 \\
    highpred\_highSNR & bsq & $5{\to}6$ & 2 & 0.60 & 0.0349 & 0.0097 & 0.0253 & 0.724 & 0.951 \\
    highpred\_highSNR & bsq & $6{\to}7$ & 2 & 0.86 & 0.0408 & 0.0122 & 0.0286 & 0.701 & 0.963 \\
    highpred\_highSNR & bsq & $7{\to}8$ & 2 & 0.42 & 0.0076 & 0.0011 & 0.0065 & 0.855 & 0.969 \\
    highpred\_highSNR & bsq & $8{\to}9$ & 2 & 0.57 & 0.0253 & 0.0073 & 0.0180 & 0.713 & 0.974 \\
    highpred\_highSNR & bsq & $9{\to}10$ & 2 & 0.77 & 0.0209 & 0.0026 & 0.0183 & 0.877 & 0.979 \\
    highpred\_highSNR & fsq & $4{\to}7$ & 6 & 3.80 & 0.1054 & 0.0269 & 0.0785 & 0.745 & 0.975 \\
    highpred\_highSNR & fsq & $7{\to}10$ & 6 & 1.90 & 0.0184 & 0.0027 & 0.0157 & 0.852 & 0.992 \\
    highpred\_highSNR & vqvae & $2{\to}3$ & 2 & 1.54 & 0.1395 & 0.0410 & 0.0986 & 0.706 & 0.884 \\
    highpred\_highSNR & vqvae & $3{\to}4$ & 2 & 1.04 & 0.0609 & 0.0117 & 0.0492 & 0.807 & 0.927 \\
    highpred\_highSNR & vqvae & $4{\to}5$ & 2 & 0.56 & 0.0310 & 0.0083 & 0.0227 & 0.731 & 0.946 \\
    highpred\_highSNR & vqvae & $5{\to}6$ & 2 & 0.44 & 0.0206 & 0.0053 & 0.0153 & 0.741 & 0.954 \\
    highpred\_highSNR & vqvae & $6{\to}7$ & 2 & 0.17 & 0.0059 & 0.0009 & 0.0051 & 0.853 & 0.959 \\
    highpred\_highSNR & vqvae & $7{\to}8$ & 2 & 0.28 & 0.0111 & 0.0018 & 0.0093 & 0.837 & 0.962 \\
    highpred\_highSNR & vqvae & $8{\to}9$ & 2 & 0.12 & 0.0042 & 0.0015 & 0.0027 & 0.650 & 0.964 \\
    highpred\_highSNR & vqvae & $9{\to}10$ & 2 & 0.06 & 0.0017 & -0.0001 & 0.0018 & 1.079 & 0.967 \\
    highpred\_lowSNR & bsq & $2{\to}3$ & 2 & 1.29 & 0.1291 & 0.0171 & 0.1120 & 0.867 & 0.961 \\
    highpred\_lowSNR & bsq & $3{\to}4$ & 2 & 1.15 & 0.1020 & 0.0073 & 0.0948 & 0.929 & 0.974 \\
    highpred\_lowSNR & bsq & $4{\to}5$ & 2 & 1.00 & 0.0683 & 0.0079 & 0.0604 & 0.884 & 0.983 \\
    highpred\_lowSNR & bsq & $5{\to}6$ & 2 & 0.61 & 0.0276 & 0.0023 & 0.0253 & 0.916 & 0.987 \\
    highpred\_lowSNR & bsq & $6{\to}7$ & 2 & 0.94 & 0.0401 & 0.0025 & 0.0377 & 0.938 & 0.990 \\
    highpred\_lowSNR & bsq & $7{\to}8$ & 2 & 0.48 & 0.0094 & 0.0002 & 0.0091 & 0.974 & 0.992 \\
    highpred\_lowSNR & bsq & $8{\to}9$ & 2 & 0.53 & 0.0221 & 0.0013 & 0.0208 & 0.942 & 0.993 \\
    highpred\_lowSNR & bsq & $9{\to}10$ & 2 & 0.72 & 0.0200 & 0.0004 & 0.0196 & 0.980 & 0.995 \\
    highpred\_lowSNR & fsq & $4{\to}7$ & 6 & 3.83 & 0.0884 & 0.0040 & 0.0844 & 0.955 & 0.993 \\
    highpred\_lowSNR & fsq & $7{\to}10$ & 6 & 1.99 & 0.0166 & 0.0003 & 0.0162 & 0.980 & 0.998 \\
    highpred\_lowSNR & vqvae & $2{\to}3$ & 2 & 1.58 & 0.1387 & 0.0104 & 0.1283 & 0.925 & 0.968 \\
    highpred\_lowSNR & vqvae & $3{\to}4$ & 2 & 0.96 & 0.0584 & 0.0030 & 0.0554 & 0.949 & 0.980 \\
    highpred\_lowSNR & vqvae & $4{\to}5$ & 2 & 0.66 & 0.0313 & 0.0018 & 0.0295 & 0.942 & 0.985 \\
    highpred\_lowSNR & vqvae & $5{\to}6$ & 2 & 0.40 & 0.0168 & 0.0005 & 0.0162 & 0.968 & 0.988 \\
    highpred\_lowSNR & vqvae & $6{\to}7$ & 2 & 0.22 & 0.0089 & 0.0002 & 0.0087 & 0.979 & 0.989 \\
    highpred\_lowSNR & vqvae & $7{\to}8$ & 2 & 0.24 & 0.0086 & 0.0003 & 0.0082 & 0.963 & 0.990 \\
    highpred\_lowSNR & vqvae & $8{\to}9$ & 2 & 0.16 & 0.0051 & 0.0003 & 0.0048 & 0.942 & 0.991 \\
    highpred\_lowSNR & vqvae & $9{\to}10$ & 2 & 0.04 & 0.0002 & -0.0002 & 0.0005 & 1.969 & 0.990 \\
    lowpred\_lowSNR & bsq & $2{\to}3$ & 2 & 1.28 & 0.1243 & 0.0167 & 0.1075 & 0.865 & 0.956 \\
    lowpred\_lowSNR & bsq & $3{\to}4$ & 2 & 1.15 & 0.1023 & 0.0066 & 0.0958 & 0.936 & 0.970 \\
    lowpred\_lowSNR & bsq & $4{\to}5$ & 2 & 0.99 & 0.0667 & 0.0047 & 0.0620 & 0.930 & 0.980 \\
    lowpred\_lowSNR & bsq & $5{\to}6$ & 2 & 0.63 & 0.0291 & 0.0021 & 0.0270 & 0.928 & 0.985 \\
    lowpred\_lowSNR & bsq & $6{\to}7$ & 2 & 0.93 & 0.0378 & 0.0019 & 0.0359 & 0.949 & 0.989 \\
    lowpred\_lowSNR & bsq & $7{\to}8$ & 2 & 0.49 & 0.0095 & 0.0004 & 0.0092 & 0.962 & 0.991 \\
    lowpred\_lowSNR & bsq & $8{\to}9$ & 2 & 0.54 & 0.0219 & 0.0007 & 0.0213 & 0.970 & 0.992 \\
    lowpred\_lowSNR & bsq & $9{\to}10$ & 2 & 0.70 & 0.0197 & 0.0007 & 0.0190 & 0.963 & 0.994 \\
    lowpred\_lowSNR & fsq & $4{\to}7$ & 6 & 3.80 & 0.0858 & 0.0015 & 0.0843 & 0.982 & 0.992 \\
    lowpred\_lowSNR & fsq & $7{\to}10$ & 6 & 2.03 & 0.0166 & 0.0001 & 0.0165 & 0.994 & 0.998 \\
    lowpred\_lowSNR & vqvae & $2{\to}3$ & 2 & 1.56 & 0.1348 & 0.0079 & 0.1269 & 0.941 & 0.964 \\
    lowpred\_lowSNR & vqvae & $3{\to}4$ & 2 & 0.95 & 0.0566 & 0.0026 & 0.0540 & 0.953 & 0.977 \\
    lowpred\_lowSNR & vqvae & $4{\to}5$ & 2 & 0.71 & 0.0335 & 0.0010 & 0.0325 & 0.970 & 0.983 \\
    lowpred\_lowSNR & vqvae & $5{\to}6$ & 2 & 0.33 & 0.0133 & 0.0004 & 0.0129 & 0.971 & 0.986 \\
    lowpred\_lowSNR & vqvae & $6{\to}7$ & 2 & 0.25 & 0.0100 & 0.0003 & 0.0097 & 0.970 & 0.987 \\
    lowpred\_lowSNR & vqvae & $7{\to}8$ & 2 & 0.25 & 0.0087 & 0.0002 & 0.0085 & 0.979 & 0.988 \\
    lowpred\_lowSNR & vqvae & $8{\to}9$ & 2 & 0.17 & 0.0062 & 0.0002 & 0.0060 & 0.967 & 0.989 \\
    lowpred\_lowSNR & vqvae & $9{\to}10$ & 2 & 0.06 & 0.0007 & -0.0000 & 0.0007 & 1.015 & 0.989 \\
    \bottomrule
  \end{tabular}
  \caption{Reconstruction-gain decomposition for \TANTransitionsB{} capacity-sweep transitions, aggregated over \TANSeeds{} initializations per cell. Nominal capacity and joint-token occupancy increments are in bits. Shares are undefined for nonpositive reconstruction gains; those transitions remain listed.}
  \label{tab:full-ledger}
\end{table}

Table~\ref{tab:regime-summary} groups the same shares by observation-noise level,
showing both the stronger noise dominance in high-noise cells and the
exceptions in low-noise cells.  Calibration and extended populations appear in
separate blocks.

% Generated by research_rebuild/revision_20260915/make_review_summary_assets.py
% Derived display values from frozen bundles.  Do not hand-edit.
\begin{table}[ht]
  \centering\small
  \begin{tabular}{@{}llrrrr@{}}
    \toprule
    population & observation-noise level & $n$ & min & median & max \\
    \midrule
    calibration & high (low pred.) & 6 & 0.935 & 0.973 & 0.994 \\
    calibration & high (high pred.) & 6 & 0.911 & 0.963 & 0.986 \\
    calibration & low & 6 & 0.690 & 0.821 & 0.918 \\
    extended & high (low pred.) & 18 & 0.865 & 0.965 & 1.015 \\
    extended & high (high pred.) & 18 & 0.867 & 0.946 & 1.969 \\
    extended & low & 18 & 0.493 & 0.736 & 1.079 \\
    \bottomrule
  \end{tabular}
  \caption{Marginal noise shares by population and observation-noise level. Medians are taken across transitions; each transition share is a ratio of seed-mean gains (\secref{app:aggregation}). Mechanisms contribute unequal numbers of transitions, so population medians weight transitions equally. Minima and maxima include all shares.}
  \label{tab:regime-summary}
\end{table}

\subsection{Linear-readout baselines}
\label{app:ceilings}

Table~\ref{tab:ceilings} compares linear readouts of observations, states, and
their combination on the same four-lag history used for utility.  The two
state-informed arms give nearly identical scores.  No tokenized arm in this
population exceeds the observation-only baseline, although a nonlinear latent
representation could allow a linear readout to do so.  These unquantized arms
are linear-readout baselines, not Bayes ceilings.

% Generated by research_rebuild/make_tokaudit_assets.py
% Every value in this file is read from a frozen bundle.  Do not hand-edit.
\begin{table}[ht]
  \centering\small
  \begin{tabular}{@{}lrrr@{}}
    \toprule
    regime & raw observation only & latent state only & observation and state \\
    \midrule
    low pred., high obs.\ noise & 0.1159 & 0.4361 & 0.4360 \\
    high pred., high obs.\ noise & 0.2367 & 0.7925 & 0.7925 \\
    high pred., low obs.\ noise & 0.4967 & 0.7925 & 0.7925 \\
    \bottomrule
  \end{tabular}
  \caption{Linear-readout baselines using the same four-step history: per-step cross-sectional RankIC for the first state channel, by regime.}
  \label{tab:ceilings}
\end{table}

\subsection{Mechanism controls}
\label{app:mechanism}

Five descriptive control arms examine whether codebook updates, parameter
count or loss weights account for the family differences.

Before those controls, the display below prints the contrast itself from the same
frozen rows as Table~\ref{tab:full-ledger}.  Each cell sums the
\mbox{$7{\to}10$}-bit span's seed-mean increments: three one-bit steps for RBQ
and VQ-VAE, one three-bit step for FSQ.  $\Delta H$ is the realized joint-token
occupancy increment in bits; $\GU$ is the same rank-information-coefficient gain
as in Table~\ref{tab:full-ledger}.  VQ-VAE adds the least occupancy in every
regime, and its utility increment is the smallest and negative in the two
high-predictability regimes.

\begin{center}\small
\begin{tabular}{@{}llrr@{}}
  \toprule
  regime & family & $\Delta H$ (bits) & $\GU$ \\
  \midrule
% The macro names carry the historical inverted SNR naming (HighPredHighNoise
% is the SNR-1.0 regime); the printed labels follow Table~\ref{tab:regimes}.
  high pred., high obs.\ noise & RBQ & \TAOccSpanHighPredLowNoiseBsqD & $\TAOccSpanHighPredLowNoiseBsqU$ \\
                               & VQ-VAE & \TAOccSpanHighPredLowNoiseVqvaeD & $\TAOccSpanHighPredLowNoiseVqvaeU$ \\
                               & FSQ & \TAOccSpanHighPredLowNoiseFsqD & $\TAOccSpanHighPredLowNoiseFsqU$ \\
  high pred., low obs.\ noise  & RBQ & \TAOccSpanHighPredHighNoiseBsqD & $\TAOccSpanHighPredHighNoiseBsqU$ \\
                               & VQ-VAE & \TAOccSpanHighPredHighNoiseVqvaeD & $\TAOccSpanHighPredHighNoiseVqvaeU$ \\
                               & FSQ & \TAOccSpanHighPredHighNoiseFsqD & $\TAOccSpanHighPredHighNoiseFsqU$ \\
  low pred., high obs.\ noise  & RBQ & \TAOccSpanLowPredLowNoiseBsqD & $\TAOccSpanLowPredLowNoiseBsqU$ \\
                               & VQ-VAE & \TAOccSpanLowPredLowNoiseVqvaeD & $\TAOccSpanLowPredLowNoiseVqvaeU$ \\
                               & FSQ & \TAOccSpanLowPredLowNoiseFsqD & $\TAOccSpanLowPredLowNoiseFsqU$ \\
  \bottomrule
\end{tabular}
\end{center}

\paragraph{Codebook schedule and parameter count.}
Replacing the codebook update with an exponential moving average (decay $0.99$)
and restarting a code that has gone unused for ten steps raises first-stage
support by a factor of \TACTEmaSupport{} and cuts reconstruction error to
\TACTEmaRecon{} of the published value, adding \TACTEmaBits{} bits of
realized occupancy.  The $7{\to}10$ utility increment does not move despite the gains in code use
and reconstruction: the plateau survives this codebook-schedule change.  One schedule
intervention does not rule out an interaction with the optimizer, the budget or
the training path, and the claim is limited to the schedule it tested.  At the largest width the
repaired first stage saturates at \TACTEmaSupportCtl{} active codes against
\TACTEmaSupportBase{} for the published baseline.

Widening the binary mechanism to match the codebook model's parameter count
(\TACTParamCount{} parameters at the largest width) moves utility by
\TACTParamUtility{}\% and changes reconstruction by a factor of
\TACTParamRecon{}, without reordering the measured quantities.  Parameter count is a real
but modest contributor; the preserved ordering is a qualitative result.

Over the extended population the pooled rank correlations with the utility gain
are \TASpearmanSignalB{} for the state gain, \TASpearmanReconB{} for total
reconstruction gain, \TASpearmanOccupancyB{} for occupancy gain and
\TASpearmanExcessB{} for the reference-relative excess.  They are descriptive:
the ordering does not reproduce inside every mechanism-by-regime stratum, where
\TAPositiveSignalB{}, \TAPositiveReconB{} and \TAPositiveOccupancyB{} of
\TAStrata{} strata are positive for the three gains and
\TAPositiveExcessB{} is for the excess.  The pooled ordering is therefore
insufficient to rank the measures across strata.

\paragraph{Hyperparameters.}
Three arms move one hyperparameter each.  Driving the entropy weight to zero
reduces realized occupancy by \TACTEntropyOff{} bits and worsens
reconstruction; raising it leaves occupancy and reconstruction essentially
unchanged; lowering the codebook commitment leaves occupancy essentially
unchanged.  The contrast is therefore not an artifact of one hyperparameter
setting.  We make no directional claim from these arms.

\FloatBarrier
\section{Robustness and controlled training contrasts}
\begin{figure}[t]
  \centering
  \includegraphics[width=\textwidth,trim=0bp 9bp 0bp 9bp,clip]{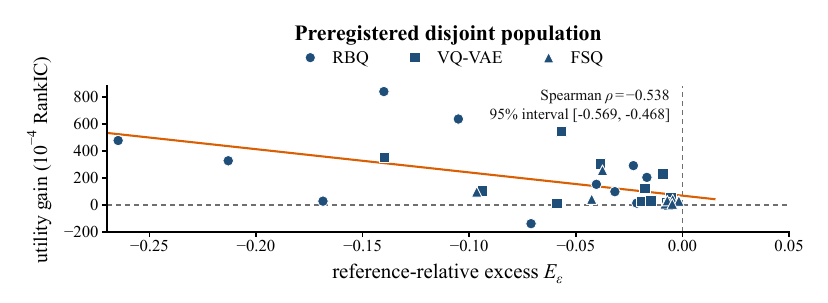}
  \caption{Higher reference-relative excess ($E_\varepsilon$) accompanies smaller utility gains in the prognostic population ($n=\TAPopTransitions$ transitions). The line is an OLS visual guide; inference uses Spearman correlation.}
  \label{fig:association}
\end{figure}

\subsection{Capacity-conditioned association}
\label{app:conditioning}

We analyze two existing populations under a fixed control scheme.
Population A is the prognostic population of \secref{sec:prognostic}: 32
transitions, 5 initializations, 4 regimes $\times$ 3 quantizer families.
Population B is the cross-generator population of
\secref{app:crossgenerator}: 24 transitions, 3 initializations, 2 generators
$\times$ 2 noise levels $\times$ 2 mechanisms.  The analysis uses existing fits.

The control scheme is S0 uncontrolled; S1 within-stratum mean; S2 control for
lower nominal bits; S3 adds nominal step; S4 (primary) adds realized occupancy
gain; S5 adds family indicators; S6 is S0 on the rung-disjoint subset.  The
seed axis is enumerated exactly: \TAPatCapacityAAtoms{} atoms on A and
\TAPatCapacityBAtoms{} on B.

\paragraph{Population A.}
\begin{center}\small
\begin{tabular}{@{}lcc@{}}
  \toprule
  spec & value (n) & exact seed interval \\
  \midrule
  Uncontrolled (S0) & $\TAPatCapacityABaseValue$ (\TAPatCapacityABaseN) & $[\TAPatCapacityABaseLo,\TAPatCapacityABaseHi]$ \\
  Within-stratum mean (S1) & $\TAPatCapacityAWithinValue$ (\TAPatCapacityAWithinN) & $[\TAPatCapacityAWithinLo,\TAPatCapacityAWithinHi]$ \\
  Lower nominal bits (S2) & $\TAPatCapacityALowerValue$ (\TAPatCapacityALowerN) & $[\TAPatCapacityALowerLo,\TAPatCapacityALowerHi]$ \\
  Nominal step (S3) & $\TAPatCapacityAStepValue$ (\TAPatCapacityAStepN) & $[\TAPatCapacityAStepLo,\TAPatCapacityAStepHi]$ \\
  Realized occupancy gain (S4) & $\TAPatCapacityAOccupancyValue$ (\TAPatCapacityAOccupancyN) & $[\TAPatCapacityAOccupancyLo,\TAPatCapacityAOccupancyHi]$ \\
  Family indicators (S5) & $\TAPatCapacityAFamilyValue$ (\TAPatCapacityAFamilyN) & $[\TAPatCapacityAFamilyLo,\TAPatCapacityAFamilyHi]$ \\
  Rung-disjoint subset (S6) & $\TAPatCapacityARungDisjointValue$ (\TAPatCapacityARungDisjointN) & $[\TAPatCapacityARungDisjointLo,\TAPatCapacityARungDisjointHi]$ \\
  \bottomrule
\end{tabular}
\end{center}

\paragraph{Population B.}
\begin{center}\small
\begin{tabular}{@{}lcc@{}}
  \toprule
  spec & value (n) & exact seed interval \\
  \midrule
  Uncontrolled (S0) & $\TAPatCapacityBBaseValue$ (\TAPatCapacityBBaseN) & $[\TAPatCapacityBBaseLo,\TAPatCapacityBBaseHi]$ \\
  Within-stratum mean (S1) & $\TAPatCapacityBWithinValue$ (\TAPatCapacityBWithinN) & $[\TAPatCapacityBWithinLo,\TAPatCapacityBWithinHi]$ \\
  Lower nominal bits (S2) & $\TAPatCapacityBLowerValue$ (\TAPatCapacityBLowerN) & $[\TAPatCapacityBLowerLo,\TAPatCapacityBLowerHi]$ \\
  Nominal step (S3) & $\TAPatCapacityBStepValue$ (\TAPatCapacityBStepN) & $[\TAPatCapacityBStepLo,\TAPatCapacityBStepHi]$ \\
  Realized occupancy gain (S4) & $\TAPatCapacityBOccupancyValue$ (\TAPatCapacityBOccupancyN) & $[\TAPatCapacityBOccupancyLo,\TAPatCapacityBOccupancyHi]$ \\
  Family indicators (S5) & $\TAPatCapacityBFamilyValue$ (\TAPatCapacityBFamilyN) & $[\TAPatCapacityBFamilyLo,\TAPatCapacityBFamilyHi]$ \\
  Rung-disjoint subset (S6) & $\TAPatCapacityBRungDisjointValue$ (\TAPatCapacityBRungDisjointN) & $[\TAPatCapacityBRungDisjointLo,\TAPatCapacityBRungDisjointHi]$ \\
  \bottomrule
\end{tabular}
\end{center}

On A, all \TAPatCapacityANegAtoms{} exact seed atoms of S4 are negative (range
$\TAPatCapacityAMin$ to $\TAPatCapacityAMax$); on B, \TAPatCapacityBPosWeightPct\% of the exact
weight is positive and the support covers zero (range $\TAPatCapacityBOccupancyLo$ to
$\TAPatCapacityBOccupancyHi$).  The pooled-vs-stratified difference is descriptive, not a
variance decomposition: A's between-stratum correlation is $\TAPatCapacityABetween$
and its within-stratum mean is $\TAPatCapacityAWithin$ (same sign), whereas B's
between-stratum correlation is $\TAPatCapacityBBetween$ and its within-stratum mean is
$\TAPatCapacityBWithin$ (opposite signs), consistent with the weaker pooled
correlation $\TAPatCapacityBBaseValue$.  At matched capacity, the
mechanism with the more negative excess has the higher utility gain in
\TAPatCapacityAConcordant{} of \TAPatCapacityAPairs{} pairs on A
($\rho=\TAPatCapacityAContrastRho$, sign-test $p=\TAPatCapacityAContrastP$) and
\TAPatCapacityBConcordant{} of \TAPatCapacityBPairs{} on B
($\rho=\TAPatCapacityBContrastRho$, $p=\TAPatCapacityBContrastP$).

The design limits how finely these effects can be separated.  A's design has
rank \TAPatCapacityADesignRank{} of \TAPatCapacityADesignRank{} with residual df
\TAPatCapacityAResidualDf; B's S4 has residual df \TAPatCapacityBDf.  Nominal step is a
function of the rung pair (VIF $\approx\TAPatCapacityAVif$ on A, $\TAPatCapacityBVif$ on B), so S3 adds
little over S2.  Realized occupancy gain is family-dependent, so S4 and S5
remove part of the mechanism contrast by construction.  The seed enumeration
resamples initializations only; the panel axis is not recomputed for the
partial statistics.  The auxiliary incremental-value regression is unaffected
and still reports no improvement from adding the state component; the estimands
differ.

\subsection{The cross-generator study}
\label{app:crossgenerator}

We test the allocation findings under two additional latent dynamics: a
nonlinear saturating recurrence and a seasonal process with stochastic
volatility.  Both use additive Gaussian observation noise.  This is a separate
population from the prognostic study of \secref{sec:prognostic}; its results
do not enlarge that population or its interval.

\paragraph{Design.}
The nonlinear recurrence combines a cyclic permutation with a saturating nonlinearity, driven
by standardized Student-$t$ innovations with a finite fourth moment.  The second
is a seasonal recurrence with a lag of one day-cycle and a stochastic-volatility
state.  We denote the two generators $g_1$ (nonlinear) and $g_2$ (seasonal).
In full, with $P(s_1,s_2,s_3,s_4)=(s_2,s_3,s_4,s_1)$ and $t_7$ a
standard Student-$t$ innovation with seven degrees of freedom,
\begin{align}
s_t &= 0.20\,s_{t-1} + 0.80\,\tanh(1.50\,P s_{t-1}) + 0.35\,\xi_t,
  & \xi_t &= \sqrt{5/7}\,t_7, \label{eq:crossg1}\\
h_t &= 0.95\,h_{t-1} + 0.15\,u_t, & u_t &\sim \mathcal{N}(0,1),
  \label{eq:crossg2vol}\\
s_t &= 0.25\,s_{t-1} + 0.55\,s_{t-24}
      + 0.20\,\sin(2\pi t/24+\phi)
      + 0.25\,e^{h_t/2}\,\eta_t,
  & \eta_t &\sim \mathcal{N}(0,1), \label{eq:crossg2}
\end{align}
with $\phi\sim\mathrm{Unif}(0,2\pi)$ per panel, entity and channel.  Both
processes run for a burn-in of \TACrossBurnIn{} steps before recording panels;
the split grid is 8/8/32 panels with 64 entities and 256 time steps.  The
latent volatility and phase are never given to the estimator.
They share one observation model,
$x_t = s_t + \sigma_\varepsilon\varepsilon_t$, whose noise scale and per-channel
normalizer are computed on the training split alone and then frozen.  The
non-Gaussianity and the heteroskedasticity therefore sit in the latent state, not
in the observation noise the estimator assumes.  The study trained
\TACrossCodecs{} codecs at one budget and target SNRs 0.1 and 1.0
(\TACrossGenerators{} generators $\times$
\TACrossSnrs{} observation-noise levels $\times$ \TACrossFamilies{} quantizer
families $\times$ \TACrossInitializations{} initializations $\times$
\TACrossWidths{} widths), and measured every transition with \TACrossStates{}
sampled states and \TACrossPairs{} antithetic pairs per state.  Its endpoint is
the \emph{stratified} form defined in the protocol: the within-noise-level
correlation taken separately, and the two averaged with equal weight.  That
endpoint is denoted $Q_g$, with $H_1\colon Q_g<0$; it differs from the pooled
single-$\rho$ estimand of \secref{sec:prognostic}.

\paragraph{Results and interpretation.}
The stratified allocation--utility association is negative on both generators:
$Q_g=\TACrossGenOneQ{}$ with interval \TACrossGenOneInterval{} for the
nonlinear process, and $Q_g=\TACrossGenTwoQ{}$ with interval
\TACrossGenTwoInterval{} for the seasonal process.  The corresponding $H_1$
verdicts are \TACrossGenOneVerdict{} and \TACrossGenTwoVerdict{}.
Median marginal noise shares are \TACrossGenOneShareMedian{} and
\TACrossGenTwoShareMedian{}, and all \TACrossTransitions{} transitions lie
below the reference: \TACrossGenOneBelowRef{} of \TACrossGenOneTransitions{}
and \TACrossGenTwoBelowRef{} of \TACrossGenTwoTransitions{}, respectively.
The pooled association, however, does not survive capacity
conditioning (\secref{app:conditioning}).

The two generators are fixed design choices.  Their intervals describe
evaluation uncertainty within these processes, not variation over a population
of dynamical models.  Within each observation-noise level, the
\TACrossPerStratum{} transitions are three consecutive rungs on each of two
quantizer ladders.  A shared capacity trend, with excess decreasing and utility
gain increasing, can itself produce a negative correlation.  Part of $Q_g$
therefore reflects capacity position rather than a cross-sectional relation.
Pooling the two observation-noise levels, which the stratified statistic
deliberately does not do, collapses the correlation to \TACrossGenOnePooled{} and
\TACrossGenTwoPooled{}, the visible size of the degeneracy.  This SNR
pooling is distinct from the capacity-conditioned analysis in
Section~\ref{app:conditioning}, which conditions on capacity position and realized
occupancy rather than averaging over observation-noise levels.

The reference contextualizes composition but adds little ranking separation on these
populations: the excess and measured share are close to rank-identical
(\TACrossGenOneRefRankCorr{} and \TACrossGenTwoRefRankCorr{}), and the
reference share nevertheless varies across transitions.
The raw gains correlate with utility in the opposite direction to the share: the
reconstruction and noise gains are strongly positive against utility, while the
share and the excess are negative.  A larger noise-related \emph{share} can accompany a smaller utility gain even when the absolute noise-related gain is larger.

\paragraph{Replica-budget calibration.}
An independent pilot of at most \TACrossPilotFits{} codecs calibrated the
replica budget against the criterion that the interval of the
\TACrossReplicaMid{}-minus-\TACrossReplicaHigh{} corrected state gain lies inside
a fixed band.  The corrected Monte-Carlo bias fell $\TACrossPilotBiasHigh$,
$\TACrossPilotBiasMid$ and $\TACrossPilotBiasLow$ across the three rungs, and
\TACrossReplicas{} replicas were adopted.  The finest rung is not a known truth, so this
bounds run precision and does not establish analytic accuracy.

\begin{figure}[t]
  \centering
  \includegraphics[width=\textwidth,trim=0bp 4.5bp 0bp 5.5bp,clip]{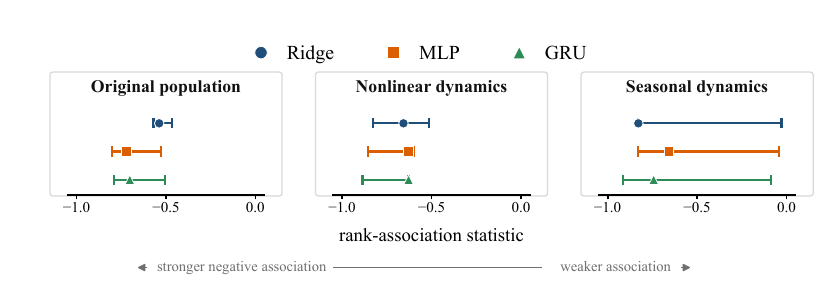}
  \caption{Allocation--utility associations across dynamics and readout families. Points and bars show rank-association estimates and intervals. Estimands vary across settings; only directions are directly comparable. Study-specific definitions: Appendices~\ref{app:conditioning}--\ref{app:readers}.}
  \label{fig:robustness}
\end{figure}

\subsection{Reader robustness and penalty sensitivity}
\label{app:readers}

We re-evaluate utility with nonlinear readouts and then widen the ridge penalty
grid.  These checks test whether the allocation--utility relation depends on
the original reader choice.

\paragraph{Nonlinear readouts.}
The prognostic population (B1, \TAReaderBOneCodecs{} existing codecs) is
re-read with ridge regression, a two-layer MLP and a one-layer GRU.  All see
the same window block.  They share a training budget and
\TAReaderInits{} paired initializations, with context lengths
\TAReaderBOneH{} and \TAReaderBTwoH{} for the prognostic and cross-generator
populations, respectively.  The same comparison is made on the
cross-generator population (B2, \TAReaderBTwoCodecs{} codecs).

Both nonlinear readers support the negative direction on the prognostic
population at per-interval alpha \TAReaderAlphaBOne{}.  All
\TAReaderBTwoClaims{} cross-generator claims are supported at alpha
\TAReaderAlphaBTwo{}.
\begin{center}\small
\begin{tabular}{@{}llcc@{}}
\toprule
population / dynamics & readout & stratified estimate & interval \\
\midrule
prognostic & MLP & \TAReaderBOneMlp & \TAReaderBOneMlpInterval \\
           & GRU & \TAReaderBOneGru & \TAReaderBOneGruInterval \\
cross-generator / nonlinear & MLP & \TAReaderBTwoMlpGenOne & \TAReaderBTwoMlpGenOneInterval \\
           & GRU & \TAReaderBTwoGruGenOne & \TAReaderBTwoGruGenOneInterval \\
cross-generator / seasonal & MLP & \TAReaderBTwoMlpGenTwo & \TAReaderBTwoMlpGenTwoInterval \\
           & GRU & \TAReaderBTwoGruGenTwo & \TAReaderBTwoGruGenTwoInterval \\
\bottomrule
\end{tabular}
\end{center}
Within the prognostic population, correlations are negative in all
\TAReaderBOneRegimes{} regimes for both of the \TAReaderNonlinear{}
nonlinear readers: \TAReaderBOneMlpLo{} to \TAReaderBOneMlpHi{} for the MLP
and \TAReaderBOneGruLo{} to \TAReaderBOneGruHi{} for the GRU.

These readouts use a shared per-channel scaler, whereas the original prognostic
readout uses a per-feature scaler.  B1 therefore evaluates existing codecs on
fresh test panels under a changed interface, rather than reproducing the
original endpoint.  B2 reuses the cross-generator codecs, test panels and audit
ledger, changing only utility; it is not an independent replication.  B1 and
B2 are analyzed separately.  Their direction statistics retain the shared
capacity-trend limitation described in \secref{app:crossgenerator}.

\paragraph{Ordering.}
A further comparison uses no excess at all: whether the readers agree on the
\emph{ordering} the
capacity sequence induces.  Kendall's $\tau_b$ between
the utility transition gains of each reader pair, averaged with equal weight over
strata, is positive with a lower interval bound above the
\TAReaderOrderThreshold{} bar in all \TAReaderOrderComputed{} intervals that were
computed; their lower bounds run from $\TAReaderOrderLoMin$ to
$\TAReaderOrderLoMax$, at a per-interval alpha of $\TAReaderOrderAlpha{}$.
The \TAReaderOrderThreshold{} threshold is a practical criterion chosen for
this study.  Agreement on capacity ordering leaves state fidelity and task
utility as distinct quantities.  The protocol declared a
family of \TAReaderOrderDeclared{} ordering claims --- three populations crossed
with three reader pairs --- and only \TAReaderOrderComputed{} were computed: the
second population averages its two generators into one point per pair, so the
per-generator ordering intervals the protocol names were never produced.  The
interval alpha is the Bonferroni allocation for the declared
\TAReaderOrderDeclared{}-claim family applied to the \TAReaderOrderComputed{}
computed claims, giving a conservative allocation.

The protocol also declares a band of $|\GU|\le \TAReaderNearZeroBand{}$ in
rank-information-coefficient units inside which a transition's utility gain is
treated as sign-unstable.  On the first population \TAReaderNearZeroBOne{}
reader-transition entries fall inside it, concentrated on the widest rung of each
ladder where the gain is smallest.  They are flagged and kept.  The band never
enters the Spearman or $\tau_b$ statistics, and it is never used to drop a
transition from any endpoint.  The learned readers themselves are fitted for at
most \TAReaderUpdates{} updates.  On a target that
is nonlinear in the window the MLP reached a mean per-step cross-sectional rank
information coefficient of \TAReaderCompetenceMlp{} against the ridge readout's
\TAReaderCompetenceRidge{}, showing that the MLP can exploit nonlinearity on this target. The GRU did not outperform the ridge readout; this check does not establish convergence of either reader.

\paragraph{The extended penalty grid.}
The ridge penalty is selected from a fixed grid of \TACrossGridPoints{} points on
inner training panels.  In \TACrossGridBoundary{} of the \TACrossCodecs{} codecs
the selected penalty is the largest grid point, so the selection is at the
boundary rather than interior.  This shows that the optimum on the searched grid lies at its edge;
a wider grid could change the selection.  The original grid is retained for the primary analysis, and the
\TAPatRidgeCodecs{} cross-generator codecs were re-read
with the ridge penalty selected from a grid extended by three decades:
\texttt{logspace(-4, 7, \TAPatRidgeExtendedGrid)} against
\texttt{logspace(-4, 4, \TAPatRidgeFrozenGrid)}.  The extended grid is a strict
superset, so the two selections are nested; features, scaler, inner split,
selection metric, target and test panels are unchanged, and
only the refit changes.

On the extended grid the largest point is selected \TAPatRidgeExtendedBoundary{}
of \TAPatRidgeCodecs{} times, and \TAPatRidgeMoved{} penalties move, all to the three new points.  Mean inner RankIC
changes by $\TAPatRidgeInnerChange$, mean test utility by $\TAPatRidgeTestMeanChange$,
and the maximum absolute test-utility change is $\TAPatRidgeTestMaxChange$.  Rank
agreement is $\TAPatRidgeRankAll$ over the \TAPatRidgeCodecs{} utilities and
$\TAPatRidgeRankTransitions$ over the 24 transition utility gains.  Both
generators' stratified $Q_g$ are identical to four decimals ($\TAPatRidgeQOneFrozen$
for g1 and $\TAPatRidgeQTwoFrozen$ for g2); pooled g1 moves from
$\TAPatRidgePooledOneFrozen$ to $\TAPatRidgePooledOneExtended$ and pooled g2 is unchanged
at $\TAPatRidgePooledTwo$.  A smaller amplitude is not the mechanism, because RankIC
is invariant to positive rescaling; a larger penalty changes the coefficient
direction, and the measured direction agreement is
$\TAPatRidgeDirectionAgreement$.  This post-hoc sensitivity leaves the claim
limited by the shared per-channel scaler, the two learned readers and the
near-rank identity between the measured share and the excess.

\subsection{Tokenizer controls: scalar binning and causal filtering}
\label{app:codec-controls}

A training-free scalar ladder and a limited causal encoder test whether noise
dominance depends on learned quantization or a pointwise receptive field.
The scalar study is descriptive.

\paragraph{Scalar-binning control.}
A two-stage residual scalar quantizer with fixed grids---uniform (equal-width
bins over the training $[0.001,0.999]$ quantile range) and quantile
(equal-frequency bins)---was applied to the frozen Exp~39 generators
(\TAPatScalarExpThirtyNineTransitions{} transitions) and the three frozen VAR regimes (Exp~16)
(\TAPatScalarExpSixteenTransitions{} transitions), \TAPatScalarTransitions{} transitions in
all.  No model is trained.  The estimator, noise replay,
second-moment-matched reference and ridge readout are the same as in the main
analyses; the reader
sees the dequantized reconstruction (4 dimensions per step) instead of a
64-dimensional latent, which is declared.  Each of the \TAPatScalarChannels{} channels
and each of the \TAPatScalarStages{} stages uses \TAPatScalarBitsLo{}--\TAPatScalarBitsHi{} bits, so the
nominal rate is \TAPatScalarRateFactor{}$b$ bits per time point and the
ladder's capacity levels are \TAPatScalarRateLevels{}; the realized
occupancy is the joint entropy of the packed \TAPatScalarChannels{}$\times$\TAPatScalarStages{}-digit
per-step code, not the sum of the per-channel entropies.  The readout keeps a
\TAPatScalarReaderHistory{}-step history of that reconstruction, a different
interface from the four-step latent history the core reader uses, so this
control's utility column is not on the same footing as the learned arms'.

\begin{center}\small
\begin{tabular}{@{}lcc@{}}
  \toprule
  quantity & scalar ladder & Exp~39 learned arms (frozen) \\
  \midrule
  transitions & \TAPatScalarTransitions & \TAPatScalarLearnedTransitions \\
  median measured noise share $\Aeps$ & \textbf{\TAPatScalarShareMedian} &
    \TAPatScalarLearnedShareOne{} / \TAPatScalarLearnedShareTwo \\
  share range & \TAPatScalarShareLo--\TAPatScalarShareHi & --- \\
  share above one half & \textbf{\TAPatScalarAboveHalf/\TAPatScalarTransitions} &
    \TAPatScalarLearnedAboveHalf/\TAPatScalarLearnedTransitions \\
  excess below the reference & \textbf{\TAPatScalarBelow/\TAPatScalarTransitions} &
    \TAPatScalarLearnedBelow/\TAPatScalarLearnedTransitions \\
  excess above the reference & \textbf{\TAPatScalarAbove/\TAPatScalarTransitions} &
    \TAPatScalarLearnedAbove/\TAPatScalarLearnedTransitions \\
  median excess & $\TAPatScalarExcessMedian$ & negative (all) \\
  pooled $\rho(\Eeps,\GU)$ & \textbf{\TAPatScalarRho} &
    \TAPatScalarLearnedRhoOne{} / \TAPatScalarLearnedRhoTwo \\
  \bottomrule
\end{tabular}
\end{center}

The breakdown by quantization scheme and environment is:
\begin{center}\small
\begin{tabular}{@{}lcccc@{}}
\toprule
group & median share & median excess & below / above ref. & pooled $\rho$ \\
\midrule
uniform & \TAPatScalarUniformShare & $\TAPatScalarUniformExcess$ & \TAPatScalarUniformBelow{} / \TAPatScalarUniformAbove & $\TAPatScalarUniformRho$ \\
quantile & \TAPatScalarQuantileShare & $\TAPatScalarQuantileExcess$ & \TAPatScalarQuantileBelow{} / \TAPatScalarQuantileAbove & $\TAPatScalarQuantileRho$ \\
cross-generator & \TAPatScalarExpThirtyNineShare & --- & \TAPatScalarExpThirtyNineBelow{} / \TAPatScalarExpThirtyNineAbove & $\TAPatScalarExpThirtyNineRho$ \\
VAR regimes & \TAPatScalarExpSixteenShare & --- & \TAPatScalarExpSixteenBelow{} / \TAPatScalarExpSixteenAbove & $\TAPatScalarExpSixteenRho$ \\
\bottomrule
\end{tabular}
\end{center}

The noise dominance reproduces
and is stronger: a codec with no learning at all shows the pattern at least as
clearly.  ``Below the reference'' does not reproduce
(\TAPatScalarBelow/\TAPatScalarAbove{} split, median excess $\TAPatScalarExcessMedian$),
and the excess--utility association changes sign ($\TAPatScalarRho$).  This
control is not an implementation of Chronos: the official quantizer was not read
for this study, and the ladder is a two-stage residual scalar quantizer written
here.  The comparison against the learned arms is a descriptive nearest-neighbour
match on the realized occupancy increment alone: it matches neither occupancy
level nor any rate--distortion condition, and its only rule is to take the
closest row in the same cell.  Over the \TAPatScalarRateMatchPairs{} frozen
Exp~39 pairs the absolute gap between the two increments runs from
\TAPatScalarRateMatchGapMin{} to \TAPatScalarRateMatchGapMax{} bits per time
step, median \TAPatScalarRateMatchGapMedian{}.  These gaps limit the comparison to descriptive nearest neighbors; they do not
establish cross-mechanism rate equivalence.

\paragraph{Causal temporal encoder control.}
\TAPatTemporalCodecs{} codecs---the frozen Exp~39 grid, 2 generators $\times$ 2 noise levels
$\times$ 2 mechanisms $\times$ 4 widths $\times$ 3 seeds---were retrained with
one change: the encoder gains a depthwise causal 8-tap filter, initialized to
identity, after its GELU.  Every shared module starts from the pointwise model's
own initialization; the token rate per time step is unchanged (one token per
step, $2w$ bits); the decoder, quantizers, objective, budget, seeds, data,
reader recipe and measurement are unchanged.  The pointwise arm is the
Exp~39 population.  The window-wise noise replay
gives both arms the same window ends and the same noise, so the comparison is
paired at the evaluated step.

\begin{center}\small
\begin{tabular}{@{}lcc@{}}
  \toprule
  & pointwise (frozen) & temporal (new) \\
  \midrule
  parameter count & \TAPatTemporalParamsPointLo--\TAPatTemporalParamsPointHi &
    \TAPatTemporalParamsTemporalLo--\TAPatTemporalParamsTemporalHi
    (+\TAPatTemporalParamsAdded) \\
  test reconstruction MSE, mean & \TAPatTemporalTestMsePoint &
    \TAPatTemporalTestMseTemporal \\
  paired test MSE ratio, median / mean & --- &
    \textbf{\TAPatTemporalTestRatioMedian} / \TAPatTemporalTestRatioMean \\
  paired test MSE ratio, range & --- &
    \TAPatTemporalTestRatioLo--\TAPatTemporalTestRatioHi \\
  median measured noise share & \TAPatTemporalPointShare & \TAPatTemporalTemporalShare \\
  share range & \TAPatTemporalPointShareLo--\TAPatTemporalPointShareHi &
    \TAPatTemporalTemporalShareLo--\TAPatTemporalTemporalShareHi \\
  transitions below the reference & \TAPatTemporalPointBelow/\TAPatTemporalTransitions &
    \TAPatTemporalTemporalBelow/\TAPatTemporalTransitions \\
  median excess & $\TAPatTemporalPointExcess$ & $\TAPatTemporalTemporalExcess$ \\
  median utility gain & \TAPatTemporalPointUtility & \TAPatTemporalTemporalUtility \\
  median reconstruction gain & \TAPatTemporalPointReconstruction &
    \TAPatTemporalTemporalReconstruction \\
  median occupancy gain (bits) & \TAPatTemporalPointOccupancy &
    \TAPatTemporalTemporalOccupancy \\
  pooled $\rho(\Eeps,\GU)$ & $\TAPatTemporalPointPooledRho$ &
    $\TAPatTemporalTemporalPooledRho$ \\
  descriptive $Q_g$, g1 / g2 & $\TAPatTemporalPointQOne$ / $\TAPatTemporalPointQTwo$ &
    $\TAPatTemporalTemporalQOne$ / $\TAPatTemporalTemporalQTwo$ \\
  \bottomrule
\end{tabular}
\end{center}

The temporal arm's training-time validation
reconstruction MSE averages \TAPatTemporalValidationMseTemporal; the pointwise
arm's validation MSE is unavailable, so the fit comparison in the table is the
like-for-like test-side measurement: both columns are produced on the same test
windows and paired over the
\TAPatTemporalSampleSize{} codec pairs.  On that measurement the arms are closely matched.

The primary endpoint is the SNR-stratified mean paired change in the marginal
noise share, $\Delta\bar\Aeps_g = \tfrac12\operatorname{mean}\{\Delta\Aeps
\mid \mathrm{SNR}\ 0.1\} + \tfrac12\operatorname{mean}\{\Delta\Aeps
\mid \mathrm{SNR}\ 1.0\}$ with $\Delta\Aeps =
\Aeps^{\mathrm{temporal}}-\Aeps^{\mathrm{pointwise}}$, expressed in share
units (97.5\% paired bootstrap over initialization and panel, with a shared draw
for both arms and a Bonferroni allocation across the two generator claims): g1 gives $\Delta\bar\Aeps_g=\TAPatTemporalOnePoint$ with
interval $[\TAPatTemporalOneLo,\TAPatTemporalOneHi]$, and g2 gives
$\TAPatTemporalTwoPoint$ with $[\TAPatTemporalTwoLo,\TAPatTemporalTwoHi]$;
both are inconclusive.  The intervals constrain the paired effect on the noise
share to roughly two tenths of a percentage point on this grid.  The $Q_g$
correlations in the table are descriptive secondary statistics of a different
estimand, the allocation--utility stratified rank correlation; the share change
above is the one the interval bounds.  The ledger is otherwise unchanged: \TAPatTemporalTemporalBelow/\TAPatTemporalTransitions{}
remain below the reference, the median share moves by \TAPatTemporalShareMove, the
median occupancy gain by $\TAPatTemporalOccupancyMove$ bits, and the association with
utility survives (pooled $\TAPatTemporalPooledMoveLow$ against
$\TAPatTemporalPooledMoveHigh$; stratified $Q_g$ moves only on g1).  Fit quality is
closely matched on the paired test measurement, so differential fit quality does
not explain the near-zero point estimate; the paired endpoint nevertheless
remains inconclusive.  This comparison covers one architecture change, two
controlled generators and one budget.  This control does not establish the behavior of other temporal tokenizer architectures; the separate window-compression study is reported in \secref{app:temporal}.

\subsection{Matched-pair negative control}
\label{app:matched}

The matched-pair study attempts to train two arms with different allocations
but similar reconstruction gains and realized occupancy increments,
with a symmetric relative reconstruction tolerance of \TAMatchedReconTol{}
and an absolute joint-entropy tolerance of \TAMatchedEntropyTol{} bit.  At
three initializations the group-level object matched on both tolerances (relative
reconstruction difference \TAMatchedReconThree{}, joint-entropy difference
\TAMatchedEntropyThree{} bits), although only \TAMatchedPerSeedThree{} of the
three per-initialization pairs did.  At ten initializations the group-level match
fails both tolerances (\TAMatchedReconTen{} and \TAMatchedEntropyTen{} bits) with
\TAMatchedPerSeedTen{} of \TAMatchedSeedsTen{} per-seed pairs matching.  Those
tolerances are the pilot's own decision rule and were originally explored with
some results already visible; they are exploratory tolerances rather than formal
thresholds.  The extended study's interpretation rule has three
outcomes, and because no group-level match holds at ten initializations the
outcome is \emph{inconclusive on matching}.

The reported occupancies use paired-stage joint entropy.

\subsection{Initial training-objective contrast at matched reconstruction and occupancy diagnostics (Exp 48)}
\label{app:discrimination}

This paired training contrast tests whether the audit distinguishes allocations
when reconstruction and occupancy diagnostics agree.  Its protocol, matching
bands and decision table were fixed before the formal fits.  A separate pilot
selected the regularization strength using matching statistics only, without
reading allocation shares.

We use the nonlinear generator of \secref{app:crossgenerator} at target SNR
1.0, residual binary quantization, one capacity span (per-stage widths $4\to7$,
a nominal step of six bits per token) and one training budget.  Five paired initializations,
shared across the two conditions and the two capacities, give
\TAExpFortyEightCodecs{} formal codecs; the data seeds are new and disjoint
from every earlier population.

\paragraph{Training conditions.}
For a state $s$ let $x^{(1)}=s+\varepsilon^{(1)}$ and
$x^{(2)}=s+\varepsilon^{(2)}$ be independent draws of the declared
observation-noise law.  Both conditions train on both views against the
\emph{observation},
\begin{equation}
L(\lambda) = \tfrac12\bigl[L_{\rm rec}(x^{(1)})+L_{\rm rec}(x^{(2)})\bigr]
  + L_{\rm q} + \lambda\,\bigl\lVert f(x^{(1)})-f(x^{(2)})\bigr\rVert^2,
\label{eq:consistency}
\end{equation}
where $L_{\rm q}$ is the quantizer term the frozen objective already carries.
The plain arm is $\lambda=0$; the consistent arm adds the consistency term at
$\lambda=\TAExpFortyEightLambda{}$, the same value at both capacities,
chosen by the pilot as the largest candidate whose reconstruction gain and
level differences sat inside the bands.  The state $s$ is never a
reconstruction label in either arm.

\paragraph{Matching.}
The three bands were frozen before the formal test: a relative difference of
at most \TAExpFortyEightBandGain{} in the transition reconstruction gain,
at most \TAExpFortyEightBandOccupancy{} bit in the realized occupancy
increment, and at most \TAExpFortyEightBandLevel{} in the reconstruction
error at each capacity.  The gate is applied on the development panels, and
there the two arms sit inside all three: their reconstruction gains differ by
\TAExpFortyEightGainRelPct\%, their occupancy increments by
\TAExpFortyEightOccGap{} bits, and the two capacities' reconstruction-error
levels by \TAExpFortyEightLevelLowPct\% and \TAExpFortyEightLevelHighPct\%.
These are practical matching bands with a deterministic decision rule, not a
formal equivalence test.  Both arms satisfy the rule, giving the verdict
\emph{matched}.

\paragraph{Result.}
The ledger below is measured on the test panels.  The gate above is applied on
the development panels; on the test panels the same descriptive verdict,
\emph{matched}, holds.
\begin{center}\small
\begin{tabular}{@{}lcc@{}}
\toprule
& plain & consistent \\
\midrule
reconstruction gain $\Gx$ & \TAExpFortyEightGainPlain & \TAExpFortyEightGainConsistent \\
occupancy increment $\Delta H$ (bits) & \TAExpFortyEightOccPlain & \TAExpFortyEightOccConsistent \\
state gain $\Gs$ & \TAExpFortyEightStatePlain & \TAExpFortyEightStateConsistent \\
noise gain $\Geps$ & \TAExpFortyEightNoisePlain & \TAExpFortyEightNoiseConsistent \\
noise share $\Aeps$ & \TAExpFortyEightSharePlain & \TAExpFortyEightShareConsistent \\
excess $\Eeps$ & \TAExpFortyEightExcessPlain & \TAExpFortyEightExcessConsistent \\
\bottomrule
\end{tabular}
\end{center}

The arms allocate that gain differently: the consistent arm moves share from
the noise component to the state component, a paired change of
\TAExpFortyEightDeltaShare{} with a
finite-sample seed-bootstrap interval $[\TAExpFortyEightDeltaShareLo,\TAExpFortyEightDeltaShareHi]$
over the \TAExpFortyEightSeeds{} initializations.  Both arms sit below the
same matched reference (\TAExpFortyEightRefShare{}), so the excess moves with
the share.  Exact enumeration of the $n=\TAExpFortyEightSeeds{}$ seed
bootstrap --- all \TAExpFortyEightAtoms{} multinomial atoms --- leaves no atom
on the other side of zero (support \TAExpFortyEightSupportMin{} to
\TAExpFortyEightSupportMax{}).  The paired utility change is
\TAExpFortyEightDeltaUtility{} rank information coefficient, which is
consistent with no movement and is reported without a direction.

This is a controlled demonstration for one generator, mechanism, capacity span,
budget and intervention.  It establishes a detectable allocation difference
at matched diagnostics, with no directional utility claim.  It establishes
neither a general capacity-selection rule nor a causal effect of allocation
on utility.  The earlier matched-pair study remains inconclusive on matching
(\secref{app:matched}).

\FloatBarrier
\subsection{Independent training-objective confirmation and a second mechanism (Exp 52)}
\label{app:exp52}

The bands inherit the Exp48 design: 0.10 and 0.05 bit retain the earlier matched-pair study's exploratory tolerances for comparability; a 0.15 relative limit at each reconstruction endpoint was added so endpoint levels would not already distinguish the arms. These investigator-specified tolerances are not relaxed for formal evaluation, and the pilot selects strength within the fixed bands.

The matched-diagnostic case of \secref{app:discrimination} was re-run with fresh
training data and initializations (Unit~A, residual binary quantization) and
extended to a second mechanism (Unit~B, VQ-VAE).  Each unit trains
\TAExpFiftyTwoNA{} independent blocks $\times$ two arms $\times$ two capacities
$=$ \TAExpFiftyTwoFits{} formal fits under one frozen recipe: 1,600 updates, batch 8,192, Adam
$10^{-4}$, cosine schedule, snapshots by validation reconstruction only,
2,048 states and 1,024 antithetic pairs, 32 test panels and 8 validation
panels.  Unit~A fixes the RBQ consistency weight at $\lambda=0.10$; Unit~B takes
$\lambda=0.10$ from an allocation-blind pilot that read reconstruction and
occupancy diagnostics only (the largest passing candidate in
$\{0.01,0.03,0.10,0.30,1.00\}$).

The matching bands are a relative reconstruction-gain difference of
at most 0.10, an occupancy-increment difference of at most 0.05 bit, and a
level difference of at most 0.15 at each capacity.  Both units pass on
validation (gains \TAExpFiftyTwoGainRelA\% and \TAExpFiftyTwoGainRelB\%;
occupancy gaps \TAExpFiftyTwoOccGapA{} and \TAExpFiftyTwoOccGapB{} bits;
levels \TAExpFiftyTwoLevelLowA/\TAExpFiftyTwoLevelHighA\% and
\TAExpFiftyTwoLevelLowB/\TAExpFiftyTwoLevelHighB\%) and on the test panels.
These are practical bands with a deterministic rule, not an equivalence test;
test matching is an additional condition on the same panels as the endpoint,
not an independent replication.

The measured paired change in marginal noise share is
\TAExpFiftyTwoDeltaA{} for Unit~A with interval \TAExpFiftyTwoIntervalA{}
(simulation-calibrated, $k=2.50$, $B=10{,}000$; not a nominal 95\% confidence
interval) and \TAExpFiftyTwoDeltaB{} for Unit~B with interval
\TAExpFiftyTwoIntervalB{} (Student-$t$, 15 degrees of freedom, hierarchical
bootstrap standard error; finite-sample approximate).  All bootstrap draws are
negative in both units.  Utility was not measured in either unit.

The supported reading is: within this design, at reconstruction and occupancy
diagnostics inside the declared practical bands, the audit reads a composition
difference; the RBQ case reproduces under new data and initializations, and the
VQ-VAE case shows the same direction under a pilot-selected weight; the units use different mechanisms and interval methods,
so their effect sizes describe their respective designs rather than a controlled
mechanism comparison.

\begin{figure}[t]
  \centering
  \includegraphics[width=\textwidth]{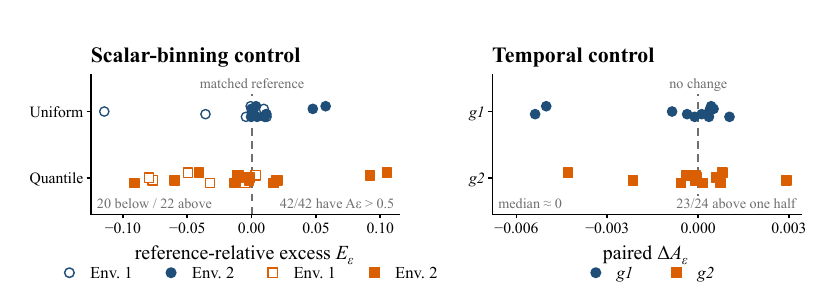}
  \caption{Tokenizer controls. Left: scalar binning shows noise-dominated gains but mixed positions relative to the reference. Right: paired noise-share changes from adding a causal temporal encoder. Dashed lines mark the reference and zero change, respectively. Full endpoints: Appendix~\ref{app:codec-controls}.}
  \label{fig:controls}
\end{figure}

\paragraph{Complete matching and post-hoc component readouts.}
\begin{table}[t]
\centering\small
\setlength{\tabcolsep}{4pt}
\begin{tabular}{@{}llllll@{}}
\toprule
Unit & Split & $G_x$ gap (\%) & $\Delta H$ (bit) & $D_x(C_1)$ (\%) & $D_x(C_2)$ (\%) \\
\midrule
A & val & 0.6427 & 0.000208 & 2.6529 & 4.9043 \\
A & test & 0.6127 & 0.000136 & 2.6376 & 4.8636 \\
B & val & 2.9622 & 0.037404 & 4.9920 & 5.7567 \\
B & test & 3.0892 & 0.037317 & 5.0298 & 5.7599 \\
\bottomrule
\end{tabular}
\caption{Validation and test matching gaps. Validation supplies the primary matching decision; test matching shares the endpoint panels.}
\label{tab:new-matching}
\end{table}

\begin{table}[t]
\centering\small
\setlength{\tabcolsep}{4pt}
\begin{tabular}{@{}llllll@{}}
\toprule
Unit & Gain & plain & consistent & $\delta$ & Interval \\
\midrule
A & GM & -0.019614 & -0.006581 & +0.013033 & [+0.010655, +0.015386] \\
A & GV & -0.049545 & -0.040902 & +0.008643 & [+0.006357, +0.010833] \\
A & Gs & 0.029891 & 0.034301 & +0.004410 & [+0.003634, +0.005197] \\
B & GM & -0.006721 & -0.004549 & +0.002172 & [+0.001324, +0.003071] \\
B & GV & -0.011946 & -0.012651 & -0.000705 & [-0.001831, +0.000369] \\
B & Gs & 0.005348 & 0.008182 & +0.002833 & [+0.002473, +0.003180] \\
\bottomrule
\end{tabular}
\caption{Post-hoc component gains and training contrasts. Gains are low minus high capacity; contrasts are consistent minus plain. Both units use 128 replay pairs and a block/panel/state bootstrap.}
\label{tab:new-readouts}
\end{table}

The readouts use Exp54's original 128-pair budget, states and noise seeds. Both units use 10,000 block--panel--state bootstrap draws, with common indices across capacities, arms and the three quantities. Percentile approximate 95\% intervals are distinct from the original 1,024-pair endpoints. $\delta$ denotes consistent minus plain. A has positive $\delta G_V$; B is unresolved; both have positive $\delta G_s$. Both arms' $G_V$ remain negative.

\subsection{Window-compression comparison: position-wise components and budget (Exp 55, 60--62)}
\label{app:temporal}
Exp55 compresses a 16-step, four-channel window to one discrete token, with no input skip connection and widths $4/7/10$. Whole reads all steps; local reads the last four; both reconstruct the window. Sixteen independent data/initialization blocks at two SNRs give 192 fits: 3,200 updates, 1,024 windows per update, Adam $10^{-4}$, snapshots by validation reconstruction. Testing uses 32 panels, 64 state windows per panel and 128 replay pairs. Joint single-token entropy is measured over all 32,768 test windows in bits/window and bits/step. In Exp55, \TATempFinalSnapshot{} of \TATempFinalFits{} fits selected the last of the eight validation snapshots, with similar selection frequencies across codebook widths and both receptive fields, so the capacity sequences are compared at one common budget rather than at a converged optimum.

The original shares in the main text are medians of block ratios, a descriptive statistic distinct from the default ratio of mean gains. First-step whole-minus-local state-gain differences are \TATempDeltaGsLow{} (\TATempDeltaGsLowInterval) and \TATempDeltaGsHigh{} (\TATempDeltaGsHighInterval), with paired bootstrap intervals over 16 blocks. Rate and parameter differences prevent attribution to temporal structure alone.
\paragraph{Position interpretation and probe.} Exp60 separates the visible last four and invisible first twelve targets post hoc; weights $12/16$ and $4/16$ recover the window-level components. The whole advantage is concentrated in the extra visible targets (Table~\ref{tab:positions}). Out-of-distribution replacement sensitivity shows input dependence, not effective temporal modeling by itself.
\begin{table}[t]
\centering\small
\setlength{\tabcolsep}{4pt}
\begin{tabular}{@{}llll@{}}
\toprule
SNR & Positions & $G_s^{\rm whole}-G_s^{\rm local}$ & Interval \\
\midrule
0.1 & first12 & +0.0046 & [+0.0043, +0.0050] \\
0.1 & last4 & -0.0023 & [-0.0031, -0.0014] \\
1.0 & first12 & +0.0177 & [+0.0163, +0.0191] \\
1.0 & last4 & +0.0022 & [-0.0001, +0.0043] \\
\bottomrule
\end{tabular}
\caption{Whole-minus-local state-gain differences by target position for the first capacity step. Intervals use paired resampling of 16 training blocks.}
\label{tab:positions}
\end{table}

\paragraph{Scale and budget.} Fixing validation and test noise scales to the train-split scale preserves the qualitative boundary. Exp61 fixed four blocks by ID for 48 fits at 6,400 updates, plus 12 bridge fits assessing scale and checkpoint selection. Doubling the budget preserves dominance directions and positive whole-minus-local state gains. Four blocks support a budget sensitivity check, not optimal convergence or broad population inference. Some near-tied snapshots change identity while component estimates remain close.
\paragraph{Same-window structural comparison.} Exp62's full and shared encoders both see the window and emit one token. Shared uses a four-step shared convolution and mean pooling, with 3,072 fewer parameters. Four development and four confirmation blocks give 48 fits each. Shared has smaller state gains on confirmation, but rate, parameters and order expressivity also change, and the declared rate-matching condition fails. No structural-causal or rate-controlled claim is made; this is not evidence that architecture has no effect.
\paragraph{Post-hoc window reference (Exp63).} The existing Gaussian construction uses 64-dimensional second moments of train-fixed-scale test windows and joint entropy in bits/window. References are computed per block and endpoint, then gains are averaged over 16 blocks before taking ratios. Before computation, the primary rule required measured/reference shares to fall on the same side of 0.5 in all four cells; the secondary rule required matching directions across SNR for both designs. Both hold (Table~\ref{tab:window-reference}), but magnitudes differ and high-SNR local lies above reference. The reference includes channel randomness; agreement supplies an explanatory expectation, not identification of rate or temporal mechanisms.
\begin{table}[t]
\centering\small
\setlength{\tabcolsep}{4pt}
\begin{tabular}{@{}llll@{}}
\toprule
SNR & Design & $A_\varepsilon$ & $A_\varepsilon^{\rm ref}$ \\
\midrule
0.1 & whole & 0.8380 & 0.9273 \\
0.1 & local & 0.8845 & 0.9263 \\
1.0 & whole & 0.1010 & 0.2602 \\
1.0 & local & 0.3475 & 0.2421 \\
\bottomrule
\end{tabular}
\caption{64-dimensional window reference. Both shares are ratios of mean gains over 16 blocks.}
\label{tab:window-reference}
\end{table}

\subsection{Selection on a specified state target (Exp56, 59)}
\label{app:selection}
Each block/condition has four candidate transitions. Eligibility is selection-panel $G_x>0$ and $0<\Delta H\le0.5$ bit/step. Independent selection panels use eight panels, 16 windows each and 64 replay pairs; test uses 32 panels, 64 windows each and 128 pairs. Rules maximize $G_x,G_M,G_V,G_s$; test regret is the candidate-set maximum $G_s$ minus the selected $G_s$, with paired resampling over 16 blocks.

Audit and reconstruction select whole $4\to7$ in every block, with zero regret. The candidate dominates both targets, so this set has weak ability to distinguish the rules: no advantage is established, rather than a general absence of advantage.
\begin{table}[t]
\centering\small
\setlength{\tabcolsep}{4pt}
\begin{tabular}{@{}lll@{}}
\toprule
Rule / SNR & Regret & Interval \\
\midrule
audit\_gs / snr0.1 & 0.000000 & [0.000000, 0.000000] \\
audit\_gs / snr1.0 & 0.000000 & [0.000000, 0.000000] \\
clean\_mse / snr0.1 & 0.004531 & [0.004089, 0.004926] \\
clean\_mse / snr1.0 & 0.007925 & [0.004911, 0.010849] \\
consistency / snr0.1 & 0.004531 & [0.004092, 0.004924] \\
consistency / snr1.0 & 0.015913 & [0.014111, 0.017455] \\
equivalent raw / snr0.1 & 0.000055 & [0.000000, 0.000164] \\
equivalent raw / snr1.0 & 0.000000 & [0.000000, 0.000000] \\
reconstruction / snr0.1 & 0.000000 & [0.000000, 0.000000] \\
reconstruction / snr1.0 & 0.000000 & [0.000000, 0.000000] \\
\bottomrule
\end{tabular}
\caption{Test-set regret in $G_s$ for capacity-transition selection. The equivalent raw rule uses $G_M-G_V$ and retains its original near-tied choice.}
\label{tab:selection-updated}
\end{table}

The fifth rule $G_M-G_V$ checks an equivalent implementation; the identity does not require finite-sample choices to coincide, and the equivalent-raw row keeps that rule's own near-tied selection. The separate ridge readout is trained on the training split and scored by test-panel RankIC; it does not define success.

\subsection{Fixed-model panel sensitivity of the old association (Exp 57)}
\label{app:panel}

The prognostic endpoint resamples initializations while holding
the test panels fixed.  To isolate panel variation at fixed models, we kept the
220 frozen Exp~31 codecs and their probes unchanged and re-evaluated the same
\TAPopTransitions{} transitions on \TAPanelDraws{} fresh panel draws.  Each draw
generates a new test split for every regime (new signal and noise seeds, 32
panels, 64 entities, 256 steps) and applies the train-split normalizer and
noise scale identically to the state and the observation; the probe is not
refitted.  The Exp~31 endpoint is recomputed unchanged on each draw.

The \TAPanelDraws{} values of the endpoint have mean \TAPanelRhoMean{}, standard deviation
\TAPanelRhoSd{}, and range \TAPanelRhoMin{} to \TAPanelRhoMax{}, against the
fixed-panel value \TAPanelRhoFrozen{}.  This is measurement fluctuation at
fixed models, not variation over training-data populations; it neither replaces
the fixed-panel endpoint nor turns the association into a population-invariant law.

\subsection{Declared-law replay stress (Exp 58)}
\label{app:replay}

The audit's state component is defined by replaying the declared observation-noise
law.  Four frozen codecs---two pointwise Exp~29 codecs and two whole-window
Exp~55 codecs---are evaluated under three declared laws at the same marginal
variance as training: iid Gaussian (the training law), iid Student-$t$ with four
degrees of freedom, and a stationary Gaussian AR(1) with $\phi=0.8$.  For each
model and law we compare three quantities on the same states: a high-budget
truth (2,048 independent draws, finite-draw bias removed), an estimate that
replays the declared law, and an estimate that replays iid Gaussian noise.

For the heavy-tailed law, the Gaussian replay biases the state component by
\TAReplayWrongPointwiseHeavyHigh{} for the pointwise codec and
\TAReplayWrongTemporalHeavyHigh{} for the window codec, while the declared-law
replay gives \TAReplayCorrectPointwiseHeavyHigh{} and
\TAReplayCorrectTemporalHeavyHigh{}.  For the AR(1) law, the pointwise codec is
unchanged relative to its Gaussian marginal (matching the analytic value:
\TAReplayCorrectPointwiseArHigh{} and \TAReplayWrongPointwiseArHigh{}), whereas
the window codec differs: \TAReplayCorrectTemporalArHigh{} for the block replay
and \TAReplayWrongTemporalArHigh{} for iid replay.  These are declared stress
conditions on codecs trained under iid Gaussian noise; they are not estimates of
bias on any real dataset and do not support natural-noise identification.  What
they show is the requirement stated in the main text: the replay law must match
the target observation-noise law, and for a window codec that means the joint law, not only the
marginal.

\section{Additional utility analyses and capacity selection}
\subsection{Prediction interface and decoding}
\label{app:interface}

Three secondary studies bound how the utility measure may be read.

A prediction-interface study held
tokenizers, information set and target fixed and
compared the closed-form ridge readout against a decoder-only model over
discrete tokens with a capacity-growing output layer, and against a
supervision-source control.  Under this scoring convention
\TAIntConfirmed{} of \TAIntTotal{} transitions confirm a negative interface
difference at the Bonferroni-adjusted level and \TAIntNegativePoints{} of
\TAIntTotal{} have negative point estimates, with no positive confirmation;
intervals containing zero are reported as unconfirmed.
The scoring convention that produces this count was declared after the test
ran, layered on a preregistered contrast, and the original convention confirmed
five transitions.  Applied to the population of \secref{sec:character}, the
same convention changes per-model rank information coefficients by at most
$\TAScoringModelMax$ (median $\TAScoringModelMedian$) and leaves every capacity
comparison unchanged.  A single-arm increment is not evidence for the contrast in
either version.

Replacing sequential greedy decoding with an expectation over the fine stage
was tested as a development pilot.  Its scaling criterion required a
negative greedy capacity gain together with a full-expectation recovery of at
least $0.01$ rank information coefficient.  The greedy gain was
\TADecGreedy{}, but the recovery was \TADecRecovery{}, so the criterion was
not met and the pilot was not scaled up.  This does not show that decoding is
irrelevant; a mechanism claim would need fresh training and independent
confirmation.

Scores use average ranks for ties, zero for exact constants and the full date
axis.

\subsection{Predictive and state measures}
\label{app:closure}

\paragraph{The indirect reference.}
The ledger's reference of \secref{sec:spec} is a second-moment-matched object at the
model's own occupancy, and it answers a reconstruction question: a code of
$x_t$ is asked to reconstruct $x_t$.  The selection rule of \secref{app:rules}
needs a different object --- the error with which a code of the \emph{current}
observation conveys the \emph{same-time} state $s_t$ --- and that is the
indirect, or remote-source, Gaussian rate--distortion problem, which is closed
form.  Decompose $s = Ax + n$ with $A = \Sigma_s\Sigma_x^{-1}$ and $n$
independent of $x$.  The unpredictable part $n$ survives every code, and the
encoder's remaining task is the rate--distortion problem for $x$ under the
weighted distortion $\mathbb{E}\lVert A(x-\hat x)\rVert^2$, which reverse
water-filling solves once the weights $\gamma_i$ are read off from
$\Sigma_x^{1/2}A^\top A\Sigma_x^{1/2}$.  Writing
$\Sigma_{s\mid x} = \Sigma_s - \Sigma_s\Sigma_x^{-1}\Sigma_s$ for the
conditional state covariance, the reference error at rate $R$ and its own
water-filling level $\theta_R$ are
\begin{equation}
e_{\rm ind}(R) = \frac1d\Bigl[\operatorname{tr}\Sigma_{s\mid x}
                 + \sum_i \min(\gamma_i,\theta_R)\Bigr],
\qquad
R = \!\sum_{\gamma_i>\theta_R}\! \tfrac12\log_2\frac{\gamma_i}{\theta_R},
\label{eq:indirect}
\end{equation}
so the rate water-fills over the $\gamma_i$, the eigenvalues of
$\Sigma_x^{1/2}A^\top A\Sigma_x^{1/2}$, whereas the reconstruction reference of
\secref{app:reference} water-fills over the eigenvalues of $\Sigma_x$ itself.
The two optimize different distortion objectives and their outputs are not
interchangeable.  At $R=0$ Eq.~\eqref{eq:indirect} is the state variance
$\operatorname{tr}\Sigma_s/d$, and as $R\to\infty$ it falls to the
observation-noise floor $\operatorname{tr}\Sigma_{s\mid x}/d$.  Its measured counterpart is the held-out linear readout of \secref{app:rules}, whose population MSE is at least that of the optimal conditional-mean readout from the same code. This does not impose an ordering against a Gaussian reference evaluated on a different source model.  The prognostic study of
\secref{sec:prognostic} evaluates its state-information measure against this same
object, and the selection rule of \secref{app:rules} scores its capture against
it; both use it for the same-time state, not for a future one.  It is an
observation on this population under a Gaussian-source assumption rather than a
general invariance, and it supplies no quantity to the ledger.  Two further
studies ask whether a \emph{predictive} measure --- one built on the conditional
mean of the future rather than on the state --- would separate from
the state-readout measure on this generator.

These closure probes use linear-readout $R^2$, distinct from the codec distortion $D_s$ (Table~\ref{tab:accounts}).

The first study is post-hoc and from the same session.  It rebuilds the oracle conditional
mean in closed form from the generator.  Across \TAPSTransitions{} transitions the two
measures agree in sign \TAPSSign{} times, the within-cell rank correlation
between them runs \TAPSCellLo{} to \TAPSCellHi{}, and the ratio of the two
measures is a constant per cell with no trend along the ladder.  A possible structural explanation is that with a linear
target read by a linear reader, the predictive measure is the state-readout measure
re-weighted by a fixed matrix.  A fixed re-weighting on its own does \emph{not}
force two configurations to keep the same ordering --- that additionally requires
the error's anisotropy to stay fixed as capacity grows, which is a property of
this population rather than of the algebra.

A follow-up tested an anisotropic low-variance target and
predicted that a low-variance channel's ratio would rise with capacity.  The
prediction failed: \TADirectSeparated{} of \TADirectComparisons{} low-variance
comparisons separate, the median absolute change in the ratio over those
comparisons is \TADirectMedianAbs{}, below its own standard error at a median
ratio of \TADirectMedianRatio{}, and exactly \TADirectFlags{} of
\TADirectTests{} cell-by-channel comparisons flag, on the reconstruction-aligned
channel with the opposite sign.  That is what a two-standard-error screen over
that many tests is expected to produce by chance.  On this generator, then, the
two measures were not separable, the measured anisotropy was stable along the
ladder, and the low-variance-target test did not separate them.  These findings concern this generator and target construction;
they leave predictive-capacity rules on other generators unresolved.

\subsection{Incremental predictive value}
Adding the state component did not improve out-of-regime prediction on this grid.
The incremental-value entry is the auxiliary regression behind this result.  Its endpoint is a difference of pooled out-of-fold mean squared
errors, not a regression coefficient.  Holding the specification,
population, folds, penalty grid and replicate count fixed, the regression gives a base out-of-fold
MSE of $\TAIVBase$ and an enhanced one of $\TAIVEnhanced$, an MSE change of
$\TAIVCorrected$ with an interval of
$\TAIVInterval$.  Adding the state component increases held-out error: the MSE change is defined as base
minus enhanced, so a negative value means worse prediction.
The selection rules score a different object, the validation
state-information capture, and do not outperform direct task validation on this
grid.

The response of that regression is the seed-mean utility gain $\GU$ of a
transition, and its base design already carries the reconstruction gain $\Gx$,
the realized occupancy increment $\Delta H$, the lower rung's nominal bits per
stage, the step between the two rungs and the quantizer family.  The test
therefore asks whether the state component adds predictive value beyond these
conventional diagnostics.  The primary design adds
$\Gs$; the secondary design adds the reference-relative excess $\Aeps-\Aref$
instead.  $\Geps$ is not entered, because $\Gx=\Gs+\Geps$ makes it collinear with
the two gains already present, and no combination is selected after the fact.
The outer loop is leave-one-regime-out over \TAIVFolds{} folds, so a whole ladder
stays on one side of the outer boundary, and the penalty is chosen inside the
training folds by five-fold cross-validation on a fixed grid with the same folds
and seed for every design.  The random inner folds can split transitions
sharing an intermediate rung, a limitation retained from the declared design.
Only the primary comparison has an interval; the secondary is a point estimate.

For the secondary analysis, the base is refit on the same sample as the enhanced
model.  Adding the reference-relative excess also increases error.  The stored values are:
\begin{center}\small
\begin{tabular}{@{}lr@{}}
\toprule
secondary comparison & out-of-fold MSE \\
\midrule
base & $\TAIVSecondaryBase$ \\
enhanced & $\TAIVSecondaryEnhanced$ \\
change (base minus enhanced) & $\TAIVSecondaryChange$ \\
\bottomrule
\end{tabular}
\end{center}
The primary comparison improves in \TAIVFoldsImproving{} of \TAIVFolds{}
folds, so its direction also varies across regimes.  This grid establishes no
stable incremental predictive value.  That finding concerns added prediction
under the specified regression, not the existence of the association reported
in \secref{sec:prognostic}.

\subsection{The selection rules and the regret}
\label{app:rules}

\paragraph{State-information capture.}
The audit-derived rule scores each rung by a normalized drop in same-time state
reconstruction error,
\begin{equation}
\text{capture} = \frac{\sigma_s^2 - \hat e_{\rm meas}}
                     {\sigma_s^2 - e_{\rm ind}},
\label{eq:capture}
\end{equation}
where $\sigma_s^2 = \operatorname{tr}\Sigma_s/d$ is the state variance, i.e.\ the
error of predicting $s_t$ by its mean.  Both numerator and denominator are
therefore mean-squared-error drops relative to that variance baseline, and the
quantity is a normalized variance-explained ratio, not a ratio of mutual
informations.

Two objects in Eq.~\eqref{eq:capture} are distinct from the ledger's, and the
distinction is the point.  The measured term $\hat e_{\rm meas}$ is the held-out
ridge mean squared error of reading the \emph{same-time} state $s_t$ from the
\emph{same-time} code, fitted on the training panels and scored on the validation
panels; its penalty is selected on an inner slice by state MSE, not by the
per-step cross-sectional rank information coefficient the utility probe
maximizes, so it is neither $\Ds$
nor a next-step utility probe.  The reference term $e_{\rm ind}$ is the
\emph{indirect} (remote-source) Gaussian reference of \secref{app:closure}: the
error with which a memoryless Gaussian code of the observation $x_t$ at the
rung's own realized joint-token entropy conveys $s_t$, the same-time state.  It
is not the reconstruction reference of \secref{app:reference}.

Within that stated problem --- the Gaussian surrogate, that rate and squared loss
--- $e_{\rm ind}$ is the best a memoryless code can do, so the denominator is the
largest drop available. For a readout at least as good as the mean predictor, the population ratio lies in $[0,1]$ under that Gaussian model.  Read on the
actual population it is bounded more weakly.  $0$ means that this linear
readout matches the variance baseline (negative values mean worse MSE), which does not establish
that the code carries no state information at all.  $1$ says only that the
measured error equals the Gaussian reference, which is not global optimality on a
non-Gaussian source.  The value is not clamped.  A ridge fitted in-sample and
scored out of sample can fall marginally below the reference, so values slightly
above one occur, and that over-shoot is not by itself evidence of a defect.  The
capture is a ratio, it is not the reference-relative excess $\Eeps$, and it does
not enter the ledger.

\paragraph{The five rules.}
Let $b$ range over the available rungs and let $\bar q_b$, $\bar r_b$ and
$\bar c_b$ be the mean over seeds at rung $b$ of the validation rank information
coefficient, the validation reconstruction error and the validation capture.
\emph{Largest} takes $\arg\max_b b$.  \emph{Lowest reconstruction error} takes
$\arg\min_b\bar r_b$.  \emph{Highest validation utility} takes
$\arg\max_b\bar q_b$.  \emph{One standard error} takes
$b^\star=\arg\max_b\bar q_b$, the standard error of the seed mean
$s_b=\operatorname{std}(\{q_{b,\text{seed}}\},\mathrm{ddof}=1)/\sqrt{n}$, the
threshold $\theta_{\rm 1SE}=\bar q_{b^\star}-s_{b^\star}$, and then the
\emph{smallest} rung with $\bar q_b\ge\theta_{\rm 1SE}$.  \emph{Highest capture}
takes $\arg\max_b\bar c_b$.  All
five depend only on validation-side scores.

\paragraph{Regret.}
With $\bar U_{\rm test}(b)$ the seed-mean test utility at rung $b$, the oracle
rung is $b^\star=\arg\max_b\bar U_{\rm test}(b)$ and
$\text{regret}=\bar U_{\rm test}(b^\star)-\bar U_{\text{test}}(\hat b)$.  The
table reports the mean of that quantity over the \TACells{} cells together
with the count of cells off the optimum.  It is a descriptive mean: no interval
is computed over cells, so the entries carry no paired uncertainty.  Mean regret
is \TARegretLargest{} for the largest rung, \TARegretOneSE{} for the
one-standard-error rule, \TARegretState{} for the audit-derived rule and
$\TARegretValidation{}$ for reading the task on validation.
Table~\ref{tab:regret} reports all five rules side by side.

% Generated by research_rebuild/make_tokaudit_assets.py
% Every value in this file is read from a frozen bundle.  Do not hand-edit.
\begin{table}[ht]
  \centering\small
  \begin{tabular}{@{}lrr@{}}
    \toprule
    selection rule & mean test regret & cells off best \\
    \midrule
    largest available rung & $0.00116$ & 2 \\
    lowest validation reconstruction error & $0.00116$ & 2 \\
    one-standard-error rule & $0.00057$ & 6 \\
    \textbf{highest validation utility} & $4.71\times 10^{-6}$ & \textbf{1} \\
    highest state-information capture (audit-derived) & $0.00126$ & 3 \\
    \bottomrule
  \end{tabular}
  \caption{Capacity selection over \TACells{} mechanism--regime cells. Regret is the difference between the best and selected capacities\textquotesingle{} seed-mean test utility, averaged over cells. Results are descriptive post-hoc means; between-rule differences are untested.}
  \label{tab:regret}
\end{table}

\FloatBarrier
\section{Simple statistics and the conditional-mean state component}
\label{app:simple}
\paragraph{Targets and populations.} All quantities use per-element squared-error units. $M=D_s+V$ and gains are low minus high capacity, hence $G_s=G_M-G_V$. Exp54 re-measures 128 frozen models of the independent training-objective confirmation (Exp52; 64 repetition-level transitions) and \TANFitB{} models of the capacity sweep (Exp29; \TAAltTransitionsB{} transitions aggregated over initializations). These are not 118 homogeneous independent samples. Replay uses 128 and 64 pairs respectively; independent views from different antithetic pairs estimate $V$.
\paragraph{Readouts.} Differences use shared draws, and correlations use average ranks. Exp52 row intervals resample 32 panels conditional on one training repetition; Exp29 row intervals resample initialization seeds and panels, paired across quantities.
\begin{center}\small\begin{tabular}{lrr}\toprule
 & Exp52 & Exp29\\\midrule
$G_M,G_s$ sign disagreements & \TAAltSignRawA & \TAAltSignRawB\\
Both gain intervals exclude zero & \TAAltSignA & \TAAltSignB\\
$G_M-G_s$ interval excludes zero & \TAAltConflA & \TAAltConflB\\
Spearman$(G_V,G_s)$ & \TAAltRhoConsA & \TAAltRhoConsB\\
Spearman$(G_x,G_s)$ & \TAAltRhoGXA & \TAAltRhoGXB\\
Median $|G_M-G_V-G_s|$ & \TAAltEquivMedA & \TAAltEquivMedB\\\bottomrule
\end{tabular}\end{center}
Every sign disagreement has $G_M<0,G_s>0$: the conditional mean improves while increasing output variance can worsen single-draw error. The identity holds for the estimands; separately estimated finite-replay quantities can differ. A post-hoc breadth screen did not establish a broad collection of composition differences at approximately matched conventional measurements; it does not support general ranking superiority.
\paragraph{Target scope.} Downstream models see single-observation tokens; $D_s$ is not proposed as a superior deployment objective. For $f_\lambda(x)=\lambda x$, $V=\lambda^2\sigma^2$ and $D_s=(1-\lambda)^2\operatorname{Var}(s)$: reduced variance can worsen conditional-mean state distortion. Consistency alone is not a state component.

\section{External and public-data studies}
\subsection{Beijing multi-site stress test}
\label{app:multisite}

A second external study injects known i.i.d. Gaussian observation noise into the
clean Beijing multi-site air-quality series (\TAExpFortySevenStations{}
stations, \TAExpFortySevenChannels{} channels, \TAExpFortySevenHours{} hourly
timestamps; per-season 60/20/20 splits; blocks of
\TAExpFortySevenPanelHours{} retained hourly observations; train-only global
per-station/channel normalization, no window statistics).  A block is cut from
the sequence of retained records and is not guaranteed to be \TAExpFortySevenPanelHours{}
consecutive natural hours: a timestamp survives only if every station reports
every channel after gap filling, so a block can span a real-time gap, and the
readout's next step is the next retained record rather than necessarily the
next hour.  Three cells (clean control, target SNR 0.3 and 1.0) $\times$ two
mechanisms (RBQ and VQ-VAE) $\times$ four widths $\times$
\TAExpFortySevenSeeds{} seeds give \TAExpFortySevenCodecs{} codecs under one
shared budget; each noise cell has \TAExpFortySevenTransitions{} transitions.

\paragraph{Endpoint.}
The primary endpoint
is the stratified Spearman
$Q=\tfrac12\rho(\Eeps,\GU\mid\text{SNR }0.3)
  +\tfrac12\rho(\Eeps,\GU\mid\text{SNR }1.0)$
with $H_1\colon Q<0$; a \TAExpFortySevenBootstrapReplicates{}-draw
season/panel/seed bootstrap gives $Q=\TAExpFortySevenQ$ with a 95\% interval
(2.5/97.5 percentiles) $\TAExpFortySevenInterval$, and \TAExpFortySevenBelowLow{}/\TAExpFortySevenTransitions{}
and \TAExpFortySevenBelowHigh{}/\TAExpFortySevenTransitions{} transitions fall
below the reference at the two noise levels (per-cell
$\rho=\TAExpFortySevenRhoLow$ and $\TAExpFortySevenRhoHigh$).  The
clean cell's noise-related component is exactly zero (worst per-model mean
\TAExpFortySevenCleanWorst{}).  Each season draws its own noise
realization, and the bootstrap weights the per-panel reference curves by the
draw's panel multiplicities.

\paragraph{Time-axis construction.}
Missing values are filled by linear interpolation that fills at most six
consecutive missing positions from each end of a gap, so a longer gap can be
filled from both sides and ``only gaps of six hours or less are filled'' is not
what the implementation guarantees; in the retained series a filled value is at
most \TAExpFortySevenInterpMaxDistance{} records from an observed one
(\TAExpFortySevenInterpFilled{} values filled).  Interpolation runs
before the seasonal split, so a filled value inside one split can draw on an
observation assigned to another split.  The split-respecting sensitivity below
measures the effect of removing this exposure.  A post-hoc structural
description of the axis
reports
\TAExpFortySevenHourlyPct\% of the \TAExpFortySevenRetainedSteps{}
steps between adjacent retained records to be exactly one hour, with
\TAExpFortySevenGapSteps{} steps crossing a gap and a largest gap of
\TAExpFortySevenLargestGapHours{} hours.  Of the readout's four-step
windows and their next-step labels, at most
\TAExpFortySevenWindowCrossTestPct\% of the windows and
\TAExpFortySevenLabelCrossTestPct\% of the labels in a test split
cross a gap (\TAExpFortySevenWindowCrossAllPct\% over every
split, training included).  These are descriptive properties of the
constructed axis; the reported endpoint is
not re-estimated under a different time axis.
\paragraph{Robustness of the sign.}
The sign survives leave-one-season-out
$[\TAExpFortySevenLooSeasonLo, \TAExpFortySevenLooSeasonHi]$, per-mechanism
(\TAExpFortySevenMechBsq{} for RBQ and \TAExpFortySevenMechVqvae{} for VQ-VAE)
and leave-one-station-out
$[\TAExpFortySevenStationLo, \TAExpFortySevenStationHi]$ checks.  This is one city
and one data domain with injected known noise, not a natural-noise
identification, and it does not replace the inconclusive Saugeen endpoint.

\paragraph{Split-respecting fill.}
The interpolation above can draw across a split boundary.  A post-hoc
sensitivity (Exp~50) removes that exposure
by filling each split from its own observations only.  \TAExpFiftyDropped{} of the
\TAExpFortySevenRetained{} retained records drop out because their fill is no
longer reachable inside their own split --- all of them in the training and
validation splits, none in the test splits --- and the endpoint moves from
$\TAExpFortySevenQ{}$ to $\TAExpFiftyQ{}$ with a 95\% interval
$\TAExpFiftyInterval{}$.  The sign is unchanged and the interval stays clear
of zero in the same direction, so the cross-split fill is not driving the
reading.  This is a sensitivity on the same archive and the same measurement
path, not an independent replication, and it does not replace the primary
Beijing endpoint.

\subsection{Saugeen stress test}
\label{app:saugeen}

The negative-association endpoint is inconclusive.  It
runs on \TASaugeenNUsed{} of the \TASaugeenNDesigned{} designed cells --- the
rule excludes one whose reconstruction gain is not positive ---
where $\rho=\TASaugeenRho{}$ with interval \TASaugeenInterval{} and sign
agreement \TASaugeenSign{}, over \TASaugeenPoints{} daily points.  It is
not a replication success or failure.

The study injects known corruption into a real hydrological series.  It compares
a scratch-trained capacity sequence with a tokenizer pretrained on other public
domains.  The latter is our own pretraining on those domains.  The design has
the following limits.

\begin{center}\small
\begin{tabularx}{\textwidth}{@{}>{\raggedright\arraybackslash}p{0.14\textwidth}>{\raggedright\arraybackslash}X>{\raggedright\arraybackslash}X@{}}
\toprule
item & design as run & interpretation \\
\midrule
budget & Scratch: batch \TASaugeenScratchBatch{}, \TASaugeenScratchUpdates{} updates, across forty-eight fits. Pretrained arm: \TASaugeenGenBatch{}, \TASaugeenGenUpdates{}.
& The arms differ by \TASaugeenSampleRatio{} in sample-updates. Matching codebook size does not isolate pretraining. \\
side information & Window mean and standard deviation reach the decoder through a continuous path.
& Occupancy measures token indices only; the scalar reference is a proxy. \\
VQ objective & Encoder-side weight $1$; codebook-side weight \TASaugeenCommitment{}. These swap the published paths.
& The arm is architecture-inspired. Retraining with the published objective would be a separate study. \\
\bottomrule
\end{tabularx}
\end{center}

\paragraph{Side information and rate convention.}
The codec reconstructs as
$\hat x = \mathrm{std}(x)\cdot\mathrm{decoder}(q(\mathrm{norm}(x))) +
\mathrm{mean}(x)$, so the window's own mean and standard deviation enter
through a continuous path that the token indices do not carry.  Holding the
token indices fixed, translating a window by
\TASaugeenTranslate{} moves the reconstruction by the same amount and leaves
the tokens unchanged.  The occupancy measure is therefore a \emph{token-index}
occupancy that bounds the discrete code, not the pipeline's total rate.  The
scalar reference receives neither the side information nor the same reader
interface, which makes it a declared proxy rather than the synthetic studies'
comparison.  One capacity doubling is \TASaugeenBitsPerToken{} bit per token,
\TASaugeenBitsPerWindow{} bits per window and
\TASaugeenBitsPerPoint{} bit per time point; the nominal step and the realized
occupancy increment are reported in the same per-window unit, with
$\Delta H=\Rocc(C_2)-\Rocc(C_1)$.

\paragraph{Clean control.}
The corruption-free cells inject no observation noise, and the noise-related component is
zero in all \TASaugeenCleanModels{} clean fits of the scratch ladder.  These zeros follow from the construction: without injected noise the
estimator's noise term vanishes identically, and accuracy with nonzero noise is
assessed separately by the failed finite-replica criterion in \secref{app:validity}.  The multi-domain tokenizer's own clean fit is also zero but is not counted among the \TASaugeenCleanModels{}.

\subsection{The public-data operational application}
\label{app:real}

The operational application declares a state and a residual.  The state is a
causal per-channel seasonal mean --- the mean of the same phase over the
previous \TARealSeasonalDays{} days, at a seasonal lag of
\TARealSeasonalLag{} rows for the dataset analysed --- with a burn-in of
\TARealBurnIn{} rows.  The residual is resampled as independent noise per
channel, and its lag-one autocorrelation is reported
(\TARealLagOneLo{}--\TARealLagOneHi{}): the residual is temporally dependent, and
independent resampling does not preserve that dependence.  That is the
diagnostic for how defensible that declaration is.  Distortions are computed on
\TARealStates{} sampled states at \TARealReplicas{} replicas over
\TARealRepeats{} independent noise repeats.

Two public datasets were carried
into the study; the guard on the state construction fails on the
second, where the declared state does not track the observation, and
only the first is carried forward.  Because the split is a modeling
choice, no number in this appendix is ground truth, and the state and noise-related
components here are not comparable to the synthetic ledger's.  On the dataset
carried forward, the measured noise share of reconstruction error runs from
\TARealShareLo{} to \TARealShareHi{}, and the transition noise share of
reconstruction gain from \TARealGainLo{} to \TARealGainHi{}; measuring the
predictive value of the state gain relative to the task increment would need
more transitions than the study has.

\subsection{Replay-law diagnostic}

A stationary Gaussian process with every parameter declared was used to study
the operational measurement under known truth: state AR(1) with $\phi=0.9$;
observation noise AR(1) with $\phi_e\in\{0,0.5,0.9\}$ plus a shared
cross-channel factor; a lagged declared state built from the past; three codec
families (pointwise and windowed nonlinear soft-threshold ladders, and a linear
control); and four replay laws (iid, block, conditional, oracle).  Two truths are computed independently
of the estimator: the conditional truth
$\mathbb{E}\lVert\hat{s}-\mathbb{E}[f(x)\mid\hat{s}]\rVert^2$ under the true
law, and the physical truth with the true state in place of the declared one.

For the pointwise soft-threshold codec at the lower rung, the estimator's output
$D_s$ is \TAPatReplayDsPhiZero, \TAPatReplayDsPhiHalf{} and \TAPatReplayDsPhiNine{} at
$\phi_e=0,0.5,0.9$, against conditional truths of \TAPatReplayTruthPhiZero,
\TAPatReplayTruthPhiHalf{} and \TAPatReplayTruthPhiNine---ratios of \TAPatReplayRatioPhiZero,
\TAPatReplayRatioPhiHalf{} and \TAPatReplayRatioPhiNine.  Across all four replay laws and
all three noise laws, target/estimator lies in \TAPatReplayTargetEstimatorLo--\TAPatReplayTargetEstimatorHi, so the estimator faithfully
converges to its own estimand; target/conditional truth lies in
\TAPatReplayTargetTruthLo--\TAPatReplayTargetTruthHi{} for every law, including the
conditional law (\TAPatReplayConditionalPhiZero{} at $\phi_e=0$) and the oracle law
(\TAPatReplayOraclePhiZero).  The linear control's $D_s$ is
identical across all four replay laws (spread $\TAPatReplaySpreadZero$,
$\TAPatReplaySpreadSmall$, $\TAPatReplaySpreadLast$
across the three noise laws), which is what the algebra requires.

The estimator forms $Y_j=(f(s+e_j)+f(s-e_j))/2$, so it converges to the
conditional mean under the symmetrized replay law.  A replay law whose draw
carries a nonzero mean given the declared state---exactly what the conditional
law adds---has that mean cancelled by the pairing.  When the declared state is
built from the past, the residual is not conditionally centred given it (the
fitted residual predictor explains \TAPatReplayExplainLo--\TAPatReplayExplainHi\% of
the residual variance in this process), and the measured state component is
\TAPatReplayStateLo--\TAPatReplayStateHi\% of the natural estimand.
At $\phi_e=0.9$, block replay preserves the residual's within-window lag-one
autocorrelation (\TAPatReplayBlockAc{} against \TAPatReplayResidualAc) and iid does not
(\TAPatReplayIidAc), yet both give the same
estimates in this diagnostic: preserving temporal dependence alone does not recover the conditional relation between residual and declared state.  This known-truth diagnostic shows why an operational state component can be
substantially below the declared-state fidelity.  It does not estimate bias in
any real dataset.  The synthetic populations satisfy the distinct conditions
of known latent state and conditionally zero-mean observation noise.

\subsection{The motivating production system}
\label{app:ashare}

The benchmark was motivated by an equity-forecasting pipeline in which a
tokenized representation is trained first and a downstream sequence model is
trained on its tokens.  Over a long continuation run of that system the
tokenizer's own health indicators improved and its reconstruction error kept
falling, while the downstream ranking quality did not follow.  That is the
observation that led to asking what a capacity increment preserves.  No number
from that system appears in this paper: its evaluation window has been consumed
and its data come from Wind, so it serves only as motivation and
carries no benchmark evidence.  Its protocol and scoring conventions are
described in \secref{app:interface} only to the extent that the scoring
convention was later adopted by the benchmark.

\FloatBarrier
\FloatBarrier
\section{Reproducibility}
\label{app:repro}

Code and protocols will be released with the paper.  All empirical data are
public series --- the Beijing multi-site air-quality data, the Monash Saugeen
streamflow series and the public datasets of the operational application of
\secref{app:real} --- alongside the controlled synthetic generators of
\secref{app:regimes} and \secref{app:crossgenerator}.

The release provides two routes.  Given the original frozen result bundles, the first
reconstructs the reported numbers on CPU, without loading models or training.
The second measures a \emph{new} transition: it trains a small codec pair at two
capacities, runs the estimator of \secref{app:controls} and the reference of
\secref{app:reference} on them, and emits a record conforming to the ledger
schema.  The codec pair it trains is far smaller than this paper's and shares
none of its data, so the numbers that run writes are new measurements and not
results of this paper.  The second route is codec-specific: it covers the
supplied RBQ, VQ-VAE and analytic uniform quantizer; measuring another codec
requires a compatible extraction path.  These two routes do not establish
whole-pipeline reproducibility.

Experiment identifiers (Exp~NN) locate implementations and records in the
source archive; the study-to-identifier map and each study's specification
status are part of the released research records.

\end{document}